\documentclass{article}

\usepackage[main,final,nonatbib]{neurips_2026}
\usepackage[utf8]{inputenc}
\usepackage[T1]{fontenc}
\usepackage{amsmath,amssymb,amsthm,mathtools}
\usepackage{booktabs}
\usepackage{graphicx}
\usepackage{microtype}
\usepackage{xcolor}
\usepackage{url}
\usepackage{placeins}

\newtheorem{theorem}{Theorem}
\newtheorem{proposition}{Proposition}
\newtheorem{lemma}{Lemma}
\newtheorem{corollary}{Corollary}
\newtheorem{remark}{Remark}

\title{How Should Parallel Langevin Chains Share Noise?}

\author{Ryan Farell\\
Department of Computer Science, The University of Texas at Austin\\
\texttt{ryan.farell@utexas.edu}}

\begin{document}
\maketitle

\begin{abstract}
Parallel Langevin sampling usually draws an array of independent Gaussian
vectors, one row per chain. The average row drives the ensemble mean, with
variance $1/K$ per coordinate for $K$ chains. Yet the unadjusted Langevin
algorithm (ULA) specifies only each row's distribution. Subtracting the
column means and rescaling, as in classical antithetic sampling, makes that
variance zero. Each chain keeps its ordinary ULA path law for any finite run.
For quadratic targets we derive the joint covariance of linear estimates
computed from the ensemble mean, including coefficient estimates, time
averages and predictions. The identity separates contributions from initial
states and Gaussian draws. The coupling changes the second through one scalar,
which is zero whenever the draws sum to zero across chains. For nonquadratic
targets we bound the remaining variance. Centering has a price: for a quadratic
form in the mean-subtracted Gaussian state, the ensemble average has variance
$1/(K-1)$ relative to one chain, compared with $1/K$ under independent draws.
On three tempered logistic posteriors, eight centered chains reduced the
variance of the posterior-mean estimate, summed over coordinates, by two to
three orders of magnitude compared with eight independent chains. They also
beat 128 independent chains on variance and on variance times sampling time.
Under Metropolis correction the reduction shrank to about three- to eightfold:
rejected draws no longer balance accepted ones. The measured variance ratios
were consistent with a leading-order prediction from rejection frequency
and noise energy.
\end{abstract}

\section{Introduction}

Bayesian computations often average many Langevin chains to estimate a
posterior mean or a linear prediction. Each chain takes gradient steps
perturbed by Gaussian noise. Accelerators make it natural to launch $K$
chains and average their outputs for the usual $1/K$ Monte Carlo variance factor. If we want a further tenfold reduction in
variance by adding independent chains of the same length, we need ten
times as many of them. The question is whether we can make better use of
the chains we already have, at the same run length.

\begin{figure}[!ht]
\centering
\begin{minipage}{.49\linewidth}
For $K\ge2$, draw independent standard Gaussian vectors
$\zeta_1,\ldots,\zeta_K$ and stack them as the rows of an array.
In each column,
\begin{equation}\label{eq:construct}
  \bar\zeta=\frac1K\sum_j\zeta_j,\qquad
  \xi_i=\sqrt{\frac K{K-1}}(\zeta_i-\bar\zeta).
\end{equation}
The $\xi_i$ sum to zero in every column, and the scale factor restores
unit variance to each entry.
\end{minipage}\hfill
\begin{minipage}{.44\linewidth}\includegraphics[width=\linewidth]{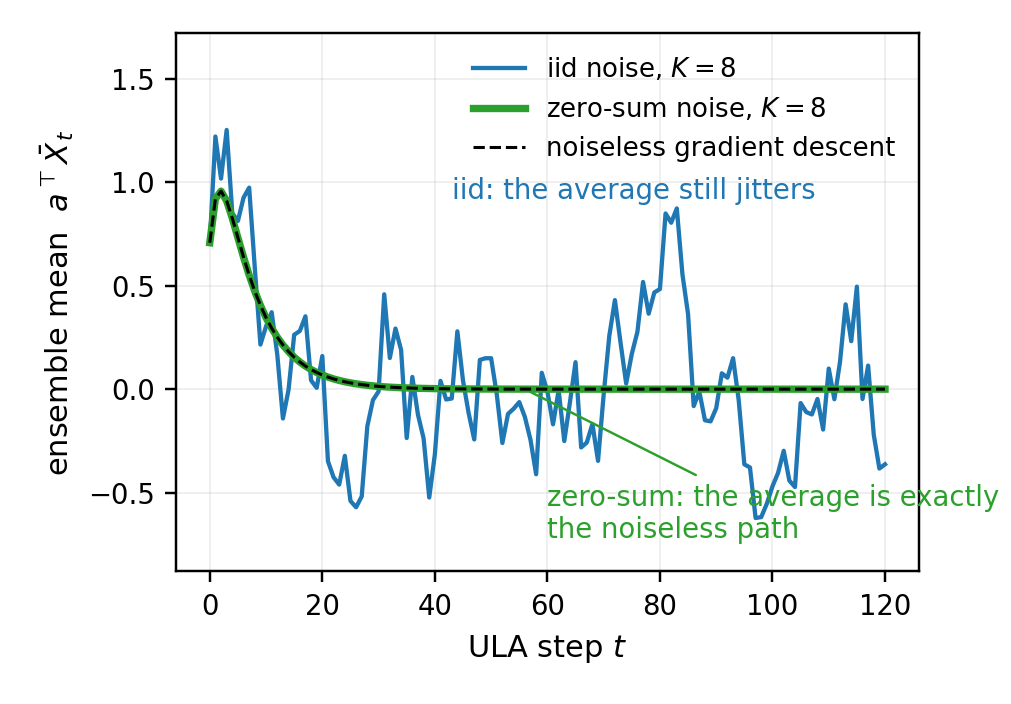}\end{minipage}
\caption{The average of eight ULA chains on a quadratic target from a common
start. With independent noise the average fluctuates. With zero-sum noise it
follows gradient descent to machine precision. A single chain has the same
law in both runs.}\label{fig:intro-schematic}
\end{figure}

Many GPU samplers already advance all the chains in one call
\cite{lao2020tfp,phan2019numpyro,cabezas2024blackjax,duffield2025posteriors}.
For Langevin sampling, the random-number generator supplies an array of
Gaussian draws, with a row for each chain and a column for each coordinate.
What if the draws in each column cancelled across chains?
Subtract the mean across chains in each coordinate, then rescale to
restore unit variance. On a quadratic target from a fixed common start,
the ensemble mean, the average of the $K$ chain states at each step,
follows gradient descent exactly (Figure~\ref{fig:intro-schematic}).

All eight chains in Figure~\ref{fig:intro-schematic} are random even when
their average follows gradient descent. A ULA chain updates its state with a
fresh standard Gaussian vector, its innovation. The algorithm does not specify
its dependence on other chains' draws at that step. Fresh means standard
Gaussian given everything all $K$ chains have done so far.
Craiu and Meng used this freedom to couple Markov chains~\cite{craiu2005multiprocess}.
Here we compare couplings through the covariances of the estimates they produce
in a finite run. Section~\ref{sec:related} separates this from couplings of the
dynamics, in which the drift changes with the noise~\cite{ma2015recipe} or a
state-dependent covariance needs its own correction~\cite{nusken2019note}.

The code changes only the Gaussian array passed to the ULA update.
Centering acts down each column, across chains, and the last line restores
the variance of each entry:

\begin{center}
\begin{minipage}{.88\linewidth}
\begin{verbatim}
xi = rng.standard_normal((K, d))  # one step: ordinary draws
xi -= xi.mean(axis=0)             # center across chains
xi *= np.sqrt(K / (K - 1))        # restore N(0, I) marginals
\end{verbatim}
\end{minipage}
\end{center}

Centering and rescaling preserve the entire path law of each chain
(Proposition~\ref{prop:marginal}), hence also the bias of a posterior-mean
estimate at a given step size and run length. The variance can change
substantially: on three tempered logistic posteriors, eight centered
chains beat 128 independent chains on the trace variance of the posterior-mean
estimate (its variance summed over coordinates) and on variance times sampling
time (Section~\ref{sec:experiments}). The second comparison includes the cost of centering,
which on one device costs one mean over the chain axis per step. Across devices, centering would require
one length-$d$ all-reduce per step.

\paragraph{Contributions.}
For a quadratic target, Theorem~\ref{thm:panel} gives the joint covariance of
a collection of linear ensemble-mean estimates, called a panel. One scalar
scales the innovation contribution to every entry, including the covariance
between a running average and a last-step prediction. One coupling serves
them all, even before the predictions are specified.

For ULA with Gaussian noise and affine drift, zero-sum draws remove the
innovation contribution to every linear ensemble-mean estimate; variance from
random starts can remain. At even $K$, pairing and centering both achieve this,
but for $K>2$ they give different variance penalties for the square of a
mean-subtracted coefficient (Theorem~\ref{thm:hermite}). When the Hessian varies,
the gradient averaged across chains can differ from the gradient at the ensemble
mean, so even a coefficient mean at the last step can retain variance from the
noise. Theorem~\ref{thm:variance-ratio} bounds this residual under curvature and
horizon assumptions. We also measure how much of the gain remains under
Metropolis adjustment and a fitted control variate~\cite{mira2013zero,oates2017functionals}.

\section{Background}\label{sec:background}
\subsection{Parallel ULA and the estimates from one run}
Let $U$ be the negative log target density and $X_t^{(i)}$ the state of
chain $i$ after $t$ steps, such as a regression coefficient vector. With
step size $\eta>0$, ULA~\cite{roberts1996exponential} updates it by

\begin{equation}\label{eq:ula}
 X_{t+1}^{(i)}=X_t^{(i)}-\eta\nabla U(X_t^{(i)})+\sqrt{2\eta}\,\xi_t^{(i)} .
\end{equation}
Here $\xi_t^{(i)}$ is the innovation of chain $i$ at step $t$. It must have
distribution $N(0,I_d)$ conditional on everything sampled before that step, by
this chain or any other. It is fresh in time, even if dependent on the
innovations of other chains at the same step.

For a coefficient mean, average its coordinate over chains at each saved
step, then over time. Write $\bar X_t=K^{-1}\sum_i X_t^{(i)}$ for the
ensemble mean. For a new covariate vector $a$, average $a^\top\bar X_t$
at those steps to estimate the mean linear prediction.
These estimates average single-chain quantities, so their expectations do not
depend on the coupling (Section~\ref{sec:marginal}). Their covariance does, and we compute it
in Section~\ref{sec:quadratic}.

\subsection{Coupling constructions}
There are two easy ways of filling a column of the array of draws
in \eqref{eq:construct}. Draw a new number for each chain and
the column mean has variance $1/K$. Copy the first number to every row and
the mean has variance one. To describe everything in between, and how to
get below $1/K$, let $C$ be the $K\times K$ correlation matrix of one
column. We use Gaussian columns and draw them independently across the $d$
coordinates, so the full covariance is $C\otimes I_d$. We call $C$
admissible if it is positive semidefinite with ones on the diagonal.
Writing $\mathbf1$ for the vector of $K$ ones, the variance of a column
mean is $\alpha(C)$. The centered construction has correlation matrix
$C_\star$, where

\begin{equation}\label{eq:alpha}
 \alpha(C)=\frac{\mathbf1^\top C\mathbf1}{K^2},\qquad
 C_\star=\frac{K}{K-1}\left(I_K-\frac{1}{K}\mathbf1\mathbf1^\top\right).
\end{equation}
Independent draws give $\alpha=1/K$ and a common draw gives $\alpha=1$.
We call $C$ zero-sum when $\alpha(C)=0$. For an admissible $C$ this means
$C\mathbf1=0$: every column sums to zero.
Under $C_\star$, each entry equals minus the sum of the other $K-1$ entries.
Each has unit variance and every pair has the same correlation $c$. Taking
covariances with the chosen entry gives $1=-(K-1)c$, so $c=-1/(K-1)$.

For even $K$, we could instead pair each draw with its negative~\cite{hammersley1956new}.
A pair $z,-z$ averages to zero, but its two squares are identical, so
averaging them gains nothing. The quadratic variance therefore depends on
which zero-sum construction we choose. Within this class, only $C_\star$ is
invariant to relabeling the chains (Table~\ref{tab:factors}).

\begin{table}[ht]
\centering\small
\caption{Of these four choices, only $C_\star$ cancels the column mean at every
$K\ge2$ and treats all chains alike. Its price is a quadratic factor of
$1/(K-1)$ in place of $1/K$. The first and last rows give ensemble-average variance relative to one chain
for a linear observable and a quadratic form in the mean-subtracted state,
under the Gaussian assumptions of
Theorem~\ref{thm:hermite}. The factors for pairs assume even $K$.}\label{tab:factors}

\begin{tabular}{lcccc}
\toprule
 & iid & common & pairwise & $C_\star$ \\
\midrule
linear factor $\alpha(C)$ & $1/K$ & $1$ & $0$ (even $K$) & $\mathbf 0$ \\
zero-sum at odd $K$ & --- & --- & no & yes \\
invariant to relabeling ($K>2$) & yes & yes & no & yes \\
quadratic factor & $1/K$ & $1$ & $2/K$ & $1/(K-1)$ \\
\bottomrule
\end{tabular}
\end{table}

\subsection{Related work}\label{sec:related}
Coupling valid Markov kernels through their random draws goes back to
antithetic MCMC~\cite{green1992metropolis,frigessi2000antithetic,craiu2005multiprocess}.
Craiu and Meng already give the exchangeable correlation in \eqref{eq:alpha}.
Their theorem gives covariance signs under monotonicity and negative
association. On a quadratic target, propagating Gaussian draws through the
ULA recursion gives the covariance they contribute to estimates at different
saved times, without monotonicity. One zero-sum coupling removes that
contribution from every linear ensemble-mean panel. Adaptive and differentiable antithetic designs for Monte
Carlo and variational inference choose dependence between samples to
suit an objective
\cite{ren2019adaptive,wu2019differentiable}. They learn the dependence for
an objective; here it is fixed in advance and serves every linear estimate.

Zero-variance MCMC~\cite{dellaportas2012control,mira2013zero} and control
functionals~\cite{oates2017functionals} act on the observable. They use
target information to correct it after sampling, so they can also be
applied to a coupled run. Section~\ref{sec:experiments} measures this
combination for fitted zero-variance control variates.

A different approach is to couple the dynamics. Applied to the product of $K$
copies of the target, the diffusion framework of Ma, Chen and
Fox~\cite{ma2015recipe} can preserve that joint law. With the skew-symmetric curl matrix set to zero,
a constant symmetric matrix sets the noise covariance and multiplies the drift.
If it is zero-sum, both drift and noise vanish on averaging the chains, so
the ensemble mean stays where it started (Appendix~\ref{app:marginal}).
We keep the ordinary drift and ask only that each chain have its usual law.
In ensemble Langevin methods a state-dependent empirical covariance enters
the update and requires its own correction
\cite{garbuno2020interacting,nusken2019note,garbuno2020affine}.
Column-centering changes only the noise passed to each ordinary ULA update;
it adds no drift interaction. Quasi-Monte Carlo and orthogonal designs change
the time stream or the sample geometry instead~\cite{liu2023langevin,choromanski2019unifying}.

\section{Marginal invariance and path laws}\label{sec:marginal}

Coupling the rows introduces dependence at every step of the simulation.
We need to check what happens to an entire trajectory, since a time
average depends on much more than the distribution at a single step.

\begin{proposition}[Marginal invariance]\label{prop:marginal}
Let $C$ be admissible. Suppose each chain starts from the ordinary ULA
initial law and, at each step, the $K\times d$ array of innovations is Gaussian
with mean zero and covariance $C\otimes I_d$, drawn independently of the past.
Then the full path of every individual chain has the ordinary ULA law.

\end{proposition}
\emph{Proof sketch.} Let $\mathcal F_t$ be the history of all $K$ chains
through step $t$. Given $\mathcal F_t$, the next innovation of chain $i$
is standard Gaussian: the array is independent of $\mathcal F_t$ and
$C_{ii}=1$. Its update uses only this innovation and $X_t^{(i)}$.
Thus for any measurable set $E$, with $P_{\rm ULA}$ the one-chain kernel,
\begin{equation}\label{eq:conditional-kernel}
 \Pr(X_{t+1}^{(i)}\in E\mid\mathcal F_t)=P_{\rm ULA}(X_t^{(i)},E).
\end{equation}
Conditioning successively gives the ordinary ULA probability for any
finite path event in chain $i$
(Appendix~\ref{app:marginal}).

\begin{remark}[Standard errors and diagnostics]\label{rem:diagnostics}
Chains in a coupled ensemble are dependent. Their spread alone therefore
does not give a valid standard error, nor can they be treated as independent
runs for diagnostics such as $\widehat R$
\cite{gelman1992inference,vehtari2021rank}. Run $G\ge2$ independent ensembles
for standard errors. Convergence diagnostics must also account for dependence
within an ensemble~\cite{margossian2024nested}. This costs $G$ times as much
work as one ensemble; all $GK$ chains contribute to the estimate.

\end{remark}

\section{Exact covariance on quadratic targets}
\label{sec:quadratic}

Consider two estimates from the same Gaussian regression run. For a
coefficient mean we might use the last draw from each chain, while for
a prediction we might average over the second half of the run. Although
the estimates weight the draws differently, both depend on the early
innovations. To choose a coupling for both estimates, or for their
difference, we need their covariance as well as their variances.

Let $U(x)=\tfrac12(x-\mu)^\top H(x-\mu)$, with $H$ symmetric, and write
$A=I-\eta H$. Averaging \eqref{eq:ula} gives
\[
 \bar X_{t+1}-\mu=A(\bar X_t-\mu)+\sqrt{2\eta}\,\bar\xi_t,\qquad
 \bar\xi_t=K^{-1}\sum_i\xi_t^{(i)},\qquad
 \operatorname{Cov}(\bar\xi_t)=\alpha(C)I_d.
\]
The starts may be random, independently of the innovations, with
$\operatorname{Cov}(X_0^{(i)},X_0^{(j)})=B_{ij}\Sigma_0$. The admissible
correlation matrix $B$ plays the role of $C$ for the starts: $B=I_K$ for
independent starts and $B=\mathbf1\mathbf1^\top$ for a common random start.
We collect our estimates in a panel with entries

\begin{equation}
    \widehat\theta_\ell
    \;=\;\sum_{t=0}^T w_{\ell t}\,a_\ell^\top \bar X_t,
    \qquad \ell=1,\ldots,m.
\end{equation}
The weights $w_{\ell t}$ select the retained steps, and $a_\ell$ selects
the coefficient or prediction. For the opening examples, use $w_{\ell t}=\mathbf1\{t=T\}$ or,
for even $T$, $(2/T)\mathbf1\{t>T/2\}$.
The dependence on the initial state is
through $W_\ell=\sum_t w_{\ell t}A^t$. The draw at $s$ first appears at $s+1$, with weight
$G_{\ell s}=\sum_{t=s+1}^T w_{\ell t}A^{t-1-s}$.

\begin{theorem}[Covariance of a panel of ensemble-mean estimates]
\label{thm:panel}
For any admissible cross-chain covariances $B$ (initial states) and $C$
(innovations),
\begin{equation}
    \operatorname{Cov}(\widehat\theta_1,\ldots,\widehat\theta_m)
    \;=\;
    \alpha(B)\,M_0+\alpha(C)\,M_\xi,
\end{equation}
where the panel sensitivity matrices
$(M_0)_{\ell r}=a_\ell^\top W_\ell\Sigma_0 W_r^\top a_r$ and
$(M_\xi)_{\ell r}=2\eta\sum_{s=0}^{T-1} a_\ell^\top G_{\ell s}G_{rs}^\top a_r$
are positive semidefinite and depend on the panel design but not on $B$ or
$C$. We call $\alpha(B)M_0$ the start contribution and $\alpha(C)M_\xi$
the innovation contribution. The coupling enters the second only through
$\alpha(C)$. When $M_\xi\ne0$, the admissible $C$ minimizing the covariance
in positive-semidefinite order are precisely those with $C\mathbf1=0$.
Each makes the innovation contribution zero for the whole panel.

\end{theorem}
\emph{Proof sketch.} Averaging a step and then unrolling the recursion
writes each entry of the panel as a sum of independent contributions.
There is one from the start, whose covariance before propagation is
$\alpha(B)\Sigma_0$. Each subsequent step supplies one averaged draw
with covariance $\alpha(C)I_d$. The matrices $W_\ell$ and $G_{\ell s}$
tell us where these contributions appear in the different entries. When
we sum their covariances, the same factor $\alpha(C)$ comes out of every
innovation term (Appendix~\ref{app:identities}). If $M_\xi=0$, the panel is insensitive to
the draws and every coupling gives the same covariance.

\begin{corollary}[Random starts]\label{cor:random-start}
For the last-step estimate $a^\top\bar X_T$,
\begin{equation}\label{eq:single-probe}
 \operatorname{Var}(a^\top\bar X_T)=
 \alpha(B)\,a^\top A^T\Sigma_0(A^T)^\top a+
 2\eta\,\alpha(C)\sum_{s=0}^{T-1}\|(A^s)^\top a\|^2 .
\end{equation}
Here $A^T$ denotes the $T$th matrix power. Under zero-sum innovations only
the start contribution remains. Independent random starts give it factor
$1/K$. Deterministic starts remove it, as do random starts whose deviations
sum to zero across chains ($\alpha(B)=0$).
\end{corollary}
Appendix~\ref{app:exp-observable} checks these predictions for
intermediate correlations (Table~\ref{tab:diagnostic-correlation}),
weighted time averages and random starts (Table~\ref{tab:init}).
Appendix~\ref{app:sgld} treats minibatch noise separately.

\section{Higher-degree observables and Hermite expansion}
\label{sec:hermite}

For two opposite draws $z$ and $-z$, we know their average before
we see either one. If we square the draws first, we have two copies
of the same random variable. How much this matters for a general
observable depends on its Hermite expansion under the Gaussian law.

\begin{theorem}[Variance by polynomial degree]
\label{thm:hermite}
Let $K\ge 2$ and let $(X^{(1)},\ldots,X^{(K)})$ be jointly Gaussian in
$\mathbb R^d$ with common mean $m$ and cross-chain covariance
\begin{equation}
    \operatorname{Cov}(X^{(i)},X^{(j)})=C_{ij}\Sigma,
\end{equation}
where $C$ is admissible, and let $f$ be square-integrable under the common
marginal law.  Decompose the centered observable
$f(m+\Sigma^{1/2}Z)-\mathbb E f(m+\Sigma^{1/2}Z)$, $Z\sim N(0,I_d)$, into
orthogonal polynomial-degree components,
\begin{equation}
 f(m+\Sigma^{1/2}Z)-\mathbb E f(m+\Sigma^{1/2}Z)
 =
 \sum_{r\ge 1} f_r(Z),\qquad v_r=\mathbb E f_r(Z)^2,
\end{equation}
where $f_r$ is the degree-$r$ Hermite component.
Then
\begin{equation}
 \operatorname{Var}\!\left(\frac1K\sum_{i=1}^K f(X^{(i)})\right)
 =
 \sum_{r\ge 1} v_r\,
 \frac{\mathbf 1^\top C^{\circ r}\mathbf 1}{K^2},
\end{equation}
where $C^{\circ r}$ is the entrywise $r$th power of $C$. For the centered
covariance $C_\star$,
\begin{equation}
 \frac{\mathbf 1^\top C_\star^{\circ r}\mathbf 1}{K^2}
 =
 \frac1K\left(1+\frac{(-1)^r}{(K-1)^{r-1}}\right).
\end{equation}
\end{theorem}
The linear term ($r=1$) vanishes. For $K>2$, every other odd degree has a
factor below $1/K$, and every even degree a factor above it, largest at
$r=2$ where it is $1/(K-1)$. The departure decays as $(K-1)^{1-r}$.
At $K=2$, pairing and centering coincide: odd degrees vanish and even
degrees have factor one.

\emph{Proof sketch.} Consider a pair of chains with correlation $c$.
Mehler's formula says that a Hermite component of degree $r$ has
covariance $c^r v_r$ across the pair. Components of different degree
have zero covariance. Thus the variance of the ensemble average is
obtained by summing powers of the entries of $C$, one power for each
degree in $f$. This also works for a singular covariance
(Appendix~\ref{app:hermite}). For ULA, the Gaussian assumption holds on a
quadratic target from a common deterministic start or Gaussian starts with
$B=C$: the state cross-covariance is $C_{ij}\Sigma_t$. If starts and innovations
have different correlations, the linear term is still given by
Theorem~\ref{thm:panel}, while the higher degrees depend on both sources.

There is a simple break-even point for an observable with just a linear
and a quadratic term. Centering removes the linear contribution and
increases the quadratic factor from $1/K$ to $1/(K-1)$. It therefore
helps precisely when $v_1>v_2/(K-1)$. What happens if we are interested
in the quadratic term itself?

\begin{proposition}[Centered quadratic observables]\label{prop:quad-observable}
Under the Gaussian assumptions of Theorem~\ref{thm:hermite}, let
$q(x)=(x-m)^\top Q(x-m)-\operatorname{tr}(Q\Sigma)$, with $Q$ symmetric
and $\operatorname{Var}q(X^{(1)})>0$. The variance of its ensemble average relative to one
chain is $\|C\|_F^2/K^2$. This gives the factors in
Table~\ref{tab:factors}. Among Gaussian zero-sum designs $C_\star$ uniquely
minimizes this penalty.
\end{proposition}
At $K=8$, $C_\star$ raises this variance by about $14\%$ relative to iid
(a factor of $8/7$), while pairing doubles it.
Appendix Figure~\ref{fig:hermite} and Table~\ref{tab:diagnostic-quadratic-observable}
show the crossover as the quadratic contribution grows.

\section{Non-asymptotic analysis for nonquadratic targets}
\label{sec:nonquadratic}

On a nonquadratic target we can still cancel the Gaussian noise exactly.
The difficulty comes one step later, when each chain evaluates the gradient
at its new position. To recover the gradient descent path we would need
the average of these gradients to equal the gradient at the average
position. If the Hessian is $\rho$-Lipschitz, Lemma~\ref{lem:first-order}
(Appendix~\ref{app:nonquadratic}) bounds the norm of the difference by
$\rho(2K)^{-1}\sum_i\|X_t^{(i)}-\bar X_t\|^2$. Thus even at a fixed
degree of curvature, how far the chains have spread matters.

\begin{theorem}[Non-asymptotic variance ratio]\label{thm:variance-ratio}
Let $U$ be $m$-strongly convex and $L$-smooth with $\rho$-Lipschitz Hessian
in operator norm. Start all $K\ge2$ chains at the same deterministic point
and fix an admissible zero-sum $C$. Write $\operatorname{Var}_C$ for
variance under innovation covariance $C\otimes I_d$ and
$\operatorname{Var}_{\mathrm{iid}}$ for $C=I_K$. For $a\ne0$, $0<\eta\le1/(2L)$ and
$T\ge\lceil\ln(3/2)/(\eta L)\rceil$,
\begin{equation}\label{eq:curvature}
 \frac{\operatorname{Var}_{C}(a^\top\bar X_T)}
      {\operatorname{Var}_{\mathrm{iid}}(a^\top\bar X_T)}
 \le 15.6\,\lambda_{\max}(C)\,\Gamma,\qquad
 \Gamma=\frac{\rho^2dL}{m^4}.
\end{equation}
\end{theorem}

Appendix~\ref{app:nonquadratic} gives the proof. The lower bound on $T$
allows the iid variance in the denominator to build up.
For $C_\star$, $\lambda_{\max}=K/(K-1)\le2$, with equality at $K=2$.
Thus $\Gamma\le1/32$ guarantees a variance reduction for every $a\ne0$:
even with two chains the bound is $15.6\cdot2/32=0.975$. By
Proposition~\ref{prop:marginal}, the bias is the same as for iid, so the
mean-squared error is smaller as well.

When $m,L,\rho$ all grow in proportion to sample size $n$, the quantity
$\Gamma$ scales as $d/n$. At $\rho=0$ the bound gives back the exact
quadratic result. Away from zero it can be loose: on constructed targets
that satisfy the assumptions, the measured ratios were $10^{3.6}$ to
$10^{6.8}$ times smaller (Table~\ref{tab:variance-ratio-check}). The logistic
posteriors below have an unpenalized intercept, so no global strong-convexity
constant $m>0$ exists (Remark~\ref{rem:vr-mse}). Their evidence is empirical.

\section{Empirical evaluation}
\label{sec:experiments}

\begin{figure}[ht]
\centering
\includegraphics[width=.41\linewidth]{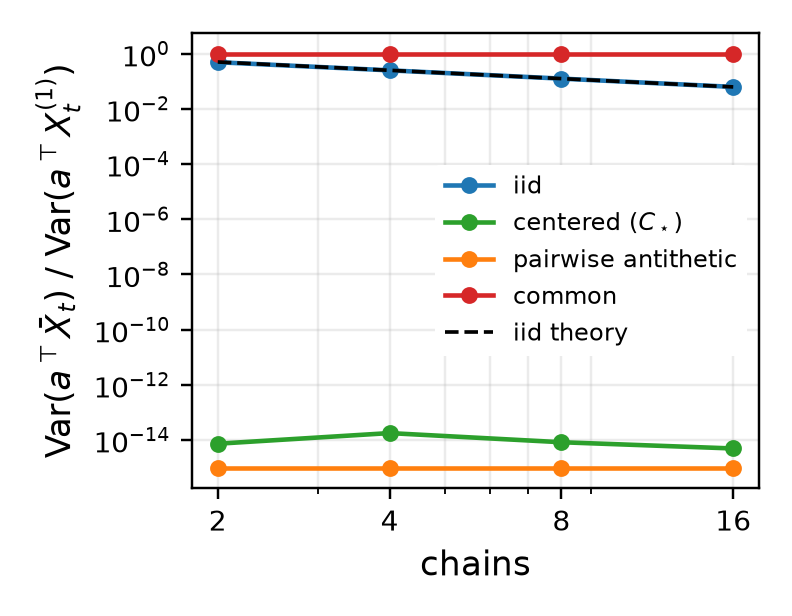}\quad
\includegraphics[width=.41\linewidth]{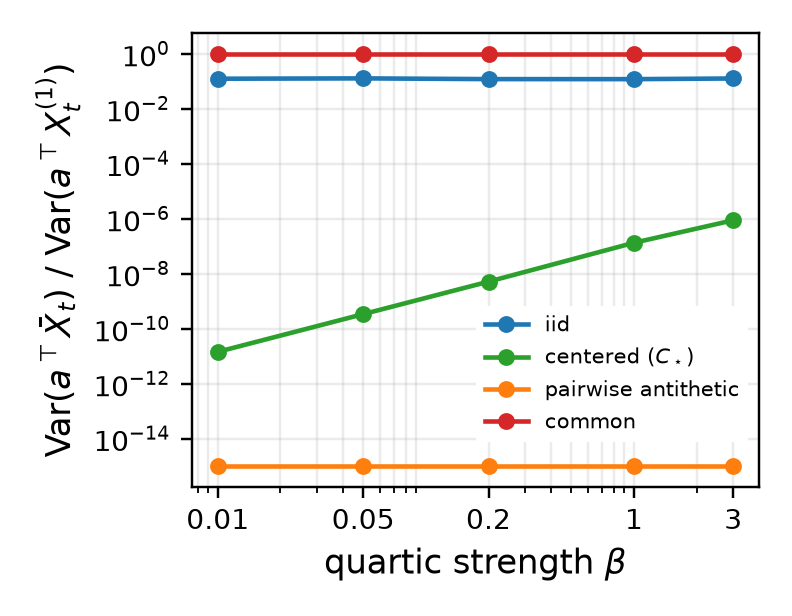}\\[-2pt]
\makebox[.94\linewidth]{\hfill WDBC\hfill\hfill Spambase\hfill\hfill Adult\hfill}\\[-2pt]
\includegraphics[width=.97\linewidth,trim=0 0 0 222bp,clip]{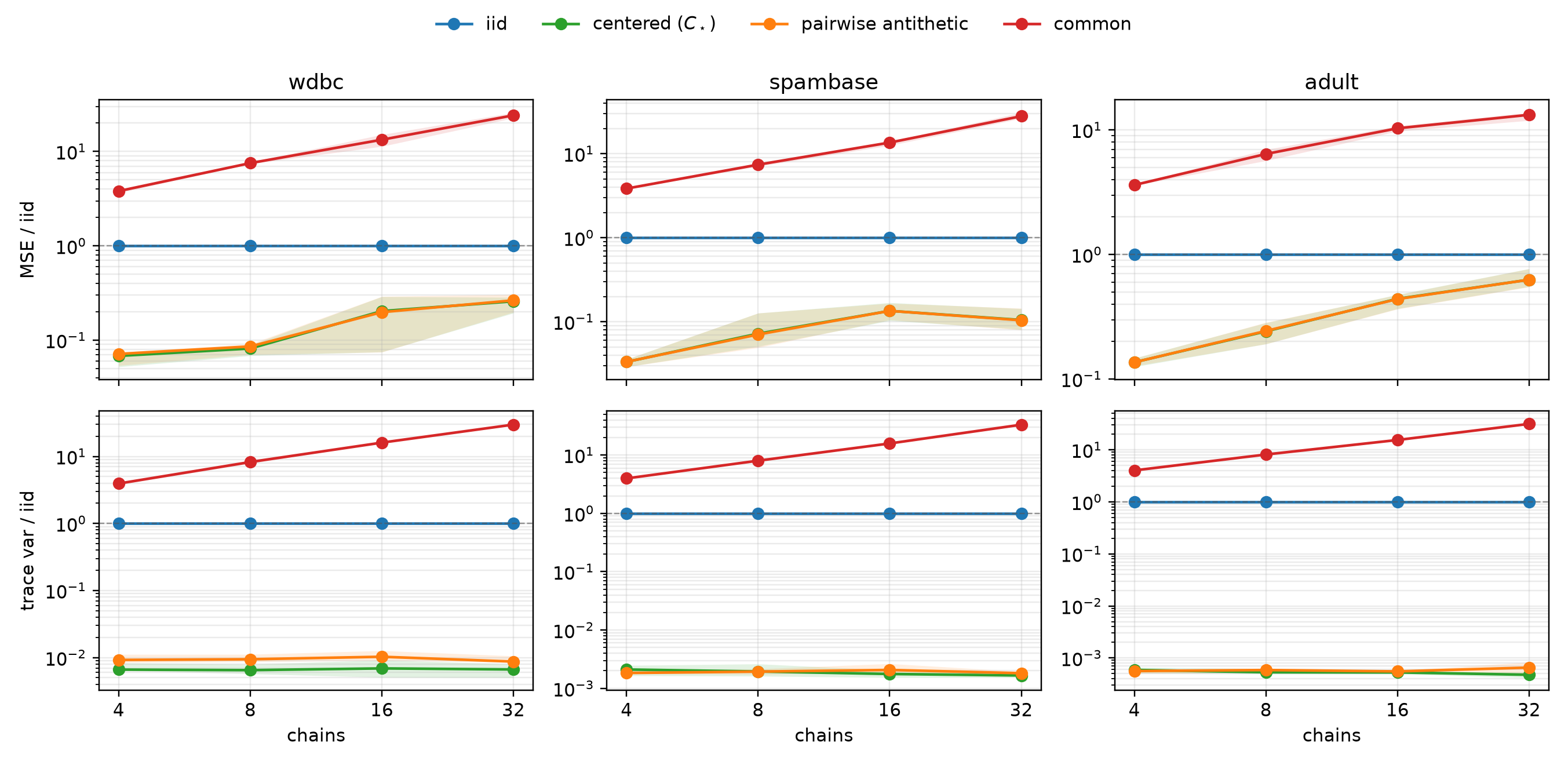}
\caption{Exact cancellation and residual variance. Top left: variance
relative to one Gaussian chain follows $1/K$ for iid (dashed), as
\eqref{eq:single-probe} predicts, and is at roundoff for both zero-sum
designs. Top right: a quartic perturbation leaves residual variance under
$C_\star$; pairing stays at the floor because the zero start and symmetric
target keep pairs mirrored. Bottom: posterior-mean trace variance relative
to iid at each $K$ on the real posteriors; centering and pairing are close.
Blue is iid, red common, orange pairwise and green $C_\star$.
Bands in the bottom panels are empirical interquartile ranges.}
\label{fig:variance-ratio}
\end{figure}

On the tempered WDBC, Spambase and Adult logistic regression
posteriors~\cite{uci_wdbc,uci_spambase,uci_adult}, eight centered ULA chains
gave lower posterior-mean trace variance than 128 independent chains.
Against eight independent chains, the reduction was two to three orders
of magnitude (Table~\ref{tab:campaign-real}). The gain persisted at every
matched chain count tested, from 4 to 32 (Figure~\ref{fig:variance-ratio}).

We raised each likelihood to the power $1/n$, with $n$ the training size,
and used two Gaussian priors per dataset. Each ensemble produces one
vector of coefficient means. We measure its variance across independent
ensembles, summed over coordinates: the \emph{trace variance}.
Coupled-to-iid ratios below one favor coupling. The main campaign followed
a pre-registered protocol. Appendix~\ref{app:experiments} gives its
replicate counts and the protocols and full results for all studies below.

\begin{table}[ht]
\centering\small
\caption{Coupling reduces variance more than reference MSE.
For posterior-mean estimates, ratios below one favor coupling.
The matched-$K$ entries use $K=8$ and are medians of six ratios
across two priors and three replicates.
The eight-versus-128 column reports the larger prior-specific median,
with a fresh-seed repeat in parentheses. Reference MSE averages squared
distances to a longer iid reference; brackets give interquartile ranges
of the matched MSE ratios. Values are rounded.}
\label{tab:campaign-real}
\resizebox{\linewidth}{!}{\begin{tabular}{lcccc}
\toprule
 & \multicolumn{2}{c}{Variance, $K=8$} & Variance, 8 vs.\ 128 & Reference MSE, $K=8$ \\
\cmidrule(lr){2-3}
Dataset & $C_\star$/iid & Pairwise/iid & $C_\star(8)$/iid(128) & $C_\star$/iid \\
\midrule
WDBC & 0.006 & 0.009 & 0.12 (0.13) & 0.082 [0.068, 0.089] \\
Spambase & 0.002 & 0.002 & 0.040 (0.045) & 0.072 [0.051, 0.13] \\
Adult & $5.3\times10^{-4}$ & $5.9\times10^{-4}$ & 0.009 (0.009) & 0.24 [0.19, 0.28] \\
\bottomrule
\end{tabular}
}
\end{table}

\paragraph{Is centering worth the extra computation?}
On a single GPU, the variance reduction outweighed the cost of centering. We
multiplied trace variance by sampling time per ensemble, in seconds,
including the time spent centering on the device. Eight centered chains were
then compared with the best iid chain count tested, up to 128. The median
centered-to-iid ratio was below 2\% in every dataset--prior setting,
including runs from dispersed starts
(Table~\ref{tab:campaign-wallclock-frontier}). To separate the time cost
from the variance reduction, we also timed both methods at $K=8$. In these
real-data runs, the median overhead was never more than about 14\% on one
GPU (Table~\ref{tab:campaign-wallclock}). For chains running on three PCIe
A100 GPUs in one node, we had to communicate at every centering step. These
steps were about 35\% slower than independent steps
(Appendix~\ref{app:exp-additional}, ``Distributed step times'').

\paragraph{How much does accuracy improve?}
On Adult at $K=8$, centering reduced trace variance by a factor of about
1,900, but MSE relative to the fixed reference by only about four
(Table~\ref{tab:campaign-real}). With the reference held fixed, the
remaining contribution to MSE is the squared difference between it and the
estimator's mean. This difference is unaffected by coupling, since the mean
is preserved; it reflects sampler bias as well as error in the reference.
Little is gained in reference MSE by reducing variance once that squared
difference is the dominant term (Figure~\ref{fig:campaign-real-k}). A simple
diagnostic would be to run the reference longer: is the squared change in
the reference larger than the variance over independent ensembles? If so,
more Monte Carlo precision alone may have little visible effect on reference
MSE. This is only a heuristic, since movement of the reference does not
determine its true error.

For the separate comparison in Table~\ref{tab:gold-endtask}, we used a long
MALA reference from dispersed starts and matched the samplers' gradient
budgets. The observed squared errors were again smaller with centered ULA
than with iid ULA. These errors were smaller for full-rank variational
inference and expectation propagation than for centered ULA; under our
conservative, pre-specified rule, however, the sampler-versus-fit rankings
remained unresolved. For every fresh replicate, the estimated stationary
bias of centered ULA was within the reference-error radius.

\paragraph{When should we use centering or pairing?}
With eight chains, ordinary antithetic pairing gave much of the variance
reduction obtained by centering (Table~\ref{tab:campaign-real}). Pairing had
lower medians under MALA, with Student-$t$ priors, and for the full
hierarchical vector. We use centering by default when there is no known
symmetry to exploit. The construction applies for any $K\ge2$, and its
worst-case variance penalty is no greater than pairing's for the averages in
Corollary~\ref{cor:noise-history}. This comparison requires the same
square-integrable function of each chain's Gaussian innovations, with all
chains sharing a deterministic starting point. Relative to iid, the variance
is bounded by a factor of $K/(K-1)$ for centering and 2 for pairing at even
$K$.

For odd $K$, conventional pairing leaves one independent draw. In our
experiments, that draw dominated the measured variance. Centering avoided
this source of noise. Pairing the remaining chains while centering a block
of three also worked well (Tables~\ref{tab:real-odd-k}
and~\ref{tab:oddk-zerosum-class}).

\paragraph{Where does the gain survive?}
Centering was effective in SGLD \cite{welling2011bayesian} even though each
chain had its own minibatches. The large variance reductions persisted at
different batch sizes and from different starts
(Table~\ref{tab:minibatch-sgld-real}). We also obtained reductions in ULA
with a fixed diagonal preconditioner (Table~\ref{tab:preconditioned-real}).

A MALA rejection \cite{roberts1996exponential} discards a Gaussian draw that
would otherwise help balance the draws in accepted proposals. We still
obtained a three- to eightfold variance reduction at the calibrated step
sizes. Much of this gain was lost when we increased the step size. The
leading-order diagnostic $ps$ matched the six dataset--prior median ratios
to within about a third, without fitting constants. Here $p$ denotes the
rejection rate; $s$ is the mean squared innovation norm on rejected steps
divided by the mean over all steps (Appendix~\ref{app:mala-law} and
Tables~\ref{tab:mala-real} and~\ref{tab:mala-sensitivity}). For validity, we
require each chain's proposal noise and acceptance uniform to be independent
with their usual laws, conditional on the ensemble past. This preserves that
chain's ordinary MALA kernel (Appendix~\ref{app:marginal}).

In HMC, we coupled only the momentum refresh
\cite{horowitz1991guided,piponi2020hmc}. There was usually a reduction when
chains shared fixed or jittered trajectory lengths. We found little gain
with chain-specific lengths or NUTS \cite{hoffman2014nuts}
(Tables~\ref{tab:hmc-real} and~\ref{tab:nuts-real}). Tuning ChEES
\cite{hoffman2021chees} on coupled chains failed the pre-specified stability
check in two dataset--prior settings. We then learned a schedule on iid
chains, kept it fixed during the coupled runs, and obtained a variance
reduction. We cannot infer from this that tuning ChEES on coupled chains is
stable (Table~\ref{tab:chees-closedloop}).

For Superconductivity's Gaussian regression posterior, the observed
fixed-start cancellation was at numerical precision
(Table~\ref{tab:bayes-linreg-exact}). MNIST and Gisette have non-Gaussian
logistic posteriors; here variance was far below iid, including for
Gisette's 5,001 parameters (Table~\ref{tab:highdim-real}). We tested a
tempered Gaussian-process hyperparameter posterior with clipped gradients as
well. It retained a reduction, as did logistic models with Student-$t$ or
hierarchical priors, though the Student-$t$ gain was much smaller
(Tables~\ref{tab:gp-hyper} and~\ref{tab:nonlogconcave-real}).

\paragraph{Which observables benefit?}
A more precise estimate of a mean does not imply a more precise estimate of
a credible interval. The bound above concerns averages of single-chain
statistics. Posterior variances and interval widths, by contrast, are
calculated after pooling draws from different chains and are not covered by
that bound. For these estimates, median variance increases across settings
reached 45\% with centering and were larger with pairing. Mean predictive
probabilities still benefited (Table~\ref{tab:uncertainty-observables}).

In the bimodal Gaussian-mixture experiment, centering roughly halved the
variance of the estimated occupancy fraction in the setting where chains
crossed frequently between modes. We found no gain at the higher barriers.
Proposition~\ref{prop:marginal} rules out a change to any chain's escape
law. Pairing was particularly effective for the equal-weight mixture: the
chains were mirrored, making the occupancy estimates essentially constant.
With unequal mixture weights, this cancellation no longer occurred
(Table~\ref{tab:bimodal}).

We fitted zero-variance (ZV) control-variate corrections on a separate iid
run, holding the fitted corrections fixed across couplings
\cite{mira2013zero}. Combining them with centering did not invariably reduce
variance further. After a linear correction, there was little to be gained
from centering on WDBC and Spambase, and variance could increase. Additional
gains depended on the target and observable. The largest additional gain was
for the hierarchical model's training-set predictive probability. We do not
interpret the linear-corrected results for Adult's posterior mean: refitting
the linear correction changed their classification under both couplings
(Table~\ref{tab:zv-grid}).

\FloatBarrier

\section{Limitations and conclusion}\label{sec:limitations}

Even under affine drift, a random start may leave variance in a linear
ensemble-mean estimate: exact cancellation removes the contribution of the
innovations. Under the Gaussian cross-covariance assumptions of
Proposition~\ref{prop:quad-observable}, the ensemble average of a quadratic
form in the mean-subtracted states has variance ratios relative to one chain
of $1/(K-1)$ for centering and $1/K$ for iid. A precise mean estimate
therefore need not give a precise estimate of credible-interval width. The
curvature and horizon conditions of Theorem~\ref{thm:variance-ratio}
restrict the extension to non-affine drift. In particular, the theorem does
not apply to our logistic examples with unpenalized intercepts.

The experiments also give reasons to prefer pairing in some settings. Its
medians were lower at even $K$ on the $t$-prior cells, for the full
hierarchical vector, and for MALA. Our multimodal evidence comes from a
single target, on which the occupancy improvement from centering vanished as
barriers rose. When tuning ChEES on coupled chains, we found two cells that
failed the stability check. We could not rank estimates in the attempted
Poisson regression benchmark because references obtained from dispersed
starts disagreed too much, so we omit the comparison.

For our comparisons we used three measures
(Table~\ref{tab:campaign-real-full}): trace variance, reference MSE, and
test log loss. We used effective sample size only to check the reference
runs. Kernel Stein discrepancy \cite{liu2016ksd} was not measured.
Distributed timings were obtained on three GPUs within one node.

We can also average the chains unequally. In Appendix~\ref{app:identities},
we ask when the Gaussian draws can have zero weighted sum. For nonnegative
weights summing to one, this is possible if and only if no weight exceeds
one half. Any larger weight prevents the other draws from balancing its
contribution. We have not resolved the analogous problem for Metropolis
decisions. Section~\ref{sec:experiments} gives conditions under which each
chain retains its MALA kernel, but an acceptance and a rejection in
different chains can break the cancellation. It remains to find out whether
coupling the acceptance decisions can recover precision without changing
these kernels.

Our recommendation is to compare couplings on independent ensembles, using
the observable of interest and including synchronization in the measured
cost. Centering leaves the law of a ULA trajectory unchanged. The gain
therefore comes from how trajectories depend on one another, and may be
undone by nonlinear terms or by accept/reject decisions.

\begin{ack}
I thank my postdoctoral supervisor, Chandrajit Bajaj of the Computational
Visualization Center (CVC), for his guidance, and the CVC Lab for its support. I acknowledge the Texas
Advanced Computing Center (TACC) at The University of Texas at Austin for
providing the Lonestar6 computing resources used in this work.
\end{ack}

\bibliographystyle{plain}
\bibliography{references}

\clearpage
\appendix

\section{Marginal-invariance proof}
\label{app:marginal}

\begin{proof}[Proof of Proposition~\ref{prop:marginal}]
Fix a chain $i$ and condition on the history $\mathcal F_t$ of the
whole ensemble just before we draw $\Xi_t$. The new array is independent
of that history. The update of chain $i$ reads only its own row of the
array, whose law is $N(0,I_d)$ since $C_{ii}=1$. Thus for any measurable
set $E$, \eqref{eq:ula} gives
\begin{equation}
 \Pr(X_{t+1}^{(i)}\in E\mid\mathcal F_t)
 =P_{\rm ULA}(X_t^{(i)},E),
\end{equation}
where $P_{\rm ULA}$ is the ordinary ULA kernel. To pass from this
one-step statement to the trajectory claim, take measurable sets
$E_0,\ldots,E_T$ and let $H_{T-1}$ be the event that chain $i$ visits
$E_0,\ldots,E_{T-1}$ in order. Since $H_{T-1}$ is measurable with
respect to $\mathcal F_{T-1}$,
\[
 \Pr(H_{T-1},X_T^{(i)}\in E_T)
 =\mathbb E[\mathbf1_{H_{T-1}}P_{\rm ULA}(X_{T-1}^{(i)},E_T)].
\]
Iterating the same conditioning step integrates the initial law against
$T$ successive copies of $P_{\rm ULA}$. This is the ordinary ULA
probability of each cylinder event, which determines the finite path law.
In particular a time average and a first hitting time computed on
the same finite path both have their ordinary single-chain laws.

\end{proof}

\emph{Consistency.} Let $\pi_\eta$ be the invariant law of ordinary ULA at
step size $\eta$. Whenever its ergodic theorem holds for $f$, the time
average in each chain converges almost surely to $\pi_\eta(f)$. For fixed
$K$, these $K$ probability-one events hold simultaneously, so averaging
the chains preserves the limit. Taking the time horizon to infinity first
and then $\eta$ to zero gives $\pi(f)$ whenever ordinary ULA has
$\pi_\eta(f)\to\pi(f)$. At finite time the expectation of any additive
estimate agrees with matched independent chains by linearity.

\emph{Why product stationarity gives a different sampler.} For the stacked
state $x$, set $\widetilde U(x)=\sum_i U(x^{(i)})$ and
$D=C\otimes I_d$. In the framework of Ma, Chen and Fox~\cite{ma2015recipe},
constant symmetric mobility gives the diffusion
\begin{equation}
 dx=-D\nabla\widetilde U(x)\,dt+\sqrt{2D}\,dW .
\end{equation}
The product density proportional to $\exp(-\widetilde U)$ has zero
probability current, even for singular $D$. Stationarity need not give
convergence from a chosen start. For a linear aggregate $Ax$, zero
diffusion means $ADA^\top=0$. Since this matrix equals
$(AD^{1/2})(AD^{1/2})^\top$, it forces $AD^{1/2}=0$ and $AD=0$.
The drift of the aggregate also vanishes. In particular, $\alpha(C)=0$
implies $C\mathbf1=0$, and the ensemble mean stays where it started.
Convergence to the product law is impossible from an initial conserved-mode
law different from its product-target law. With $C_\star$, chain $i$ has
drift
$-\tfrac{K}{K-1}(\nabla U(x^{(i)})-K^{-1}\sum_j\nabla U(x^{(j)}))$.
It already depends on the other chains. Adding a skew-symmetric curl matrix can move the
mean while preserving the product target in continuous time. Euler
discretization still needs a separate discretization analysis.

\begin{remark}
The conditional-kernel proof applies to any chain-local transition whose
whole random seed has the prescribed law given $\mathcal F_t$. For MALA it
suffices that each proposal innovation and its acceptance uniform have their
usual product law within a chain. Acceptance then preserves the coordinate
kernel, although the accepted increments need no longer sum to zero.
\end{remark}

\section{Initial-condition, weighted-average, and panel identities}
\label{app:identities}

There are two sources of randomness when the chains have random initial
states. Centering the draws removes the innovations from the ensemble mean,
but the initial state can still contribute variance. We calculate that
contribution first for a terminal projection. We then allow averaging over
time and compute covariances between averages to obtain the panel identity.
As in Section~\ref{sec:quadratic}, throughout this section we take
$U(x)=\tfrac12(x-\mu)^\top H(x-\mu)$ and $A=I-\eta H$.

\begin{theorem}[Initial-condition decomposition]
\label{thm:init}
Let the initial states in the terminal-projection setting of
Theorem~\ref{thm:panel} be random, independent of the innovations, and
satisfy $\operatorname{Cov}(X_0^{(i)},X_0^{(j)})=B_{ij}\Sigma_0$.
Suppose $B$ is a positive semidefinite cross-chain matrix with unit diagonal.
Then, for every $a\in\mathbb R^d$,
\[
 \operatorname{Var}\!\left(a^\top\bar X_t\right)
 =
 \frac{\mathbf 1^\top B\mathbf 1}{K^2}\,
 a^\top A^t\Sigma_0(A^t)^\top a
 +
 \frac{\mathbf 1^\top C\mathbf 1}{K^2}\,
 2\eta\sum_{s=0}^{t-1} a^\top A^s(A^s)^\top a .
\]
\end{theorem}

\begin{proof}
The average of the initial states has covariance
$K^{-2}\sum_{i,j}B_{ij}\Sigma_0=(\mathbf 1^\top B\mathbf 1/K^2)\Sigma_0$.
Its contribution at time $t$ is obtained by multiplication by $A^t$ on the
left and $(A^t)^\top$ on the right. Theorem~\ref{thm:panel} supplies the
innovation variance. Independence of the starts and innovations makes their
cross-covariance zero, so we add the two variances after projection onto $a$.
\end{proof}

\begin{remark}
For iid overdispersed initial states, $\mathbf 1^\top B\mathbf 1/K^2=1/K$.
The corresponding factors for a common random start and for zero-sum
randomized starts are $1$ and $0$. Thus centering the innovations leaves the
first term in Theorem~\ref{thm:init}, propagated by the deterministic
matrix $A^t$. When the initial draws are Gaussian, centering can be applied
at time zero as well: the zero-sum construction preserves their marginal
laws and removes this remaining contribution.
\end{remark}

\begin{theorem}[Weighted estimator identity]
\label{thm:weighted}
Let $\widehat\theta_w(a)=\sum_{t=0}^T w_t\,a^\top\bar X_t$, where the
time weights $w_0,\ldots,w_T$ are deterministic. Assume the conditions of
Theorem~\ref{thm:init}, and put $W=\sum_{t=0}^T w_t A^t$ and
$G_s=\sum_{t=s+1}^T w_t A^{t-1-s}$. Then
\[
 \operatorname{Var}\!\bigl(\widehat\theta_w(a)\bigr)
 =
 \frac{\mathbf 1^\top B\mathbf 1}{K^2}\,
 a^\top W\Sigma_0 W^\top a
 +
 \frac{\mathbf 1^\top C\mathbf 1}{K^2}\,
 2\eta\sum_{s=0}^{T-1} a^\top G_sG_s^\top a .
\]
\end{theorem}

\begin{proof}
This is the one-entry case of the panel identity in Theorem~\ref{thm:panel}.
Use $a$ as the projection and $w_0,\ldots,w_T$ as its time weights.
The two sensitivity matrices are then scalars:
$M_0=a^\top W\Sigma_0W^\top a$ and
$M_\xi=2\eta\sum_{s=0}^{T-1}a^\top G_sG_s^\top a$.
Substituting them in that identity gives the result.
\end{proof}

\begin{corollary}[Simultaneous covariance minimality across estimator panels]
\label{cor:loewner}
For a finite panel $\widehat\theta_\ell=\sum_t w_{\ell t}a_\ell^\top\bar X_t$,
$\ell=1,\ldots,m$, condition on the initial ensemble $X_0$. For some
positive semidefinite matrix $M$ independent of $C$, we have
$\operatorname{Cov}((\widehat\theta_1,\ldots,\widehat\theta_m)\mid X_0)
=\frac{\mathbf 1^\top C\mathbf 1}{K^2}\,M$;
a zero-sum covariance therefore removes the innovation contribution
simultaneously for any such collection of estimators. If $M\ne0$,
covariance minimality requires $C\mathbf 1=0$.
\end{corollary}

\begin{proof}
Unroll the affine recursion in both estimators and take their covariance
conditional on $X_0$. After factoring out $\mathbf 1^\top C\mathbf 1/K^2$,
the $(\ell,r)$ entry is
$M_{\ell r}=2\eta\sum_s a_\ell^\top G_{\ell s}G_{rs}^\top a_r$.
In particular, $M$ is a covariance matrix and is positive semidefinite. The remaining
minimization is scalar: $\mathbf 1^\top C\mathbf 1=\|C^{1/2}\mathbf 1\|^2$
cannot be negative, and it vanishes if and only if $C\mathbf 1=0$.
These choices give zero covariance. For $M\ne0$, any other choice gives a
nonzero positive semidefinite contribution.
\end{proof}

\begin{remark}[Beyond coordinate-isotropic couplings]
Our definition of admissibility uses $C\otimes I_d$. The same question can
be asked for a centered Gaussian coupling with arbitrary cross-chain
covariance blocks $S_{ij}\in\mathbb R^{d\times d}$. To preserve the
marginals, the diagonal blocks must be $S_{ii}=I_d$, and the joint covariance
must be positive semidefinite. In this case
$\operatorname{Cov}(\bar\xi_s)=K^{-2}\sum_{i,j}S_{ij}$. Thus all panels
lose their innovation contribution precisely when the blocks sum to zero.
For the coordinate-isotropic choice $S_{ij}=C_{ij}I_d$, the covariance of
$\bar\xi_s$ is again $(\mathbf 1^\top C\mathbf 1/K^2)I_d$.
\end{remark}

\begin{proof}[Proof of Theorem~\ref{thm:panel}]
Solving the recursion for chain $i$ gives
$X_t^{(i)}-\mu=A^t(X_0^{(i)}-\mu)+
\sqrt{2\eta}\sum_{s=0}^{t-1}A^{t-1-s}\xi_s^{(i)}$.
Substitution in $\widehat\theta_\ell=\sum_t w_{\ell t}a_\ell^\top\bar X_t$
gives a contribution $a_\ell^\top W_\ell\bar X_0$ from initialization,
up to a deterministic term. For estimators $\ell$ and $r$, its covariance
is $(\mathbf 1^\top B\mathbf 1/K^2)a_\ell^\top W_\ell\Sigma_0W_r^\top a_r$.
Now interchange the time sums in the innovation part. The draw $\bar\xi_s$
has coefficient $\sqrt{2\eta}a_\ell^\top G_{\ell s}$ in estimator $\ell$
and covariance $(\mathbf 1^\top C\mathbf 1/K^2)I_d$. Cross terms between
two different innovation times vanish by independence, as do those between
an innovation and an initial state. The remaining terms give the displayed
$M_0$ and $M_\xi$. To check positive semidefiniteness, note that $M_0$ has
the form $V\Sigma_0V^\top$ and $M_\xi$ is a sum of Gram matrices.

For the minimization over $C$, the argument in Corollary~\ref{cor:loewner}
applies again. The factor $\mathbf 1^\top C\mathbf 1=\|C^{1/2}\mathbf 1\|^2$
is minimized exactly when $C\mathbf 1=0$. Provided $M_\xi\ne0$, these and
only these couplings minimize the panel covariance. If $M_\xi=0$, there is
no innovation contribution to minimize and every admissible $C$ gives the
same covariance.
\end{proof}

\emph{Unequal chain weights.} For fixed chain weights $w_i$ in place of
$1/K$, the factor multiplying the innovation covariance becomes $w^\top Cw$.
Over admissible correlation matrices,
\begin{equation}\label{eq:weighted-optimum}
  \min_C w^\top Cw = \bigl(\max\{0,2\|w\|_\infty-\|w\|_1\}\bigr)^2 .
\end{equation}
A correlation matrix $C$ is the Gram matrix of unit vectors $u_i$, and
hence $w^\top Cw=\|\sum_i w_i u_i\|^2$. Let $L=\|w\|_\infty$ be the
longest length and $R=\|w\|_1-L$ the sum of the other lengths. The triangle
inequality bounds the norm below by $\max\{0,L-R\}$. If $L>R$, attain
equality by pointing every other weighted vector against a longest one.
Otherwise, segments with lengths $|w_i|$ can be arranged as a closed polygon
in the plane, possibly degenerate. The signs of the weights can be absorbed into the unit vectors
$u_i$. Their Gram matrix attains \eqref{eq:weighted-optimum} and is admissible.

\section{Balanced finite-sum SGLD}
\label{app:sgld}

Suppose we have centered the Gaussian draws in SGLD. The average step
can still fluctuate because each chain estimates its gradient from a
minibatch. For a finite sum of quadratics we can identify the term
that remains when the batches together cover the entire dataset. Write the $j$th summand as
$u_j(x)=\tfrac12 x^\top H_jx-c_j^\top x$, put $H=\sum_j H_j$ and
$c=\sum_j c_j$, and use
$g_B(x)=(n/b)\sum_{j\in B}(H_jx-c_j)$ for a batch of size $b$.
Suppose $n=Kb$. At each step select a fresh partition of the data into
$K$ batches, independently of the past and of the innovations. Require
each chain's batch to have the ordinary uniform size-$b$ marginal law.
For example, randomly permute the observations and divide the permutation
into $K$ consecutive blocks. Give one block to each chain and update by
\[
    X_{t+1}^{(i)}=X_t^{(i)}-\eta g_{B_t^{(i)}}(X_t^{(i)})
        +\sqrt{2\eta}\,\xi_t^{(i)}.
\]

\begin{theorem}[Balanced finite-sum cancellation]
\label{thm:finite-sum}
Let $H_B=(n/b)\sum_{j\in B}H_j$ and
$c_B=(n/b)\sum_{j\in B}c_j$. Each partition uses every observation
once, so averaging a step across the chains gives
\[
    \bar X_{t+1}
    =
    \bar X_t-\eta(H\bar X_t-c)-\eta R_t+\sqrt{2\eta}\,\bar\xi_t,
    \qquad
    R_t=\frac1K\sum_{i=1}^K(H_{B_t^{(i)}}-H)(X_t^{(i)}-\bar X_t).
\]
For common component Hessians ($H_j=H/n$) we have $R_t=0$ identically.
With common deterministic initialization and zero-sum innovations, the
ensemble mean then follows a deterministic full-gradient path,
$\bar X_{t+1}=\bar X_t-\eta(H\bar X_t-c)$.  More generally with $R_t=0$, the
linear variance identity of Theorem~\ref{thm:weighted} holds with
$A=I-\eta H$.
\end{theorem}

\begin{proof}
Balanced batches give $K^{-1}\sum_i H_{B_t^{(i)}}=H$ and
$K^{-1}\sum_i c_{B_t^{(i)}}=c$, so
$K^{-1}\sum_i g_{B_t^{(i)}}(X_t^{(i)})
=H\bar X_t-c+K^{-1}\sum_i(H_{B_t^{(i)}}-H)(X_t^{(i)}-\bar X_t)$.
When every component has the same Hessian, all the terms in $R_t$
are zero even though the states at which we evaluated the gradients
differ. The remaining recursion is affine, so the variance calculation
in Theorem~\ref{thm:weighted} applies.
\end{proof}

\section{Gaussian polynomial and quadratic-observable details}
\label{app:hermite}

Two Gaussian draws $Z$ and $-Z$ have exactly the same square, so
averaging their squares buys us nothing. For centered, jointly Gaussian draws with correlation $c$, the squares have correlation $c^2$. This explains why the
value of $c$ for an individual pair matters even after we have made
the sum of the draws zero. The calculation for a general centered
quadratic form follows the same argument.

\begin{proof}[Proof of Proposition~\ref{prop:quad-observable}]
For jointly Gaussian vectors, Isserlis' identity gives
$\operatorname{Cov}(q_i,q_j)
=2\,\operatorname{tr}(Q\operatorname{Cov}(X^{(i)},X^{(j)})Q\operatorname{Cov}(X^{(j)},X^{(i)}))
=C_{ij}^2\operatorname{Var}(q_1)$.  Averaging over $i,j$ yields the ratio
$\|C\|_F^2/K^2$.  The displayed special cases follow from $\|C\|_F^2$ for
each coupling.  For the minimization claim, the zero-sum condition gives
$C\mathbf 1=0$ and $\operatorname{tr}(C)=K$; if $\lambda_1,\ldots,\lambda_r$
are the nonzero eigenvalues then $r\le K-1$, $\lambda_j\ge 0$,
$\sum_j\lambda_j=K$, so $\|C\|_F^2=\sum_j\lambda_j^2\ge K^2/r\ge K^2/(K-1)$.
Equality requires $r=K-1$ and all nonzero eigenvalues equal to $K/(K-1)$,
with nullspace spanned by $\mathbf 1$, which gives $C=C_\star$.
\end{proof}

\begin{corollary}[Minimum quadratic-degree penalty]
\label{cor:min-second-chaos}
Among admissible zero-sum covariances, $C_\star$ uniquely minimizes
$\|C\|_F^2$ at $K^2/(K-1)$, and is therefore the unique minimizer of the
quadratic-degree ratio $\|C\|_F^2/K^2$ in
Proposition~\ref{prop:quad-observable}.
\end{corollary}

\begin{proof}
Immediate from the inequality in the proof of
Proposition~\ref{prop:quad-observable}.
\end{proof}

\begin{proposition}[A non-Gaussian paired-magnitude benchmark]
\label{prop:paired-magnitude-benchmark}
Let $K=2m$.  There exists a joint law for $Y_1,\ldots,Y_m$, each with
marginal distribution $\chi^2_1$, attaining the Wang--Wang value below.
Draw iid signs $\sigma_j$ uniformly from $\{-1,1\}$, independently of
the squared magnitudes. The $j$th pair consists of
\[
(\xi_{2j-1},\xi_{2j})=(\sigma_j\sqrt{Y_j},-\sigma_j\sqrt{Y_j}),
\qquad j=1,\ldots,m.
\]
Then each coordinate has marginal law $N(0,1)$ and
$\sum_{i=1}^{2m}\xi_i=0$ pathwise.  Let $G$ be the $\chi^2_1$ quantile and
define
\[
H_m(x)=G(1-x)+(m-1)G((m-1)x),\qquad x\in[0,1/m].
\]
Let $c_m$ be the smallest value in $[0,1/m]$ satisfying
\[
\int_{c_m}^{1/m} H_m(t)\,dt
\ge
\left(\frac1m-c_m\right)H_m(c_m).
\]
The Wang--Wang monotone-density convex-minimization formula
\cite{wangwang2011completemixability} gives, for convex $\phi$ with finite
expectation,
\[
\inf_{\substack{\text{couplings}\\Y_j\sim\chi^2_1}}
\mathbb E\,\phi\!\left(\sum_{j=1}^m Y_j\right)
=
m\int_0^{c_m}\phi(H_m(x))\,dx
+(1-mc_m)\phi(H_m(c_m)).
\]
The quantity we want to average, $\xi_i^2-1$, takes the same value
on both members of a pair. Averaging over $2m$ draws thus gives the
same variance as averaging $Y_j-1$ over the $m$ pairs,
\[
Q_{2m}
=
\frac{\operatorname{Var}\!\left((2m)^{-1}
\sum_{i=1}^{2m}(\xi_i^2-1)\right)}
{\operatorname{Var}(\xi_1^2-1)}
=
\frac{\operatorname{Var}(\sum_{j=1}^mY_j)}{2m^2}.
\]
To minimize this ratio we can use $\phi(s)=(s-m)^2$, since the $m$
squared magnitudes each have mean one. This gives the first column of
variance ratios below.
\[
\begin{array}{c|cccc}
K & \text{paired magnitude} & \text{iid} & C_\star & \text{pairwise}\\
\hline
6  & 0.098987 & 0.166667 & 0.200000 & 0.333333\\
8  & 0.039929 & 0.125000 & 0.142857 & 0.250000\\
10 & 0.017612 & 0.100000 & 0.111111 & 0.200000
\end{array}
\]
At $K=8$, for instance, dependence among the four squared magnitudes
brings the ratio down from $0.25$ for independent pairs to about
$0.04$. The formula specifies the attainable minimum for this one
observable. Equation~\eqref{eq:construct} gives an explicit jointly
Gaussian construction for every $K\ge2$.
\end{proposition}

\begin{proof}
Apply the Wang--Wang formula to $W=-Y$, where $Y\sim\chi^2_1$.  The law of
$W$ has increasing density on $(-\infty,0]$, with quantile
$G_W(u)=-G(1-u)$.  Their increasing-density expression
$G_W(x)+(m-1)G_W(1-(m-1)x)$ maps to $-H_m(x)$, and the sign reversal gives the
displayed inequality and objective for the sum of the $Y_j$ variables.  The
endpoint integrals at $x=0$ are improper; for
$\phi(s)=(s-m)^2$ they are finite because $\chi^2_1$ has finite second moment.
The independent sign pairing then gives the stated Gaussian marginals and
pathwise zero-sum identity.
\end{proof}

\begin{proof}[Proof of Theorem~\ref{thm:hermite} (full)]
The nonsingular case follows by whitening
$X^{(i)}=m+\Sigma^{1/2}Z^{(i)}$, so
$\operatorname{Cov}(Z^{(i)},Z^{(j)})=C_{ij}I_d$.  The Mehler--Hermite
covariance identity~\cite{janson1997gaussian} gives
$\mathbb E[f_r(Z^{(i)})f_s(Z^{(j)})]=\mathbf 1_{\{r=s\}}C_{ij}^r v_r$.  For a finite truncation $F_R=\sum_{r=1}^R f_r$, we can sum these
covariances over all $K^2$ pairs and divide by $K^2$. This gives the
claimed expression through order $R$. To let $R$ increase, observe that
the $L^2$ error of the chain average is bounded by the average of the
$K$ marginal $L^2$ errors, each of which tends to zero. Thus the
variance of the truncated average converges to the desired variance.
For $C_\star$, summing order $r$ along one row of the correlation matrix
gives $1+(K-1)(-1/(K-1))^r$. All rows give the same answer, leaving
the factor $K^{-1}(1+(-1)^r/(K-1)^{r-1})$. If $\Sigma$ is singular
the Gaussian vector lives on an affine subspace, where the whitening
argument applies on the range of $\Sigma$.

\end{proof}

\begin{corollary}[When zero-sum beats iid]
\label{cor:hermite-winning}
For an observable with only linear and quadratic Gaussian-polynomial energy,
$C_\star$ has smaller variance than iid averaging iff $v_1>v_2/(K-1)$.  More
generally, any admissible zero-sum $C$ deletes the linear term and has
quadratic coefficient $\alpha_2(C)=\|C\|_F^2/K^2$; it beats iid averaging iff
$v_1>(K\alpha_2(C)-1)v_2$.
\end{corollary}

\begin{corollary}[Degree law for a history of Gaussian draws]\label{cor:noise-history}
Suppose every chain starts from the same deterministic point. At each of
$T$ steps draw a fresh Gaussian array with covariance $C\otimes I_d$
independently of the past. Each chain computes the same square-integrable
scalar function $F$ of its own $T$ innovations. This includes a chain-local
trajectory statistic with a common, independently fitted correction held
fixed. Relative to iid chains, the variance factor at innovation-history
Hermite order $r\ge1$ is
\begin{equation}
 R_r(C_\star)=1+(-1)^r(K-1)^{1-r},\qquad R_r(C_{\rm pair})=1+(-1)^r
 \quad\text{(fixed pairs, even $K$)} .
\end{equation}
Thus the full variance is at most $K/(K-1)$ times iid with $C_\star$
and at most twice iid with pairs. Both bounds are attained at $r=2$.
\end{corollary}
\begin{proof}
Stack the $dT$ Gaussian innovations of each chain. The cross-covariance of
two stacks is $C_{ij}I_{dT}$. Apply Theorem~\ref{thm:hermite} to $F$ on
these Gaussian inputs, conditional on any independently fitted map. At
order $r$, division by the iid factor $1/K$ gives $K^{-1}\sum_{ij}C_{ij}^r$.
For pairwise coupling each row contributes $1+(-1)^r$. For $C_\star$ each
contributes $1+(K-1)(-1/(K-1))^r$. The full variance ratio, when the iid
variance is positive, averages these factors with nonnegative weights.
Summing coordinate variances gives the same upper bound for trace variance.
\end{proof}
It is the randomness fed to a chain that is Gaussian here. The resulting
states can have quite a different distribution. In particular, $r$ is a
degree in the Gaussian innovations. It need not agree with the polynomial
degree used for a ZV fit. If we fit the correction on the ensemble
itself, each corrected estimate also depends on draws in the other
chains, and the corollary no longer applies.

\section{Nonquadratic remainder: proof and perturbative law}
\label{app:nonquadratic}

When the drift is affine, taking one step commutes with averaging the
chains. With a varying Hessian the two orders of operation give different
answers. We bound the discrepancy over a full trajectory, then compare
the resulting variance with a lower bound for independent chains along
the same projection. The perturbative calculation at
the end of the section identifies the first term that survives a small
departure from a quadratic.

\begin{lemma}[First-order cancellation]
\label{lem:first-order}
Assume $\nabla^2 U$ is $\rho$-Lipschitz.  Under zero-sum coupling,
\[
 \bar X_{t+1}
 =
 \bar X_t-\eta \nabla U(\bar X_t)-\eta R_t,
 \qquad
 \|R_t\|\le \frac{\rho}{2K}\sum_{i=1}^K\|X_t^{(i)}-\bar X_t\|^2 .
\]
\end{lemma}

\begin{proof}[Proof of Lemma~\ref{lem:first-order}]
Average the $K$ updates, use $\sum_i\xi_t^{(i)}=0$, and Taylor expand
$\nabla U(X_t^{(i)})$ around $\bar X_t$.  The linear Hessian terms cancel
because $\sum_i(X_t^{(i)}-\bar X_t)=0$, and the Lipschitz-Hessian remainder
gives the bound.
\end{proof}

\begin{theorem}[Nonasymptotic curvature-residual bound]
\label{thm:nonlinear-bound}
Assume $U$ is $m$-strongly convex, $L$-smooth, with $\rho$-Lipschitz
Hessian, and let $\eta\le 1/L$, $q=1-\eta m$.  Run zero-sum coupled ULA from
a common deterministic point $x_0$ and let $y_{t+1}=y_t-\eta\nabla U(y_t)$
with $y_0=x_0$ be the noiseless gradient-descent reference.  Then, with
$S_t=K^{-1}\sum_i\|X_t^{(i)}-\bar X_t\|^2$ the realized ensemble spread,
\[
    \|\bar X_t-y_t\|
    \le
    \frac{\eta\rho}{2}\sum_{s=0}^{t-1}q^{t-1-s}S_s,
    \qquad
    \operatorname{Var}\!\bigl(a^\top\bar X_t\bigr)
    \le
    \tfrac14\eta^2\rho^2\|a\|^2\,
    \mathbb E\!\bigl[(\textstyle\sum_s q^{t-1-s}S_s)^2\bigr].
\]
\end{theorem}

\begin{proof}[Proof of Theorem~\ref{thm:nonlinear-bound}]
Let $T(x)=x-\eta\nabla U(x)$.  Since $U$ is $m$-strongly convex and
$L$-smooth and $\eta\le 1/L$, the Hessian integral formula
$T(x)-T(y)=\bigl[I-\eta\int_0^1\nabla^2U(y+\theta(x-y))\,d\theta\bigr](x-y)$
shows that $T$ is a $q$-contraction with $q=1-\eta m$.  Lemma~\ref{lem:first-order}
gives $\bar X_{t+1}=T(\bar X_t)-\eta R_t$ while $y_{t+1}=T(y_t)$, so
$\|\bar X_{t+1}-y_{t+1}\|\le q\|\bar X_t-y_t\|+\frac{\eta\rho}{2}S_t$.
Unrolling from the common start proves the pathwise inequality.  The
variance bound follows from $\operatorname{Var}(Z)\le \mathbb E[(Z-c)^2]$
with $Z=a^\top\bar X_t$ and $c=a^\top y_t$.
\end{proof}

\begin{lemma}[Gaussian Poincar\'e, possibly singular covariance]
\label{lem:poincare}
Let $G\sim N(0,\Sigma)$ on $\mathbb R^N$ with $\Sigma\succeq0$, and let
$f\in C^1(\mathbb R^N)$ with $\mathbb E[\nabla f(G)^\top\Sigma\nabla
f(G)]<\infty$.  Then
$\operatorname{Var} f(G)\le\mathbb E[\nabla f(G)^\top\Sigma\nabla f(G)]$.
Moreover the right side does not depend on the choice of $C^1$
extension off the support: two extensions agreeing on the support
$\operatorname{range}(\Sigma)$ have gradient difference orthogonal to
$\operatorname{range}(\Sigma^{1/2})$ there.
\end{lemma}

\begin{proof}
Write $G=\Sigma^{1/2}W$ with $W$ standard normal and apply the
classical Gaussian Poincar\'e inequality to $w\mapsto f(\Sigma^{1/2}w)$,
whose gradient is $\Sigma^{1/2}\nabla f(\Sigma^{1/2}w)$.
Extension-independence: at support points, $(\nabla f_1-\nabla
f_2)\perp\operatorname{range}(\Sigma)$, so
$\Sigma^{1/2}(\nabla f_1-\nabla f_2)=0$.
\end{proof}

\paragraph{Component bounds for Theorem~\ref{thm:variance-ratio}.}
Let $U$ be $m$-strongly convex and $L$-smooth with $\rho$-Lipschitz
Hessian (operator norm), $\eta\le 1/(2L)$, $K\ge2$ chains from a common
deterministic start, and let $C$ be admissible and zero-sum
($C\succeq0$, $C_{ii}=1$, $C\mathbf 1=0$), $\lambda_C:=\lambda_{\max}(C)$;
for $C_\star$, $\lambda_C=K/(K-1)$.  Then for every probe $a\ne0$:
\begin{itemize}
\item[(i)] for every $T\ge1$,
$\displaystyle
\operatorname{Var}_{\mathrm{zs}}\bigl(a^\top\bar X_T\bigr)
\le\frac{128}{81}\,\frac{\lambda_C\,\rho^2 d}{K\,m^4}\,\|a\|^2$;
\item[(ii)] under iid noise, only $L$-smoothness, and any start
independent of the innovations, for every
$T\ge T_0:=\lceil\ln(3/2)/(\eta L)\rceil$,
$\displaystyle
\operatorname{Var}_{\mathrm{iid}}\bigl(a^\top\bar X_T\bigr)
\ge\frac{\ln(3/2)}{4}\,\frac{\|a\|^2}{K\,L}$;
\item[(iii)] hence, with $\Gamma:=\rho^2dL/m^4$ (scale-invariant), for
every $T\ge T_0$,
\[
\frac{\operatorname{Var}_{\mathrm{zs}}\bigl(a^\top\bar X_T\bigr)}
     {\operatorname{Var}_{\mathrm{iid}}\bigl(a^\top\bar X_T\bigr)}
\;\le\;
\frac{512}{81\ln(3/2)}\,\lambda_C\,\Gamma
\;\le\;15.6\,\lambda_C\,\Gamma,
\]
strictly below one for $C_\star$ and $a\ne0$ whenever
$\Gamma<(K-1)/(16K)$---and also at $\Gamma\le1/32$ for every $K\ge2$,
since $512/(81\ln(3/2))<16$. The bound tends to zero as $\rho\to0$,
recovering the exact quadratic cancellation of Theorem~\ref{thm:panel}.
\end{itemize}

\begin{proof}
Fix the convention $J^{(i)}_{s+1\to T}
:=\prod_{u=T-1}^{s+1}\bigl(I-\eta\nabla^2U(X^{(i)}_u)\bigr)$
(time-ordered, later factors left; $n:=T-1-s$ factors; empty product
$=I$), and $q=1-\eta m$; each factor has operator norm at most $q$
since $\eta\le1/(2L)\le2/(m+L)$.

\emph{Step 1 (spread moment, zero-sum).}  Let $D^{(i)}_t=X^{(i)}_t-\bar
X_t$ and $\Sigma_t=\sum_i\|D^{(i)}_t\|^2$.  Zero-sum noise has
$\bar\xi_t=0$ a.s., so with the integral Hessians
$\hat H^{(i)}_t=\int_0^1\nabla^2U(\bar X_t+\theta D^{(i)}_t)\,d\theta$,
\[
D^{(i)}_{t+1}=(I-\eta\hat H^{(i)}_t)D^{(i)}_t+r_t+\sqrt{2\eta}\,\xi^{(i)}_t,
\qquad r_t=\frac\eta K\sum_j\hat H^{(j)}_tD^{(j)}_t .
\]
Since $\sum_i(I-\eta\hat H^{(i)}_t)D^{(i)}_t=-Kr_t$, pointwise
$\sum_i\|(I-\eta\hat H^{(i)}_t)D^{(i)}_t+r_t\|^2
=\sum_i\|(I-\eta\hat H^{(i)}_t)D^{(i)}_t\|^2-K\|r_t\|^2
\le q^2\Sigma_t$: the mean-residual cross term helps.  Fresh
innovations are mean zero and independent of $\mathcal F_t$ with
$\mathbb E\|\xi^{(i)}_t\|^2=d$, so
$\mathbb E\Sigma_{t+1}\le q^2\,\mathbb E\Sigma_t+2\eta Kd$ and, from
$\Sigma_0=0$ and $1-q^2=\eta m(2-\eta m)\ge\tfrac32\eta m$,
$\sup_t\mathbb E\Sigma_t\le\tfrac43Kd/m$.

\emph{Step 2 (Poincar\'e with zero-sum centering).}  Parameterize the
noise by the unit-marginal innovations $\xi^{(i)}_s$, whose stacked
covariance is $\Sigma=I_T\otimes(C\otimes I_d)$; the factor
$\sqrt{2\eta}$ enters through the dynamics.  Chains interact only
through the noise coupling, so $a^\top\bar X_T$ is a $C^1$, globally
Lipschitz function of the stacked innovations whose only nonzero
partials are
$\nabla_{\xi^{(i)}_s}(a^\top\bar X_T)
=\tfrac{\sqrt{2\eta}}K\,(J^{(i)}_{s+1\to T})^\top a=:g_i$.
Lemma~\ref{lem:poincare} with this $\Sigma$ tensorizes over the
independent time blocks.  For each block, since
$(C\otimes I)(\mathbf 1\otimes w)=0$ for every $w$, the quadratic form
satisfies, pointwise and for ANY (possibly random) common vector $w$,
\[
\textstyle\sum_{ij}C_{ij}\,g_i^\top g_j
=\sum_{ij}C_{ij}(g_i-\bar g)^\top(g_j-\bar g)
\le\lambda_C\sum_i\|g_i-\bar g\|^2
\le\lambda_C\sum_i\|g_i-w\|^2,
\]
the last step because $\bar g$ minimizes the scatter.  The common
Jacobian mode---the entire $O(q^{2n})$ bulk---is annihilated exactly;
for $C_\star$ the first inequality is an equality.

\emph{Step 3 (telescope against the mean path).}  Take
$w=\tfrac{\sqrt{2\eta}}K(J^{\mathrm{ref}}_{s+1\to T})^\top a$ with
$J^{\mathrm{ref}}$ the product of $(I-\eta\nabla^2U(\bar X_u))$ along
the realized ensemble-mean path (each factor of norm $\le q$).  The
product-difference identity and the $\rho$-Lipschitz Hessian give,
pathwise,
\[
\bigl\|J^{(i)}_{s+1\to T}-J^{\mathrm{ref}}_{s+1\to T}\bigr\|
\le\eta\rho\,q^{\,n-1}\sum_{u=s+1}^{T-1}\|D^{(i)}_u\|,
\]
each of the $n$ summands carrying prefix/suffix norm
$q^{(T-1-u)+(u-1-s)}=q^{\,n-1}$.  Cauchy--Schwarz over the $n$
summands, Step 1, and
$\sum_{n\ge1}n^2x^{n-1}=(1+x)/(1-x)^3$ at $x=q^2$ yield
\[
\begin{aligned}
\operatorname{Var}_{\mathrm{zs}}
&\le\frac{2\eta\lambda_C}{K^2}\,\eta^2\rho^2\|a\|^2
\sum_{n\ge1}n^2q^{2(n-1)}\,\sup_u\mathbb E\Sigma_u\\
&\le\frac{2\eta^3\rho^2\lambda_C\|a\|^2}{K^2}\cdot
\frac{2}{(\tfrac32\eta m)^3}\cdot\frac{4Kd}{3m}
=\frac{128}{81}\,\frac{\lambda_C\rho^2d}{Km^4}\,\|a\|^2 ,
\end{aligned}
\]
the $\eta^3$ cancelling exactly: the bound is uniform in $\eta$ and
$T$.  This proves (i).

\emph{Step 4 (iid lower bound).}  Condition on the starts (any law
independent of the innovations); by the law of total variance it
suffices to bound the conditional variance.  For a Lipschitz function
$f$ of a standard Gaussian vector $W$,
$\operatorname{Var} f\ge\|\mathbb E\nabla f\|^2$
(Stein's identity gives $\operatorname{Cov}(f,u^\top W)=u^\top\mathbb
E\nabla f$; maximize over unit $u$); apply it conditionally to
$f=a^\top\bar X_T$ as a function of the stacked iid innovations, whose
partials are $\tfrac{\sqrt{2\eta}}K(J^{(i)}_{s+1\to T})^\top a$, so
that the conditional bound is
$(2\eta/K^2)\sum_s\sum_i\|\mathbb E[(J^{(i)}_{s+1\to T})^\top a\mid
\text{starts}]\|^2$.  (Under the theorem's standing
$\rho$-Lipschitz-Hessian hypothesis $U\in C^2$ and this display is
classical.  With $\nabla U$ merely Lipschitz, $\nabla^2U$ exists a.e.\ by
Rademacher, and $x\mapsto x-\eta\nabla U(x)$ is bi-Lipschitz for
$\eta L\le1/2$, so preimages of Lebesgue-null sets are null along the
recursion and the a.e.\ chain rule applies to the Lipschitz $f$.)  Keep the lags
$n=T-1-s\in\{0,\dots,\tau-1\}$ with
$\tau=\lceil\ln(3/2)/(\eta L)\rceil\wedge$ the available horizon: the
ordered expansion of $J=\prod(I-\eta H_u)$ around $I$ gives
$\|J-I\|\le(1+\eta L)^n-1$ using only $\|H_u\|\le L$, so pathwise
$a^\top Ja\ge\|a\|^2\bigl(2-(1+\eta L)^n\bigr)\ge\|a\|^2/2$ whenever
$(1+\eta L)^n\le3/2$, which holds for all $n\le\tau-1$ since
$\ln(1+x)\le x$; the bound is pathwise, hence survives the conditional
expectation:
$\|\mathbb E[J^\top a\mid\text{starts}]\|\ge\|a\|/2$ on those lags for
every chain.  Summing $K$ chains and $\tau$ lags,
$\operatorname{Var}_{\mathrm{iid}}(a^\top\bar X_T)
\ge(2\eta/K^2)\cdot K\tau\cdot\|a\|^2/4
\ge\ln(3/2)\,\|a\|^2/(4KL)$ (if $\eta L>\ln(3/2)$ the single lag $n=0$
already gives $2\eta\|a\|^2/(4K)\ge\ln(3/2)\|a\|^2/(4KL)$).  No
convexity, Hessian continuity, or start condition is used beyond
independence of the innovations.  Dividing (i) by (ii) gives (iii), and
$512/(81\ln(3/2))\le15.59$.
\end{proof}
The last inequality has a factor of two to spare: the same display gives
$\ln(3/2)\,\|a\|^2/(2KL)$, which would halve the constant to $7.80$. We keep
$15.6$ because the checks in Table~\ref{tab:variance-ratio-check} used it.

\begin{remark}[Sharpness of the coupling constant; class scope]
\label{rem:vr-lambda}
On the zero-sum subspace, $\operatorname{tr}C=K$ over at most $K-1$
nonzero eigenvalues forces $\lambda_C\ge K/(K-1)$, with equality iff
$C=C_\star$: the centered coupling uniquely minimizes the Poincar\'e
constant of Theorem~\ref{thm:variance-ratio}.  Pairwise antithetic
blocks have $\lambda_C=2$ (the familiar $2/K$ against $1/(K-1)$), and
$\lambda_C\le K$ with equality attainable for even $K$ (balanced
sign-flip common noise), which forfeits the $K$-gain: the guarantee,
for a fixed target and step size, depends on the coupling through
$\lambda_C$.
\end{remark}

\begin{remark}[MSE dominance; regime; conservatism]
\label{rem:vr-mse}
Marginal invariance (Proposition~\ref{prop:marginal}) makes the bias
identical under the two couplings, so on the guarantee class the $C_\star$ ensemble has strictly smaller mean-squared error for
every nonzero terminal projection simultaneously when $\Gamma\le1/32$.  For posteriors whose curvature scales uniformly with the data size
($m,L,\rho=\Theta(n)$, as for strongly log-concave likelihoods with
parameter-uniform third derivatives), $\Gamma=\Theta(d/n)$: the class
$\Gamma\le1/32$ then corresponds to the $n\gtrsim d$ regime.  The
real-data campaign posteriors lie outside the class because the
unpenalized logistic intercept admits no global $m>0$. Restricting
attention to a neighborhood of the MAP or of the reference mean also
gives little purchase. At each of the twelve evaluation points,
pointwise estimates for the implemented posterior put $\Gamma$ at least
$1.3\times10^6$ times above $1/32$, reaching $2.1\times10^{15}$ at the
Adult/$0.3$ reference mean (Table~\ref{tab:certified-gamma}). On the
constructed targets inside the class, the measured ratios were
$10^{3.6}$--$10^{6.8}$ times smaller than the upper bound.
The exponents fitted to $\rho$ lay between $1.9$ and $2.1$ on both
the product and dense-direction cosine families. With these projections
we saw no systematic increase with dimension
(Table~\ref{tab:variance-ratio-check}).

Initialization: (i) uses the common start only through the zero
initial spread; nonzero initial spread adds an explicit start-dependent
term.  The lower bound (ii) holds for any start independent of the
innovations.  State-adaptive couplings are outside the argument (the
Poincar\'e step needs a fixed Gaussian).
\end{remark}

\begin{table}[htbp]
\centering
\small
\caption{For each of the six campaign posteriors, the numerical lower
bound on $\Gamma=\rho^2 dL/m^4$ for a ball containing the MAP or the
archived reference mean is at least $1.3\times10^6$ times the $1/32$
threshold of Theorem~\ref{thm:variance-ratio}.
At each point we used $m_{\mathrm{ub}}=\lambda_{\min}(H)$,
$L_{\mathrm{lb}}=\lambda_{\max}(H)$ and the largest absolute third
directional derivative along a unit Hessian eigenvector for
$\rho_{\mathrm{lb}}$, all computed in float64 without outward rounding.
The least-curved direction loads on the unpenalized intercept, and
sigmoid saturation at the reference mean further lowers
$m_{\mathrm{ub}}$, which enters the denominator to the fourth power.}
\label{tab:certified-gamma}
\resizebox{.9\linewidth}{!}{\begin{tabular}{llrrrrrrr}
\toprule
Dataset & prior & point & $d$ & $m_{\mathrm{ub}}$ & $L_{\mathrm{lb}}$ & $\rho_{\mathrm{lb}}$ & $\Gamma_{\mathrm{lb}}$ & $\Gamma_{\mathrm{lb}}/(1/32)$ \\
\midrule
WDBC & 0.1 & MAP & 31 & 0.0609 & 0.577 & 0.745 & $7.22\times10^{5}$ & $2.31\times10^{7}$ \\
WDBC & 0.1 & gold mean & 31 & 0.00831 & 0.161 & 0.073 & $5.58\times10^{6}$ & $1.79\times10^{8}$ \\
WDBC & 0.3 & MAP & 31 & 0.1 & 1.07 & 1.67 & $9.06\times10^{5}$ & $2.90\times10^{7}$ \\
WDBC & 0.3 & gold mean & 31 & 0.0202 & 0.488 & 0.118 & $1.27\times10^{6}$ & $4.05\times10^{7}$ \\
Spambase & 0.1 & MAP & 58 & 0.0922 & 0.385 & 0.716 & $1.58\times10^{5}$ & $5.07\times10^{6}$ \\
Spambase & 0.1 & gold mean & 58 & 0.0209 & 0.222 & 0.595 & $2.37\times10^{7}$ & $7.58\times10^{8}$ \\
Spambase & 0.3 & MAP & 58 & 0.161 & 0.736 & 0.814 & $4.19\times10^{4}$ & $1.34\times10^{6}$ \\
Spambase & 0.3 & gold mean & 58 & 0.0456 & 0.534 & 0.869 & $5.41\times10^{6}$ & $1.73\times10^{8}$ \\
Adult & 0.1 & MAP & 105 & 0.0888 & 0.739 & 6.35 & $5.04\times10^{7}$ & $1.61\times10^{9}$ \\
Adult & 0.1 & gold mean & 105 & 0.00142 & 0.181 & 0.182 & $1.56\times10^{11}$ & $4.99\times10^{12}$ \\
Adult & 0.3 & MAP & 105 & 0.132 & 1.02 & 9.88 & $3.47\times10^{7}$ & $1.11\times10^{9}$ \\
Adult & 0.3 & gold mean & 105 & 0.000728 & 0.382 & 0.671 & $6.42\times10^{13}$ & $2.06\times10^{15}$ \\
\bottomrule
\end{tabular}
}
\end{table}

\begin{table}[htbp]
\centering
\scriptsize
\caption{Validation of Theorem~\ref{thm:variance-ratio} on constructed
targets inside its class, part i/ii: the $96$ product-family cells
($48+48$ rows side by side) ($U=\tfrac12\sum h_jx_j^2+\rho$-scaled
cosine perturbations, product and dense-direction families; analytic
$(m,L,\rho)$---exact for the product family, Weyl bounds for the dense
family, which inflate the reported bound but never the measured ratio;
$4096$ ensembles
$\times$ three replicates per cell).  Across both parts the proven
bound dominates every
one of the $576$ perturbed replicate runs ($192$ parameter cells
$\times$ three replicates; median [max] shown); the $72$ quadratic
sanity runs measure at the float64 floor ($\sim10^{-31}$).  The slack
range quoted in the text ($10^{3.6}$--$10^{6.8}$) is over individual
replicate runs; the displayed cell-level values span
$4.27\times10^{3}$--$5.53\times10^{6}$.  The dense-direction family is
in Table~\ref{tab:variance-ratio-check-dense}.}
\label{tab:variance-ratio-check}
\resizebox{\linewidth}{!}{\begin{tabular}{lrrrrrrr}
\toprule
family & $\rho$ & $d$ & $K$ & $\eta L$ & measured ratio & proven bound & slack \\
\midrule
product & 0.001 & 2 & 4 & 0.025 & 1.39e-08 [1.60e-08] & 8.35e-05 & 5221.5x \\
product & 0.001 & 2 & 4 & 0.062 & 1.43e-08 [1.47e-08] & 8.35e-05 & 5673.0x \\
product & 0.001 & 2 & 4 & 0.125 & 1.42e-08 [1.44e-08] & 8.35e-05 & 5781.7x \\
product & 0.001 & 2 & 8 & 0.025 & 1.59e-08 [1.66e-08] & 7.16e-05 & 4307.6x \\
product & 0.001 & 2 & 8 & 0.062 & 1.52e-08 [1.59e-08] & 7.16e-05 & 4502.9x \\
product & 0.001 & 2 & 8 & 0.125 & 1.55e-08 [1.68e-08] & 7.16e-05 & 4273.0x \\
product & 0.001 & 4 & 4 & 0.025 & 2.30e-09 [2.42e-09] & 1.67e-04 & 69154.5x \\
product & 0.001 & 4 & 4 & 0.062 & 2.30e-09 [2.40e-09] & 1.67e-04 & 69584.8x \\
product & 0.001 & 4 & 4 & 0.125 & 2.37e-09 [2.48e-09] & 1.67e-04 & 67237.2x \\
product & 0.001 & 4 & 8 & 0.025 & 2.49e-09 [2.50e-09] & 1.43e-04 & 57240.3x \\
product & 0.001 & 4 & 8 & 0.062 & 2.55e-09 [2.58e-09] & 1.43e-04 & 55566.8x \\
product & 0.001 & 4 & 8 & 0.125 & 2.66e-09 [2.72e-09] & 1.43e-04 & 52591.0x \\
product & 0.001 & 8 & 4 & 0.025 & 7.31e-09 [7.32e-09] & 3.34e-04 & 45626.8x \\
product & 0.001 & 8 & 4 & 0.062 & 7.71e-09 [7.98e-09] & 3.34e-04 & 41857.7x \\
product & 0.001 & 8 & 4 & 0.125 & 7.96e-09 [8.01e-09] & 3.34e-04 & 41693.6x \\
product & 0.001 & 8 & 8 & 0.025 & 7.99e-09 [8.03e-09] & 2.86e-04 & 35659.0x \\
product & 0.001 & 8 & 8 & 0.062 & 8.08e-09 [8.59e-09] & 2.86e-04 & 33326.5x \\
product & 0.001 & 8 & 8 & 0.125 & 9.00e-09 [9.00e-09] & 2.86e-04 & 31809.7x \\
product & 0.001 & 16 & 4 & 0.025 & 6.33e-09 [6.76e-09] & 6.68e-04 & 98829.4x \\
product & 0.001 & 16 & 4 & 0.062 & 6.78e-09 [6.97e-09] & 6.68e-04 & 95849.7x \\
product & 0.001 & 16 & 4 & 0.125 & 6.78e-09 [7.11e-09] & 6.68e-04 & 94000.1x \\
product & 0.001 & 16 & 8 & 0.025 & 6.99e-09 [7.19e-09] & 5.73e-04 & 79626.5x \\
product & 0.001 & 16 & 8 & 0.062 & 7.43e-09 [7.57e-09] & 5.73e-04 & 75696.2x \\
product & 0.001 & 16 & 8 & 0.125 & 7.50e-09 [7.72e-09] & 5.73e-04 & 74215.0x \\
product & 0.003 & 2 & 4 & 0.025 & 1.25e-07 [1.28e-07] & 7.58e-04 & 5947.0x \\
product & 0.003 & 2 & 4 & 0.062 & 1.24e-07 [1.28e-07] & 7.58e-04 & 5909.5x \\
product & 0.003 & 2 & 4 & 0.125 & 1.37e-07 [1.43e-07] & 7.58e-04 & 5299.4x \\
product & 0.003 & 2 & 8 & 0.025 & 1.44e-07 [1.46e-07] & 6.50e-04 & 4439.9x \\
product & 0.003 & 2 & 8 & 0.062 & 1.36e-07 [1.49e-07] & 6.50e-04 & 4370.7x \\
product & 0.003 & 2 & 8 & 0.125 & 1.38e-07 [1.47e-07] & 6.50e-04 & 4414.4x \\
product & 0.003 & 4 & 4 & 0.025 & 2.01e-08 [2.05e-08] & 1.52e-03 & 74114.7x \\
product & 0.003 & 4 & 4 & 0.062 & 2.08e-08 [2.15e-08] & 1.52e-03 & 70436.0x \\
product & 0.003 & 4 & 4 & 0.125 & 2.10e-08 [2.30e-08] & 1.52e-03 & 65865.9x \\
product & 0.003 & 4 & 8 & 0.025 & 2.13e-08 [2.19e-08] & 1.30e-03 & 59421.5x \\
product & 0.003 & 4 & 8 & 0.062 & 2.28e-08 [2.35e-08] & 1.30e-03 & 55246.4x \\
product & 0.003 & 4 & 8 & 0.125 & 2.49e-08 [2.56e-08] & 1.30e-03 & 50832.8x \\
product & 0.003 & 8 & 4 & 0.025 & 6.62e-08 [7.47e-08] & 3.03e-03 & 40637.5x \\
product & 0.003 & 8 & 4 & 0.062 & 6.78e-08 [7.12e-08] & 3.03e-03 & 42590.9x \\
product & 0.003 & 8 & 4 & 0.125 & 7.07e-08 [7.21e-08] & 3.03e-03 & 42100.7x \\
product & 0.003 & 8 & 8 & 0.025 & 7.21e-08 [7.29e-08] & 2.60e-03 & 35693.7x \\
product & 0.003 & 8 & 8 & 0.062 & 7.73e-08 [7.88e-08] & 2.60e-03 & 32990.1x \\
product & 0.003 & 8 & 8 & 0.125 & 7.67e-08 [7.77e-08] & 2.60e-03 & 33457.5x \\
product & 0.003 & 16 & 4 & 0.025 & 5.86e-08 [5.93e-08] & 6.07e-03 & 102401.9x \\
product & 0.003 & 16 & 4 & 0.062 & 5.78e-08 [6.14e-08] & 6.07e-03 & 98821.7x \\
product & 0.003 & 16 & 4 & 0.125 & 6.15e-08 [6.19e-08] & 6.07e-03 & 97979.4x \\
product & 0.003 & 16 & 8 & 0.025 & 6.51e-08 [6.71e-08] & 5.20e-03 & 77530.1x \\
product & 0.003 & 16 & 8 & 0.062 & 6.34e-08 [6.43e-08] & 5.20e-03 & 80833.1x \\
product & 0.003 & 16 & 8 & 0.125 & 6.61e-08 [7.17e-08] & 5.20e-03 & 72528.5x \\
\bottomrule
\end{tabular}\hspace{2em}\begin{tabular}{lrrrrrrr}
\toprule
family & $\rho$ & $d$ & $K$ & $\eta L$ & measured ratio & proven bound & slack \\
\midrule
product & 0.01 & 2 & 4 & 0.025 & 1.38e-06 [1.51e-06] & 8.70e-03 & 5762.0x \\
product & 0.01 & 2 & 4 & 0.062 & 1.44e-06 [1.48e-06] & 8.70e-03 & 5894.1x \\
product & 0.01 & 2 & 4 & 0.125 & 1.52e-06 [1.54e-06] & 8.70e-03 & 5659.3x \\
product & 0.01 & 2 & 8 & 0.025 & 1.53e-06 [1.60e-06] & 7.46e-03 & 4662.8x \\
product & 0.01 & 2 & 8 & 0.062 & 1.49e-06 [1.53e-06] & 7.46e-03 & 4887.1x \\
product & 0.01 & 2 & 8 & 0.125 & 1.54e-06 [1.59e-06] & 7.46e-03 & 4695.9x \\
product & 0.01 & 4 & 4 & 0.025 & 2.26e-07 [2.27e-07] & 1.74e-02 & 76801.8x \\
product & 0.01 & 4 & 4 & 0.062 & 2.35e-07 [2.40e-07] & 1.74e-02 & 72479.7x \\
product & 0.01 & 4 & 4 & 0.125 & 2.54e-07 [2.56e-07] & 1.74e-02 & 67961.3x \\
product & 0.01 & 4 & 8 & 0.025 & 2.47e-07 [2.61e-07] & 1.49e-02 & 57117.2x \\
product & 0.01 & 4 & 8 & 0.062 & 2.53e-07 [2.62e-07] & 1.49e-02 & 56956.5x \\
product & 0.01 & 4 & 8 & 0.125 & 2.79e-07 [2.84e-07] & 1.49e-02 & 52508.2x \\
product & 0.01 & 8 & 4 & 0.025 & 7.25e-07 [8.01e-07] & 3.48e-02 & 43421.6x \\
product & 0.01 & 8 & 4 & 0.062 & 7.37e-07 [7.73e-07] & 3.48e-02 & 45002.8x \\
product & 0.01 & 8 & 4 & 0.125 & 7.65e-07 [7.79e-07] & 3.48e-02 & 44677.4x \\
product & 0.01 & 8 & 8 & 0.025 & 8.43e-07 [8.44e-07] & 2.98e-02 & 35327.5x \\
product & 0.01 & 8 & 8 & 0.062 & 8.31e-07 [8.56e-07] & 2.98e-02 & 34853.4x \\
product & 0.01 & 8 & 8 & 0.125 & 8.50e-07 [8.67e-07] & 2.98e-02 & 34397.2x \\
product & 0.01 & 16 & 4 & 0.025 & 6.46e-07 [6.92e-07] & 6.96e-02 & 100510.4x \\
product & 0.01 & 16 & 4 & 0.062 & 6.81e-07 [7.11e-07] & 6.96e-02 & 97863.7x \\
product & 0.01 & 16 & 4 & 0.125 & 6.72e-07 [6.97e-07] & 6.96e-02 & 99877.1x \\
product & 0.01 & 16 & 8 & 0.025 & 7.14e-07 [7.26e-07] & 5.96e-02 & 82165.6x \\
product & 0.01 & 16 & 8 & 0.062 & 7.32e-07 [7.44e-07] & 5.96e-02 & 80196.1x \\
product & 0.01 & 16 & 8 & 0.125 & 7.39e-07 [7.48e-07] & 5.96e-02 & 79793.6x \\
product & 0.03 & 2 & 4 & 0.025 & 1.31e-05 [1.33e-05] & 8.58e-02 & 6440.8x \\
product & 0.03 & 2 & 4 & 0.062 & 1.26e-05 [1.31e-05] & 8.58e-02 & 6534.0x \\
product & 0.03 & 2 & 4 & 0.125 & 1.32e-05 [1.34e-05] & 8.58e-02 & 6388.3x \\
product & 0.03 & 2 & 8 & 0.025 & 1.43e-05 [1.49e-05] & 7.35e-02 & 4939.2x \\
product & 0.03 & 2 & 8 & 0.062 & 1.47e-05 [1.51e-05] & 7.35e-02 & 4858.4x \\
product & 0.03 & 2 & 8 & 0.125 & 1.38e-05 [1.49e-05] & 7.35e-02 & 4946.1x \\
product & 0.03 & 4 & 4 & 0.025 & 2.01e-06 [2.05e-06] & 1.72e-01 & 83816.9x \\
product & 0.03 & 4 & 4 & 0.062 & 2.05e-06 [2.13e-06] & 1.72e-01 & 80467.2x \\
product & 0.03 & 4 & 4 & 0.125 & 2.14e-06 [2.20e-06] & 1.72e-01 & 77847.1x \\
product & 0.03 & 4 & 8 & 0.025 & 2.28e-06 [2.37e-06] & 1.47e-01 & 62131.5x \\
product & 0.03 & 4 & 8 & 0.062 & 2.30e-06 [2.39e-06] & 1.47e-01 & 61593.5x \\
product & 0.03 & 4 & 8 & 0.125 & 2.53e-06 [2.57e-06] & 1.47e-01 & 57308.7x \\
product & 0.03 & 8 & 4 & 0.025 & 6.95e-06 [7.23e-06] & 3.43e-01 & 47437.0x \\
product & 0.03 & 8 & 4 & 0.062 & 7.00e-06 [7.19e-06] & 3.43e-01 & 47749.4x \\
product & 0.03 & 8 & 4 & 0.125 & 7.32e-06 [7.33e-06] & 3.43e-01 & 46800.4x \\
product & 0.03 & 8 & 8 & 0.025 & 7.50e-06 [7.51e-06] & 2.94e-01 & 39163.7x \\
product & 0.03 & 8 & 8 & 0.062 & 7.40e-06 [7.90e-06] & 2.94e-01 & 37214.9x \\
product & 0.03 & 8 & 8 & 0.125 & 7.41e-06 [7.78e-06] & 2.94e-01 & 37800.2x \\
product & 0.03 & 16 & 4 & 0.025 & 5.80e-06 [6.27e-06] & 6.86e-01 & 109502.6x \\
product & 0.03 & 16 & 4 & 0.062 & 6.08e-06 [6.53e-06] & 6.86e-01 & 105151.9x \\
product & 0.03 & 16 & 4 & 0.125 & 6.39e-06 [6.55e-06] & 6.86e-01 & 104784.6x \\
product & 0.03 & 16 & 8 & 0.025 & 6.61e-06 [6.76e-06] & 5.88e-01 & 86985.3x \\
product & 0.03 & 16 & 8 & 0.062 & 6.67e-06 [6.74e-06] & 5.88e-01 & 87294.9x \\
product & 0.03 & 16 & 8 & 0.125 & 6.86e-06 [7.03e-06] & 5.88e-01 & 83682.5x \\
\bottomrule
\end{tabular}
}
\end{table}

\begin{table}[htbp]
\centering
\scriptsize
\caption{Validation of Theorem~\ref{thm:variance-ratio}, part ii/ii:
the $96$ dense-direction cosine-family cells ($48+48$ rows side by
side; Weyl bounds on $(m,L,\rho)$ inflate the reported bound, never
the measured ratio).  Columns, protocol, and the sanity-run summary
as in Table~\ref{tab:variance-ratio-check}.}
\label{tab:variance-ratio-check-dense}
\resizebox{\linewidth}{!}{\begin{tabular}{lrrrrrrr}
\toprule
family & $\rho$ & $d$ & $K$ & $\eta L$ & measured ratio & proven bound & slack \\
\midrule
dense & 0.001 & 2 & 4 & 0.025 & 2.19e-09 [2.32e-09] & 8.35e-05 & 35976.5x \\
dense & 0.001 & 2 & 4 & 0.062 & 2.40e-09 [2.45e-09] & 8.35e-05 & 34080.2x \\
dense & 0.001 & 2 & 4 & 0.125 & 2.27e-09 [2.34e-09] & 8.35e-05 & 35668.9x \\
dense & 0.001 & 2 & 8 & 0.025 & 2.40e-09 [2.70e-09] & 7.16e-05 & 26483.7x \\
dense & 0.001 & 2 & 8 & 0.062 & 2.58e-09 [2.61e-09] & 7.16e-05 & 27402.9x \\
dense & 0.001 & 2 & 8 & 0.125 & 2.55e-09 [2.57e-09] & 7.16e-05 & 27874.4x \\
dense & 0.001 & 4 & 4 & 0.025 & 2.68e-09 [2.83e-09] & 1.67e-04 & 59042.4x \\
dense & 0.001 & 4 & 4 & 0.062 & 2.94e-09 [2.95e-09] & 1.67e-04 & 56642.7x \\
dense & 0.001 & 4 & 4 & 0.125 & 3.03e-09 [3.31e-09] & 1.67e-04 & 50455.0x \\
dense & 0.001 & 4 & 8 & 0.025 & 3.19e-09 [3.24e-09] & 1.43e-04 & 44224.5x \\
dense & 0.001 & 4 & 8 & 0.062 & 3.11e-09 [3.40e-09] & 1.43e-04 & 42165.4x \\
dense & 0.001 & 4 & 8 & 0.125 & 3.45e-09 [3.46e-09] & 1.43e-04 & 41423.3x \\
dense & 0.001 & 8 & 4 & 0.025 & 6.59e-11 [6.96e-11] & 3.34e-04 & 4799456.2x \\
dense & 0.001 & 8 & 4 & 0.062 & 6.27e-11 [6.42e-11] & 3.34e-04 & 5203602.6x \\
dense & 0.001 & 8 & 4 & 0.125 & 7.10e-11 [7.35e-11] & 3.34e-04 & 4545678.8x \\
dense & 0.001 & 8 & 8 & 0.025 & 6.87e-11 [6.95e-11] & 2.86e-04 & 4119121.9x \\
dense & 0.001 & 8 & 8 & 0.062 & 7.08e-11 [7.37e-11] & 2.86e-04 & 3885844.1x \\
dense & 0.001 & 8 & 8 & 0.125 & 7.10e-11 [7.60e-11] & 2.86e-04 & 3766363.1x \\
dense & 0.001 & 16 & 4 & 0.025 & 3.95e-10 [4.30e-10] & 6.68e-04 & 1555493.9x \\
dense & 0.001 & 16 & 4 & 0.062 & 3.95e-10 [4.01e-10] & 6.68e-04 & 1667817.7x \\
dense & 0.001 & 16 & 4 & 0.125 & 4.39e-10 [4.55e-10] & 6.68e-04 & 1469055.1x \\
dense & 0.001 & 16 & 8 & 0.025 & 4.45e-10 [4.50e-10] & 5.73e-04 & 1273705.5x \\
dense & 0.001 & 16 & 8 & 0.062 & 4.53e-10 [4.62e-10] & 5.73e-04 & 1239409.7x \\
dense & 0.001 & 16 & 8 & 0.125 & 4.65e-10 [4.86e-10] & 5.73e-04 & 1179169.6x \\
dense & 0.003 & 2 & 4 & 0.025 & 2.01e-08 [2.02e-08] & 7.58e-04 & 37499.3x \\
dense & 0.003 & 2 & 4 & 0.062 & 2.10e-08 [2.17e-08] & 7.58e-04 & 35002.2x \\
dense & 0.003 & 2 & 4 & 0.125 & 2.09e-08 [2.15e-08] & 7.58e-04 & 35252.8x \\
dense & 0.003 & 2 & 8 & 0.025 & 2.18e-08 [2.20e-08] & 6.50e-04 & 29551.6x \\
dense & 0.003 & 2 & 8 & 0.062 & 2.34e-08 [2.45e-08] & 6.50e-04 & 26555.4x \\
dense & 0.003 & 2 & 8 & 0.125 & 2.30e-08 [2.40e-08] & 6.50e-04 & 27080.5x \\
dense & 0.003 & 4 & 4 & 0.025 & 2.51e-08 [2.68e-08] & 1.52e-03 & 56621.7x \\
dense & 0.003 & 4 & 4 & 0.062 & 2.60e-08 [2.80e-08] & 1.52e-03 & 54100.2x \\
dense & 0.003 & 4 & 4 & 0.125 & 2.78e-08 [2.86e-08] & 1.52e-03 & 53104.0x \\
dense & 0.003 & 4 & 8 & 0.025 & 2.86e-08 [2.97e-08] & 1.30e-03 & 43710.6x \\
dense & 0.003 & 4 & 8 & 0.062 & 2.85e-08 [2.93e-08] & 1.30e-03 & 44335.2x \\
dense & 0.003 & 4 & 8 & 0.125 & 2.99e-08 [3.04e-08] & 1.30e-03 & 42729.2x \\
dense & 0.003 & 8 & 4 & 0.025 & 5.46e-10 [5.60e-10] & 3.03e-03 & 5413132.5x \\
dense & 0.003 & 8 & 4 & 0.062 & 5.90e-10 [6.08e-10] & 3.03e-03 & 4990308.9x \\
dense & 0.003 & 8 & 4 & 0.125 & 5.81e-10 [6.22e-10] & 3.03e-03 & 4873886.7x \\
dense & 0.003 & 8 & 8 & 0.025 & 6.43e-10 [6.56e-10] & 2.60e-03 & 3962540.7x \\
dense & 0.003 & 8 & 8 & 0.062 & 6.04e-10 [6.23e-10] & 2.60e-03 & 4170839.4x \\
dense & 0.003 & 8 & 8 & 0.125 & 6.41e-10 [6.64e-10] & 2.60e-03 & 3919009.2x \\
dense & 0.003 & 16 & 4 & 0.025 & 3.58e-09 [3.75e-09] & 6.07e-03 & 1619030.7x \\
dense & 0.003 & 16 & 4 & 0.062 & 3.45e-09 [3.65e-09] & 6.07e-03 & 1660192.5x \\
dense & 0.003 & 16 & 4 & 0.125 & 3.84e-09 [4.10e-09] & 6.07e-03 & 1481184.5x \\
dense & 0.003 & 16 & 8 & 0.025 & 3.99e-09 [4.01e-09] & 5.20e-03 & 1297886.5x \\
dense & 0.003 & 16 & 8 & 0.062 & 4.04e-09 [4.09e-09] & 5.20e-03 & 1272147.2x \\
dense & 0.003 & 16 & 8 & 0.125 & 4.18e-09 [4.39e-09] & 5.20e-03 & 1185327.9x \\
\bottomrule
\end{tabular}\hspace{2em}\begin{tabular}{lrrrrrrr}
\toprule
family & $\rho$ & $d$ & $K$ & $\eta L$ & measured ratio & proven bound & slack \\
\midrule
dense & 0.01 & 2 & 4 & 0.025 & 2.34e-07 [2.36e-07] & 8.70e-03 & 36789.3x \\
dense & 0.01 & 2 & 4 & 0.062 & 2.25e-07 [2.28e-07] & 8.70e-03 & 38210.2x \\
dense & 0.01 & 2 & 4 & 0.125 & 2.47e-07 [2.50e-07] & 8.70e-03 & 34788.7x \\
dense & 0.01 & 2 & 8 & 0.025 & 2.52e-07 [2.60e-07] & 7.46e-03 & 28638.3x \\
dense & 0.01 & 2 & 8 & 0.062 & 2.62e-07 [2.74e-07] & 7.46e-03 & 27225.5x \\
dense & 0.01 & 2 & 8 & 0.125 & 2.58e-07 [2.68e-07] & 7.46e-03 & 27801.2x \\
dense & 0.01 & 4 & 4 & 0.025 & 2.93e-07 [3.06e-07] & 1.74e-02 & 56935.3x \\
dense & 0.01 & 4 & 4 & 0.062 & 2.79e-07 [2.90e-07] & 1.74e-02 & 59890.4x \\
dense & 0.01 & 4 & 4 & 0.125 & 3.21e-07 [3.28e-07] & 1.74e-02 & 53000.3x \\
dense & 0.01 & 4 & 8 & 0.025 & 3.07e-07 [3.18e-07] & 1.49e-02 & 46820.1x \\
dense & 0.01 & 4 & 8 & 0.062 & 3.18e-07 [3.20e-07] & 1.49e-02 & 46606.8x \\
dense & 0.01 & 4 & 8 & 0.125 & 3.37e-07 [3.42e-07] & 1.49e-02 & 43580.9x \\
dense & 0.01 & 8 & 4 & 0.025 & 6.24e-09 [6.49e-09] & 3.48e-02 & 5363897.7x \\
dense & 0.01 & 8 & 4 & 0.062 & 6.65e-09 [6.69e-09] & 3.48e-02 & 5199941.5x \\
dense & 0.01 & 8 & 4 & 0.125 & 6.24e-09 [7.16e-09] & 3.48e-02 & 4861841.3x \\
dense & 0.01 & 8 & 8 & 0.025 & 6.51e-09 [6.79e-09] & 2.98e-02 & 4394088.2x \\
dense & 0.01 & 8 & 8 & 0.062 & 6.81e-09 [7.25e-09] & 2.98e-02 & 4113015.9x \\
dense & 0.01 & 8 & 8 & 0.125 & 7.47e-09 [7.50e-09] & 2.98e-02 & 3974599.1x \\
dense & 0.01 & 16 & 4 & 0.025 & 3.92e-08 [3.96e-08] & 6.96e-02 & 1757140.9x \\
dense & 0.01 & 16 & 4 & 0.062 & 4.03e-08 [4.34e-08] & 6.96e-02 & 1603035.7x \\
dense & 0.01 & 16 & 4 & 0.125 & 4.27e-08 [4.56e-08] & 6.96e-02 & 1525842.7x \\
dense & 0.01 & 16 & 8 & 0.025 & 4.69e-08 [4.82e-08] & 5.96e-02 & 1238238.9x \\
dense & 0.01 & 16 & 8 & 0.062 & 4.70e-08 [4.79e-08] & 5.96e-02 & 1245270.4x \\
dense & 0.01 & 16 & 8 & 0.125 & 4.73e-08 [4.91e-08] & 5.96e-02 & 1215622.5x \\
dense & 0.03 & 2 & 4 & 0.025 & 2.08e-06 [2.11e-06] & 8.58e-02 & 40645.6x \\
dense & 0.03 & 2 & 4 & 0.062 & 2.21e-06 [2.34e-06] & 8.58e-02 & 36647.6x \\
dense & 0.03 & 2 & 4 & 0.125 & 2.22e-06 [2.26e-06] & 8.58e-02 & 37951.0x \\
dense & 0.03 & 2 & 8 & 0.025 & 2.33e-06 [2.41e-06] & 7.35e-02 & 30521.7x \\
dense & 0.03 & 2 & 8 & 0.062 & 2.32e-06 [2.36e-06] & 7.35e-02 & 31212.7x \\
dense & 0.03 & 2 & 8 & 0.125 & 2.49e-06 [2.50e-06] & 7.35e-02 & 29459.1x \\
dense & 0.03 & 4 & 4 & 0.025 & 2.66e-06 [2.78e-06] & 1.72e-01 & 61813.1x \\
dense & 0.03 & 4 & 4 & 0.062 & 2.88e-06 [2.94e-06] & 1.72e-01 & 58454.0x \\
dense & 0.03 & 4 & 4 & 0.125 & 2.74e-06 [2.87e-06] & 1.72e-01 & 59738.6x \\
dense & 0.03 & 4 & 8 & 0.025 & 2.85e-06 [3.10e-06] & 1.47e-01 & 47417.0x \\
dense & 0.03 & 4 & 8 & 0.062 & 2.88e-06 [2.92e-06] & 1.47e-01 & 50446.0x \\
dense & 0.03 & 4 & 8 & 0.125 & 3.11e-06 [3.12e-06] & 1.47e-01 & 47084.6x \\
dense & 0.03 & 8 & 4 & 0.025 & 5.77e-08 [6.21e-08] & 3.43e-01 & 5529640.8x \\
dense & 0.03 & 8 & 4 & 0.062 & 6.07e-08 [6.26e-08] & 3.43e-01 & 5478893.4x \\
dense & 0.03 & 8 & 4 & 0.125 & 6.18e-08 [6.28e-08] & 3.43e-01 & 5467469.4x \\
dense & 0.03 & 8 & 8 & 0.025 & 6.42e-08 [6.44e-08] & 2.94e-01 & 4566475.3x \\
dense & 0.03 & 8 & 8 & 0.062 & 6.68e-08 [6.77e-08] & 2.94e-01 & 4347034.8x \\
dense & 0.03 & 8 & 8 & 0.125 & 6.60e-08 [6.83e-08] & 2.94e-01 & 4304518.9x \\
dense & 0.03 & 16 & 4 & 0.025 & 3.65e-07 [3.72e-07] & 6.86e-01 & 1846793.2x \\
dense & 0.03 & 16 & 4 & 0.062 & 3.52e-07 [3.55e-07] & 6.86e-01 & 1935798.6x \\
dense & 0.03 & 16 & 4 & 0.125 & 4.18e-07 [4.63e-07] & 6.86e-01 & 1482623.2x \\
dense & 0.03 & 16 & 8 & 0.025 & 4.17e-07 [4.31e-07] & 5.88e-01 & 1366160.0x \\
dense & 0.03 & 16 & 8 & 0.062 & 4.17e-07 [4.51e-07] & 5.88e-01 & 1303899.0x \\
dense & 0.03 & 16 & 8 & 0.125 & 4.45e-07 [4.53e-07] & 5.88e-01 & 1299921.2x \\
\bottomrule
\end{tabular}
}
\end{table}

\begin{theorem}[Perturbative nonlinear variance law]
\label{thm:perturb}
Let $U_\epsilon(x)=\frac12(x-\mu)^\top H(x-\mu)+\epsilon V(x)$,
$A=I-\eta H$, $0\prec H\prec 2\eta^{-1}I$, and $V\in C_b^3$.  Fix finite
$t$, use common deterministic initialization, and draw zero-sum coupled
innovations independently across time and independently of the deterministic
start.  Let $X_s^{0,i}$ denote the quadratic reference process at
$\epsilon=0$, $m_s=K^{-1}\sum_i X_s^{0,i}$ (deterministic), and
\[
 D_s=\frac1K\sum_{i=1}^K
 \bigl[\nabla V(X_s^{0,i})-\nabla V(m_s)-\nabla^2V(m_s)(X_s^{0,i}-m_s)\bigr].
\]
Then for every fixed probe $a$,
$\operatorname{Var}(a^\top\bar X_t^\epsilon)
=\epsilon^2\eta^2\,\operatorname{Var}\!\bigl(\sum_{s=0}^{t-1}a^\top A^{t-1-s}D_s\bigr)
+O(\epsilon^3)$.
The remainder is uniform in probes with $\|a\|\le A_0$ and in
$|\epsilon|\le\epsilon_0$ for constants depending on $t,\eta,H,V,A_0,K$ and
the initial point.
\end{theorem}

\begin{proof}
For finite $t$, the coupled recursion is differentiable in $\epsilon$ by
iterating smooth maps.  Let
$Y_s^{(i)}=\partial_\epsilon X_s^{\epsilon,i}|_{\epsilon=0}$.
Differentiating the recursion gives $Y_0^{(i)}=0$ and
$Y_{s+1}^{(i)}=AY_s^{(i)}-\eta\nabla V(X_s^{0,i})$.  Averaging and expanding
$\nabla V(X_s^{0,i})$ around $m_s$, the constant term is deterministic and
the linear term cancels because $\sum_i(X_s^{0,i}-m_s)=0$; the only random
first-order forcing of the ensemble mean is therefore $D_s$.  Solving the
linear recursion gives the displayed first-order term.  Bounded derivatives
of $V$ and the fixed finite horizon yield, by another induction, a uniform
second-order Taylor expansion in $L^2$ for all probes with $\|a\|\le A_0$,
$a^\top\bar X_t^\epsilon=c_t(a)+\epsilon Z_t(a)+\epsilon^2 R_{\epsilon,t}(a)$,
with the stated bound on $R_{\epsilon,t}$.  Then
$\operatorname{Var}(c+\epsilon Z+\epsilon^2 R)
=\epsilon^2\operatorname{Var}(Z)+O(\epsilon^3)$ by Cauchy--Schwarz.
\end{proof}

\section{Metropolis acceptance: the rejection-leak law}
\label{app:mala-law}

Suppose seven of eight chains accept a proposal. Before the Metropolis
decisions, the eight Gaussian draws summed to zero. The seven that
remain must therefore sum to the negative of the rejected draw.
With more than one rejection we need the sum of the missing draws,
which is just as easy to log. We use independent acceptance uniforms
across chains, independent also of the proposals.

\begin{proposition}[Rejection-leak decomposition]
\label{prop:rejection-leak}
Let coupled MALA propose increments
$\delta_i=-\eta\nabla U(X_t^{(i)})+\sqrt{2\eta}\,\xi_t^{(i)}$ with any
zero-sum coupling ($\sum_i\xi_t^{(i)}=0$) and accept with per-chain
indicators $A_i\in\{0,1\}$.  Then
\[
\bar X_{t+1}=\bar X_t+\frac1K\sum_{i=1}^K A_i\delta_i,
\qquad
\sum_{i=1}^K A_i\,\xi_t^{(i)}=-\sum_{i=1}^K (1-A_i)\,\xi_t^{(i)}.
\]
If all chains accept, the Gaussian term in the average step vanishes
as it did in ULA. A rejection removes one chain from the
zero-sum constraint and leaks that chain's innovation, scaled by
$\sqrt{2\eta}/K$, into the average (simultaneous rejections leak their
zero-sum-correlated sum).
\end{proposition}

\begin{proof}
Immediate from $\sum_i\xi_t^{(i)}=0$ and $A_i=1-(1-A_i)$.
\end{proof}

\paragraph{Leading-order law.}
The decomposition above is exact. To estimate the variance it leaves,
suppose rejections are rare and consider a stationary Gaussian target.
Writing $\xi_a$ for the projection of the proposal noise along $a$, the
leading-order calculation suggests
\[
\frac{\operatorname{Var}\!\bigl(a^\top\bar X\bigr)_{C_\star}}
     {\operatorname{Var}\!\bigl(a^\top\bar X\bigr)_{\mathrm{iid}}}
\;=\;
p\,s+o(p),
\qquad
s=\frac{\mathbb E\bigl[\xi_a^2\mid\text{reject}\bigr]}
        {\mathbb E\bigl[\xi_a^2\bigr]},
\]
where $p$ is the per-chain rejection probability.  On a one-dimensional Gaussian target $s\to2$ as $\eta\to0$, by the
identity $\mathbb E[(x\xi)_+\xi^2]=2\mathbb E[(x\xi)_+]$ for independent
standard normals. In more dimensions, rejection selects less strongly
on any one coordinate, so the Gaussian leading-order factor lies between
one and two. At fixed dimension the smooth-target MALA rejection rate has
order $\eta^{3/2}$. On the Gaussian example this predicts the same order
for the retained variance at small steps. A nonlinear drift can itself
leave variance even when every chain accepts. For the logistic
experiments we therefore use $p s$ as a measured diagnostic, with $s$
computed from the proposal-noise energy. We have no general bound on
the remainder in this leading-order calculation.

\paragraph{Verification.}
On a one-dimensional Gaussian target ($K=8$, $10^5$ independent ensembles
per step size; script \texttt{verify\_mala\_law\_1d.py}), the measured
ratio matches $2p$ within $5\%$ for rejection rates below $1\%$, within
$10\%$ up to $2\%$, and within $40\%$ up to rejection rate $0.28$, and the
ratio is unchanged across $K\in\{4,8,16\}$ ($0.0438$--$0.0441$ at
$\eta=0.2$), confirming the predicted $K$-independence.  Table~\ref{tab:mala-law} shows
the real-data check on the tempered logistic posteriors: the per-cell
measured $(p,s)$, obtained by logging each rejected proposal's noise
energy, reproduce the matched $C_\star$/iid trace-variance ratios of
Table~\ref{tab:mala-real} with no fitted constants.

\begin{table}[htbp]
\centering
\small
\caption{How much variance returns when proposals are rejected? At
$K=8$ we estimated the rejection probability $p$ and the mean-square
noise on a rejected step relative to all steps, $s$, from the centered
runs at the calibrated step sizes. Entries are medians over three
replicates. The median replicate-level product falls within the measured replicate range in
four cells out of six. The largest discrepancy as a fraction of $p s$
is $31.4\%$ (WDBC, prior $0.1$). Dividing by the measured ratio
instead gives a maximum of $27.6\%$ (Adult, prior $0.1$).}

\label{tab:mala-law}
\resizebox{.85\linewidth}{!}{\begin{tabular}{llrrrr}
\toprule
Dataset & prior & rejection rate $p$ & selection $s$ & predicted $p\,s$ & measured $C_\star$/iid \\
\midrule
WDBC & 0.1 & 0.114 & 1.117 & 0.128 & 0.168 [0.140, 0.178] \\
WDBC & 0.3 & 0.259 & 1.128 & 0.293 & 0.267 [0.256, 0.328] \\
Spambase & 0.1 & 0.130 & 1.083 & 0.141 & 0.130 [0.129, 0.141] \\
Spambase & 0.3 & 0.332 & 1.082 & 0.359 & 0.337 [0.333, 0.342] \\
Adult & 0.1 & 0.189 & 1.047 & 0.198 & 0.155 [0.154, 0.236] \\
Adult & 0.3 & 0.160 & 1.056 & 0.169 & 0.163 [0.155, 0.189] \\
\bottomrule
\end{tabular}
}
\end{table}
\FloatBarrier

\section{Full experimental protocol and additional diagnostics}
\label{app:experiments}

The chain-count and timing comparisons in Section~\ref{sec:experiments}
use the protocols in Sections~\ref{app:exp-protocol}--\ref{app:exp-cost},
with checks for different observables in Section~\ref{app:exp-observable}.
For changes to the transition see Sections~\ref{app:exp-transition}
and~\ref{app:exp-hmc}, or Section~\ref{app:exp-reference} for accuracy
against the long MALA reference.
Sections~\ref{app:exp-posterior}--\ref{app:exp-occupancy} cover the other
posteriors, fitted control variates and mode occupancy, and
Section~\ref{app:exp-additional} gives the finite-sum and timing checks.

\subsection{Shared protocols and synthetic targets}\label{app:exp-protocol}
We measure variance across independent ensembles, each containing $K$ coupled
chains. In the logistic campaign, a cell is a dataset--prior pair, and a
replicate batches $G$ independent ensembles under a fresh seed. Cell medians
are taken over replicates. Unless noted otherwise, matched runs share the
cell, replicate, chain count and sampling schedule.

The main logistic campaign uses three replicates per cell. At $K=8$, $G=256$
for WDBC, $192$ for Spambase and $64$ for Adult. The control-variate study
uses its own ensemble counts. Its tables and the mode-occupancy tables retain
their pooled ratios, pre-registered bands, unresolved classifications and
two-standard-error intervals.

\FloatBarrier

Table~\ref{tab:summary} summarizes median variance ratios for Gaussian,
quartic and logistic targets. The four couplings share 240 seeded target
configurations, yielding 960 rows. Dimensions start at 64 and extend to 4096
for Gaussian quadratics, 2048 for quartic perturbations and 512 for logistic
regression. We vary the quartic and Gaussian-prior strengths, with
quadratic-part condition numbers up to 100. Chain counts are
$K\in\{2,4,8,16\}$ where applicable. Separate odd-$K$ ablations use
$K\in\{5,7\}$ for real data and $K\in\{3,5,7,9,15\}$ for synthetic targets.

A paired initialization sweep holds the target and probe design fixed across
couplings, giving 1,296 seeded designs and 5,184 coupling-specific rows. It
compares common zero, iid random, common random and zero-sum random starts
over horizons of 50, 250 and 1,000 steps, designated short, medium and long,
respectively. Each configuration in these two synthetic sweeps uses 128--4096
independent ensembles.

A single random-number stream is used for the appendix diagnostics, with its
seed specified in the released code. Covariance, weighted-average, marginal and quadratic-observable
checks use 100,000 ensembles, nonlinear checks use 50,000, and the CPU
finite-sum SGLD diagnostic uses 2,000.

The repository at \mbox{\url{https://github.com/rfarell/parallel-langevin-chains}}
holds the code, raw results and the full tables. Unless captions specify an interquartile range or a min--max range of
replicate ratios, reported Monte Carlo intervals are approximate, with a
nominal 95\% level.

\FloatBarrier
\begin{table}[htbp]
\centering
\small
\caption{Median synthetic variance ratios.  Gaussian cancellation is exact
(at numerical roundoff) and nonlinear variance ratios remain far below iid.
Tail behavior is summarized separately for the time-averaged logistic
estimator in Table~\ref{tab:logistic-mse}.}
\label{tab:summary}
\begin{tabular}{lrrrr}
\toprule
Target & iid & $C_\star$ & pairwise & common \\
\midrule
gaussian & 0.170 & 7.79e-15 & 1.00e-15 & 1.000 \\
quartic & 0.122 & 6.85e-09 & 1.00e-15 & 1.000 \\
logistic & 0.121 & 3.12e-09 & 2.62e-13 & 1.000 \\
\bottomrule
\end{tabular}

\end{table}

\begin{table}[htbp]
\centering
\small
\caption{Time-averaged posterior means on $360$ synthetic logistic posteriors.
Dividing MSE by the iid MSE at the same target and chain count gives a
median of $0.002$ for $C_\star$. Even the largest ratio in the sweep
was only $0.011$.}
\label{tab:logistic-mse}
\begin{tabular}{lrrrrr}
\toprule
Method & med MSE/iid & q90 & max & med var/iid & med MSE/single \\
\midrule
iid & 1.000 & 1.000 & 1.000 & 1.000 & 0.125 \\
$C_\star$ & 0.002 & 0.008 & 0.011 & 1.30e-08 & 2.71e-04 \\
antithetic & 0.005 & 0.202 & 0.207 & 7.20e-12 & 5.13e-04 \\
common & 8.011 & 16.046 & 16.365 & 8.032 & 1.000 \\
\bottomrule
\end{tabular}

\end{table}
\FloatBarrier

\begin{figure}[htbp]
\centering\includegraphics[width=.59\linewidth]{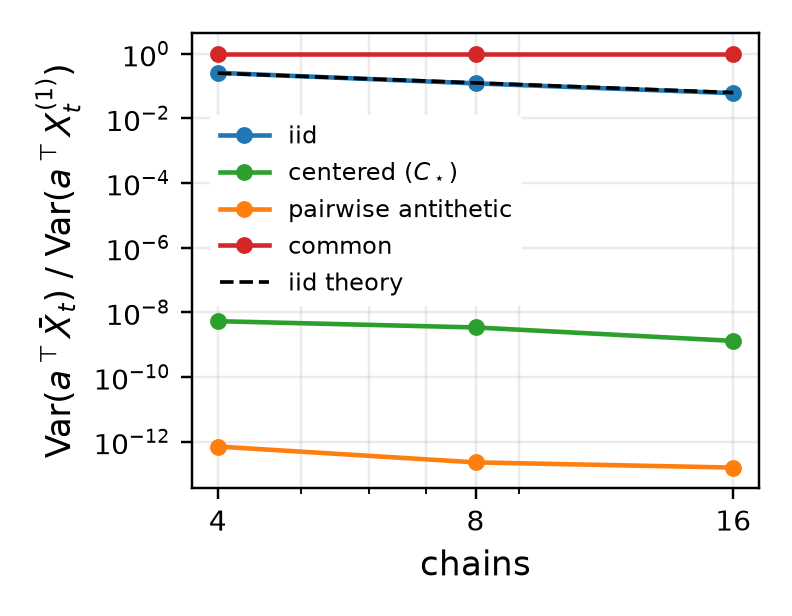}
\caption{On synthetic logistic targets the variance of a linear projection
of the ensemble mean rises above the Gaussian numerical floor. All curves
use the single-chain variance as denominator.}\label{fig:synthetic-logistic}
\end{figure}

\subsection{Chain count and sampling cost}\label{app:exp-cost}
\begin{figure}[htbp]
\centering
\includegraphics[width=.98\textwidth]{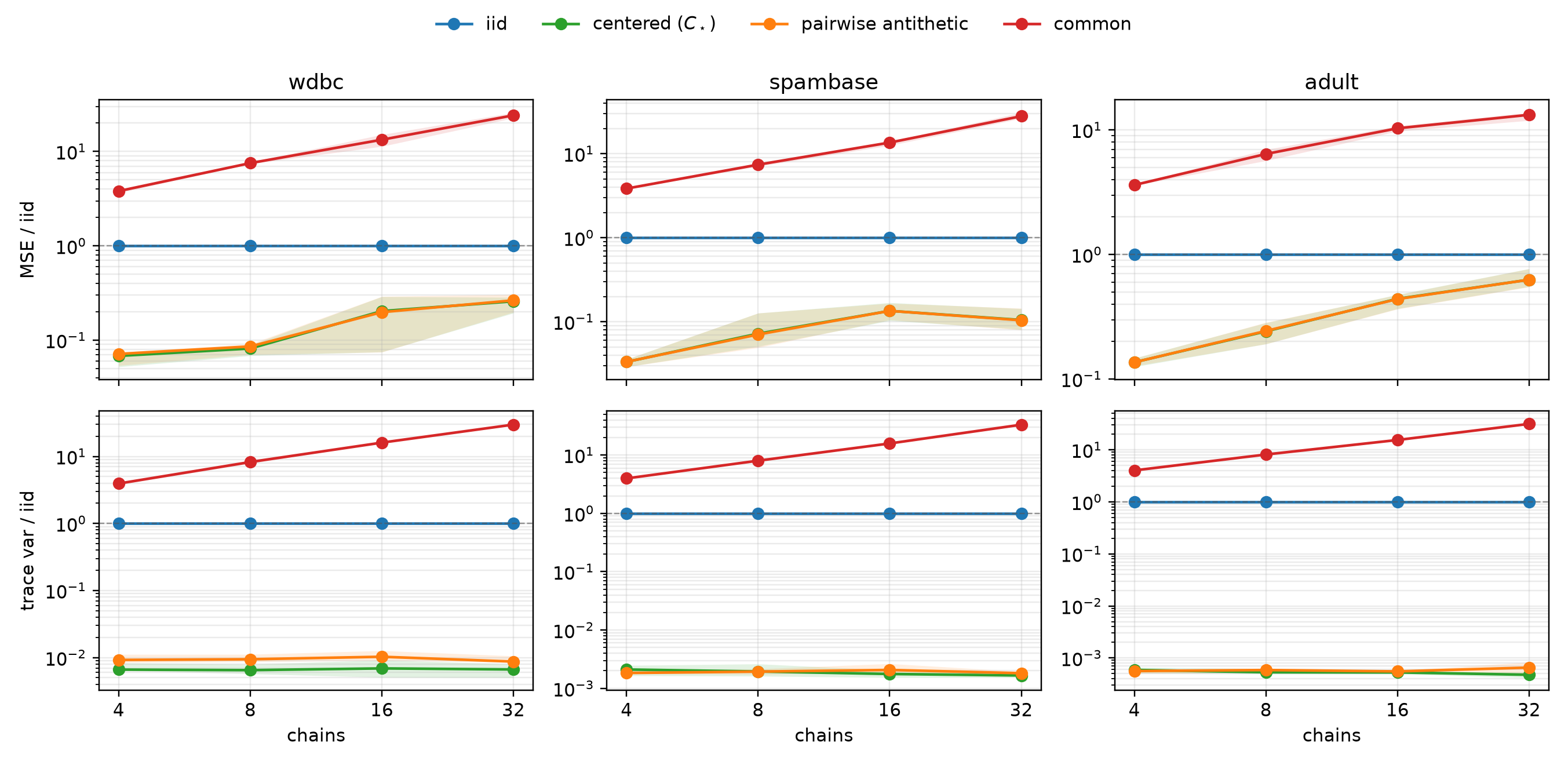}
\caption{What changes when both samplers get more chains? At each $K$ we divide
by iid on the same dataset, prior and replicate. In the bottom panels
the trace variance of the centered estimator stays two to three orders
of magnitude below iid. In the top panels the advantage in MSE begins
to level off as the discrepancy from the finite reference comes to
dominate. Lines give medians over matched priors and replicates, and
shading gives their interquartile ranges.}
\label{fig:campaign-real-k}
\end{figure}

\begin{figure}[htbp]
\centering
\includegraphics[width=.98\textwidth]{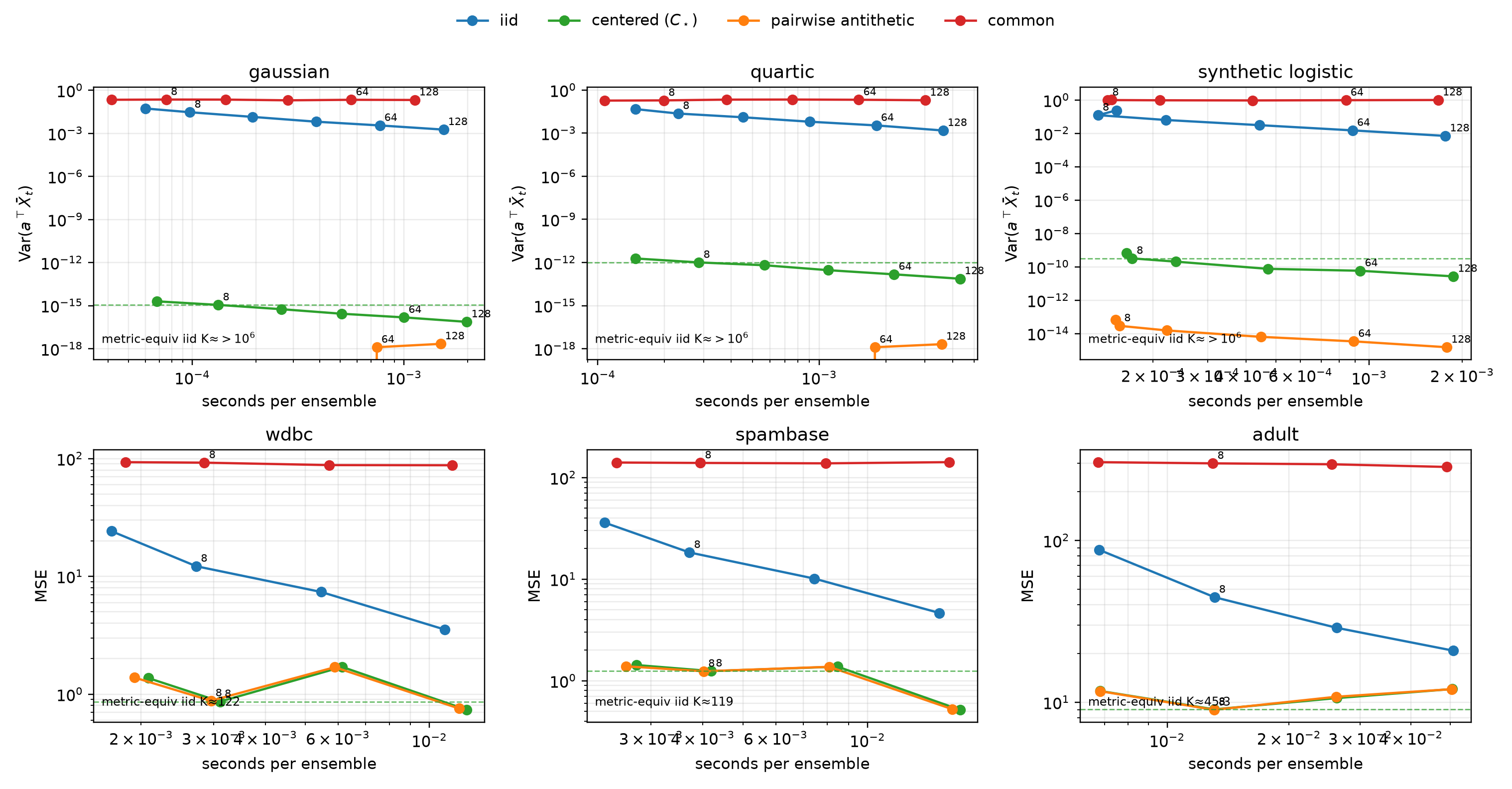}
\caption{Timing and metric campaign across synthetic and real targets.
$C_\star$ reaches the same metric as iid with smaller $K$ on each reported
target.  Synthetic panels sit at the numerical floor, and real panels are
capped by reference discrepancy rather than by the coupling itself.
In-panel labels are iid $1/K$-extrapolated chain counts for the same
metric, reported as a scale reference rather than a wall-clock equivalence.}
\label{fig:campaign-metric-equivalent}
\end{figure}

The campaign in Figures~\ref{fig:campaign-real-k}
and~\ref{fig:campaign-metric-equivalent} uses three independent
replicates on A100 GPUs.  Synthetic Gaussian, quartic, and logistic targets
use chain counts $K\in\{4,8,16,32,64,128\}$ after an unmeasured warm-up
step, so the timing excludes one-time kernel compilation.  Real-data
logistic regression uses full-gradient ULA on standardized UCI datasets.
WDBC has $455/114$ train/test points and $30$ features, Spambase has
$3681/920$ and $57$, and Adult has $30162/15060$ and $104$ after dropping
rows with missing categorical values and one-hot encoding.  Each real-data
state includes an unpenalized intercept.  We divide the logistic loss by $n$ before adding the prior penalty
$\tfrac\lambda2\|\beta\|^2$ on the non-intercept coordinates. The resulting
likelihood contribution is $p(y\mid\beta)^{1/n}$, which is what we mean
by a tempered posterior throughout the paper.  Non-intercept coefficients use
Gaussian priors with precision $0.1$ or $0.3$, and the original $K$-sweep
uses chain counts $K\in\{4,8,16,32\}$.  WDBC and Spambase use $900$ ULA steps
with $450$ burn-in and a longer iid reference of $1800$ steps.  Adult uses
$700/350$ and $1400/700$ to keep the large-data stress test within the same GPU
campaign.  Raw campaign JSONL rows record elapsed seconds, seeds, GPU id,
posterior-mean MSE against the longer reference, trace variance, probe
variance, and finite-budget test log loss.  The preprocessing and download
script records the UCI DOI and license metadata for each dataset.

Table~\ref{tab:campaign-real-full} gives the longer-reference error, test log loss and sampling time alongside trace variance for all four couplings.

The main runs and fresh-seed repeats in Tables~\ref{tab:campaign-highk-iid}
and~\ref{tab:real-odd-k} use the same full-gradient tempered logistic targets,
preprocessing, priors, ULA horizons, and posterior-mean trace-variance
estimator.  They add measured iid $K=128$ baselines for $K=8$ $C_\star$ and
odd-$K$ runs at $K=7$ and $K=5$.  Ratios are computed within matched
dataset/prior/replicate cells and then medianed; bracketed ranges report the
minimum and maximum matched replicate ratio in that cell.  Trace
variances use the population ($1/n_{\mathrm{ens}}$) normalization; this
cancels in matched same-$K$ ratios, and applying the unbiased
$1/(n_{\mathrm{ens}}-1)$ correction instead shifts the cross-$K$
($K{=}8$ vs $K{=}128$) comparison by $2.7$--$3.7\%$ without changing its
conclusion.

Table~\ref{tab:campaign-wallclock-frontier} uses the same real-data targets
and charges each trace-variance estimate by measured sampling-loop seconds per
independent ensemble.  The ``Start scale'' column is the standard deviation of
the iid Gaussian initial offsets; $0.00$ denotes zero initialization.  The iid
frontier is the best measured cost among $K\in\{8,16,32,64,128\}$ in each
matched replicate, and the odd-$K$ frontier is the best of iid and conventional
pair+1 antithetic within that replicate at the same odd $K$. The Max variance-time ratio column is the
larger of the iid-frontier and odd-$K$ cost ratios after taking worst-cell
medians.

\FloatBarrier
\begin{table}[htbp]
\centering
\scriptsize
\caption{Full real-data full-gradient tempered logistic ULA campaign at
$K=8$.  Ratios are normalized to iid at the same dataset, prior, replicate,
and chain count, and aggregate matched runs across three GPU replicates and
two Gaussian-prior strengths.}
\label{tab:campaign-real-full}
\resizebox{\linewidth}{!}{\begin{tabular}{lllrrrr}
\toprule
Dataset & coupling & ref-MSE/iid [IQR] & var/iid & ref-floor/MSE & log loss & sec/ens \\
\midrule
wdbc & iid & 1.000 [1.000, 1.000] & 1.000 & 0.083 & 0.342 & 0.003 \\
wdbc & $C_\star$ & 0.082 [0.068, 0.089] & 0.006 & 0.924 & 0.254 & 0.003 \\
wdbc & antithetic & 0.086 [0.071, 0.091] & 0.009 & 0.892 & 0.255 & 0.003 \\
wdbc & common & 7.538 [7.405, 7.730] & 8.252 & 0.013 & 0.887 & 0.003 \\
\addlinespace
spambase & iid & 1.000 [1.000, 1.000] & 1.000 & 0.081 & 0.704 & 0.004 \\
spambase & $C_\star$ & 0.072 [0.051, 0.127] & 0.002 & 0.974 & 0.462 & 0.004 \\
spambase & antithetic & 0.071 [0.049, 0.127] & 0.002 & 0.972 & 0.462 & 0.004 \\
spambase & common & 7.409 [7.073, 7.684] & 8.006 & 0.010 & 2.028 & 0.004 \\
\addlinespace
adult & iid & 1.000 [1.000, 1.000] & 1.000 & 0.256 & 1.063 & 0.013 \\
adult & $C_\star$ & 0.242 [0.191, 0.283] & 5.31e-04 & 0.998 & 0.677 & 0.013 \\
adult & antithetic & 0.244 [0.192, 0.285] & 5.86e-04 & 0.998 & 0.678 & 0.013 \\
adult & common & 6.415 [5.665, 6.961] & 8.139 & 0.041 & 2.551 & 0.013 \\
\addlinespace
\bottomrule
\end{tabular}
}
\end{table}
\FloatBarrier

\begin{table}[htbp]
\centering
\scriptsize
\caption{Measured large-$K$ iid comparisons for the main
posterior-mean trace-variance metric.  The main run and repeat columns report
median matched replicate ratios above their min--max ranges.  The ensemble-count
column lists $C_\star K=8$ ensembles / iid $K=128$ ensembles used for each
trace estimate; these are independent ensemble posterior-mean estimates, not
numbers of chains.  $K=8$ $C_\star$ remains below $0.135\times$ measured
$K=128$ iid in every cell of the repeat, while iid $K=8$/iid $K=128$ is close to the
expected $16\times$ scaling.  The even-$K$ pairwise column serves as a
calibration check; the odd-$K$ tables carry the empirical distinction from
conventional pair+1 antithetic coupling.}
\label{tab:campaign-highk-iid}
\setlength{\tabcolsep}{2.5pt}
\begin{tabular}{lcccccc}
\toprule
 & & & \multicolumn{2}{c}{$C_\star(8)$/iid(128)} & iid(8)/iid(128) & $C_\star(8)$/pair(8) \\
\cmidrule(lr){4-5}
Dataset & Prior & Ensembles & Main run & Repeat & Repeat & Repeat \\
\midrule
WDBC & 0.1 & 256/24 & \shortstack{0.0971 \\{} [0.0918, 0.0994]} & \shortstack{0.0867 \\{} [0.0865, 0.0887]} & \shortstack{15.3823 \\{} [14.8143, 18.4268]} & \shortstack{0.6387 \\{} [0.5645, 0.6864]} \\[2pt]
WDBC & 0.3 & 256/24 & \shortstack{0.1189 \\{} [0.0948, 0.1418]} & \shortstack{0.1344 \\{} [0.1204, 0.1439]} & \shortstack{15.4385 \\{} [14.9612, 16.8423]} & \shortstack{0.7365 \\{} [0.6997, 0.7433]} \\[2pt]
Spambase & 0.1 & 192/24 & \shortstack{0.0255 \\{} [0.0233, 0.0255]} & \shortstack{0.0245 \\{} [0.0244, 0.0264]} & \shortstack{16.8097 \\{} [16.2010, 17.3489]} & \shortstack{0.8424 \\{} [0.8101, 0.9145]} \\[2pt]
Spambase & 0.3 & 192/24 & \shortstack{0.0398 \\{} [0.0339, 0.0493]} & \shortstack{0.0453 \\{} [0.0409, 0.0472]} & \shortstack{17.1544 \\{} [16.9403, 18.1698]} & \shortstack{1.0682 \\{} [1.0372, 1.4424]} \\[2pt]
Adult & 0.1 & 64/24 & \shortstack{0.0089 \\{} [0.0087, 0.0090]} & \shortstack{0.0090 \\{} [0.0081, 0.0090]} & \shortstack{16.6287 \\{} [16.1336, 16.7076]} & \shortstack{0.9748 \\{} [0.9267, 1.1122]} \\[2pt]
Adult & 0.3 & 64/24 & \shortstack{0.0086 \\{} [0.0086, 0.0104]} & \shortstack{0.0085 \\{} [0.0085, 0.0104]} & \shortstack{16.5967 \\{} [16.1803, 17.3905]} & \shortstack{0.7976 \\{} [0.7378, 1.0281]} \\[2pt]
\bottomrule
\end{tabular}

\end{table}
\FloatBarrier

\begin{table}[htbp]
\centering
\scriptsize
\caption{Cost of estimating the posterior mean on one A100. We multiply trace
variance by measured sampling seconds per independent ensemble. The
first benchmark gives iid whichever of $K=8,16,32,64,128$ has the
smallest product within each matched replicate, then divides the product for eight
$C_\star$ chains by it. The other benchmark compares $C_\star$ with
the best odd baseline within that replicate at the same odd $K$. We report the worst
cell-median ratio across three datasets and two priors. Where both
comparisons apply, the last column takes the larger ratio. Both ratios
are largest for overdispersed starts.}
\label{tab:campaign-wallclock-frontier}
\resizebox{\linewidth}{!}{\begin{tabular}{llclccc}
\toprule
Block & Replicates & Start scale & Worst cell & $C_\star K=8$/best iid cost & $C_\star$ odd-$K$/best odd cost & Max variance-time ratio \\
\midrule
primary & 300--305 & 0.00 & WDBC, 0.3 & 0.0119 & 0.0523 ($K=7$) & 0.0523 \\
cross-node stress & 306--311 & 0.00 & WDBC, 0.3 & 0.0114 & 0.0555 ($K=7$) & 0.0555 \\
independent starts & 312--317 & 0.25 & WDBC, 0.3 & 0.0115 & 0.0508 ($K=7$) & 0.0508 \\
overdispersed starts & 318--323 & 1.00 & WDBC, 0.3 & 0.0162 & 0.0661 ($K=7$) & 0.0661 \\
$K=5$ edge & 324--329 & 0.00 & WDBC, 0.3 & -- & 0.0421 ($K=5$) & 0.0421 \\
\bottomrule
\end{tabular}
}
\end{table}
\FloatBarrier

\FloatBarrier
\subsection{Checks of the covariance identity and the choice of coupling}\label{app:exp-observable}

Full tables for the time-weighted, nonlinear and odd-chain checks below
are in the code repository.

For the Gaussian target, the simulated variances were $0.9982$--$0.9989$
times the exact variances from Theorem~\ref{thm:weighted}. We tested three sets of time
weights, uniform or favoring either end of the run, with fresh draws for
each set. The single-chain terminal second moment likewise agreed among
couplings to within sampling error (Table~\ref{tab:diagnostic-marginal}).

Taking each quantity as a fraction of its bound, the largest values we
observed on the two cosine targets were $0.5359$ for the local gradient
remainder and $0.1574$ for the terminal distance between the ensemble mean
and the gradient descent path. In a separate sweep over perturbation size,
we obtained a log--log slope of $1.86$ for the ensemble-mean variance.
This is close to the exponent two in Theorem~\ref{thm:perturb}.

\begin{table}[htbp]
\centering
\small
\caption{The effect of cross-chain correlation on variance. The factor
$\mathbf 1^\top C\mathbf 1/K^2$ in \eqref{eq:single-probe} is exact for each
equicorrelation $r$. The simulation tracks it between the iid, common and
centered cases.}
\label{tab:diagnostic-correlation}
\begin{tabular}{lrrr}
\toprule
$r$ & exact Var(mean) & MC Var(mean) & MC-exact \\
\midrule
-0.1429 & 1.56e-17 & 2.31e-33 [2.29e-33, 2.33e-33] & -1.56e-17 \\
-0.1000 & 0.0375 & 0.0373 [0.0370, 0.0377] & -1.53e-04 \\
0 & 0.1250 & 0.1252 [0.1241, 0.1263] & 1.88e-04 \\
0.2000 & 0.3000 & 0.3025 [0.2998, 0.3051] & 0.0025 \\
0.5000 & 0.5625 & 0.5633 [0.5584, 0.5683] & 8.27e-04 \\
0.8000 & 0.8250 & 0.8264 [0.8191, 0.8336] & 0.0014 \\
1.0000 & 1.0000 & 0.9974 [0.9887, 1.0062] & -0.0026 \\
\bottomrule
\end{tabular}

\end{table}

\begin{table}[htbp]
\centering
\small
\caption{Zero-sum coupling changes cross-chain covariance without changing a
single chain's marginal law.  The first chain's terminal squared norm and a
one-dimensional probe match the iid-coupled reference up to sampling noise.}
\label{tab:diagnostic-marginal}
\begin{tabular}{lrrrr}
\toprule
coupling & exact $E\|X_1\|^2/d$ & MC estimate & MC/exact & KS vs iid \\
\midrule
iid & 0.9361 & 0.9340 [0.9299, 0.9380] & 0.9977 & 0 \\
$C_\star$ & 0.9361 & 0.9360 [0.9320, 0.9401] & 0.9999 & 0.0043 \\
antithetic & 0.9361 & 0.9367 [0.9326, 0.9407] & 1.0006 & 0.0029 \\
common & 0.9361 & 0.9336 [0.9296, 0.9376] & 0.9973 & 0.0033 \\
\bottomrule
\end{tabular}

\end{table}

\begin{table}[htbp]
\centering
\caption{Median Gaussian variance ratios under randomized starts.  This
validates Theorem~\ref{thm:init}.  Zero-sum innovation coupling removes
only the injected-noise term, so iid initial scatter remains early in time
but contracts as the chain mixes, while zero-sum randomized starts retain
the exact cancellation throughout.}
\label{tab:init}
\begin{tabular}{llrrr}
\toprule
Steps & Initialization & iid & $C_\star$ & common \\
\midrule
50 & common zero & 0.125 & 3.80e-15 & 1.000 \\
50 & zero-sum random & 0.061 & 4.61e-15 & 0.528 \\
50 & iid random & 0.129 & 0.032 & 0.637 \\
50 & common random & 0.530 & 0.472 & 1.000 \\
250 & common zero & 0.125 & 8.04e-15 & 1.000 \\
250 & zero-sum random & 0.102 & 9.04e-15 & 0.975 \\
250 & iid random & 0.124 & 0.003 & 0.975 \\
250 & common random & 0.248 & 0.028 & 1.000 \\
1000 & common zero & 0.126 & 9.35e-15 & 1.000 \\
1000 & zero-sum random & 0.116 & 8.98e-15 & 1.000 \\
1000 & iid random & 0.125 & 6.18e-07 & 1.000 \\
1000 & common random & 0.149 & 4.08e-06 & 1.000 \\
\bottomrule
\end{tabular}

\end{table}

\begin{table}[htbp]
\centering
\small
\caption{A Gaussian observable with a linear and a centered quadratic term.
Increasing the weight on the quadratic term moves the measured
variance under $C_\star$ toward $1/(K-1)$ of the single-chain variance.
The crossover with iid occurs when the saving on the linear term
just pays for the larger quadratic contribution
(Proposition~\ref{prop:quad-observable} and Corollary~\ref{cor:hermite-winning}).}
\label{tab:diagnostic-quadratic-observable}
\begin{tabular}{lrrr}
\toprule
quadratic $\beta$ & exact ratio & MC ratio & MC/exact \\
\midrule
0 & 0 & 7.59e-33 [7.52e-33, 7.65e-33] & -- \\
0.5000 & 0.0286 & 0.0283 [0.0281, 0.0285] & 0.9905 \\
1.0000 & 0.0714 & 0.0706 [0.0700, 0.0713] & 0.9890 \\
2.0000 & 0.1143 & 0.1130 [0.1120, 0.1140] & 0.9885 \\
\bottomrule
\end{tabular}

\end{table}

\begin{figure}[htbp]
\centering
\includegraphics[width=.55\textwidth]{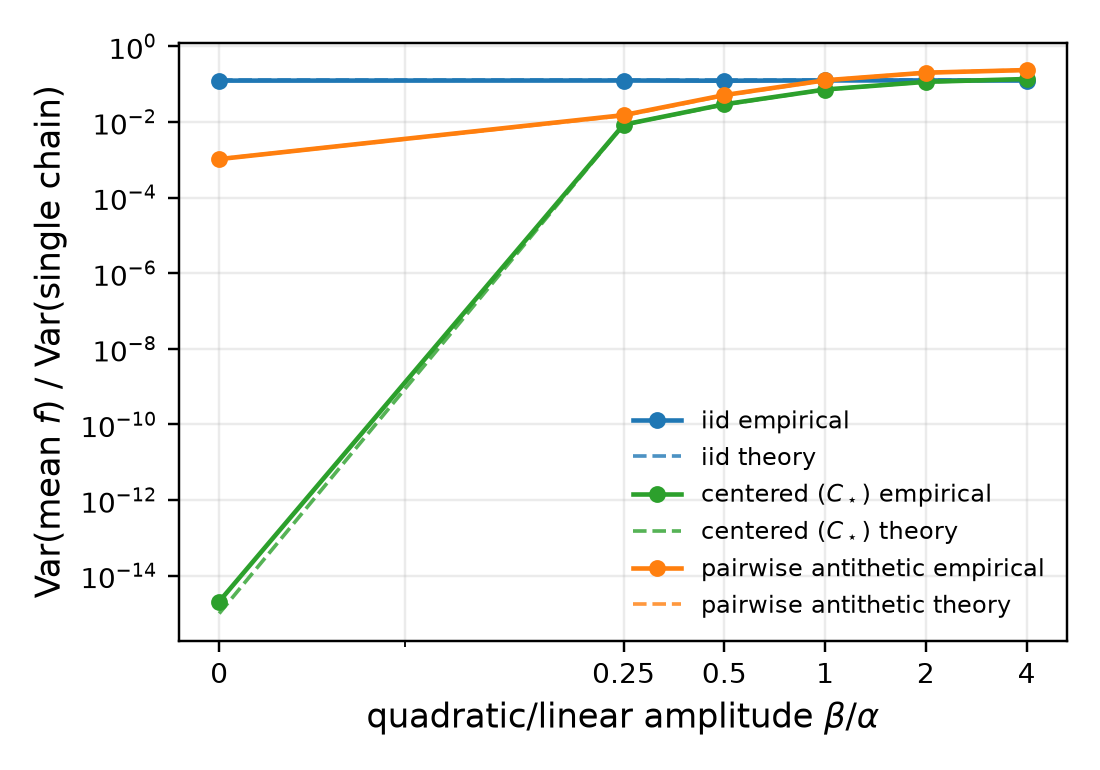}
\caption{The crossover for a mixed Gaussian observable. We varied the weights
in $f_{\alpha,\beta}(x)=\alpha u^\top z+\beta(\|z\|^2-d)/\sqrt{2d}$
and pooled medians over $K\in\{3,\ldots,63\}$. At each $K$ the variance
for $C_\star$ follows the dashed prediction of Theorem~\ref{thm:hermite}
and crosses iid at $\alpha^2=\beta^2/(K-1)$. At odd $K$ an unpaired
chain still contributes linear noise, which gives $C_\star$ an
advantage over pair+1 in that regime.}
\label{fig:hermite}
\end{figure}

\paragraph{An odd number of chains.} Leaving a draw unmatched in pair+1
gave median ratios of ensemble variance to single-chain variance between
$0.020$ and $0.025$ on the Gaussian, quartic and logistic targets. With
$C_\star$, none of these medians exceeded $1.12\times10^{-7}$.
The block baseline consists of two sign-flip
pairs plus one three-chain simplex block at $K=7$, and one pair plus one
three-chain block at $K=5$; the protocol is otherwise unchanged.

\begin{table}[htbp]
\centering
\scriptsize
\caption{Real-data comparison with an odd number of chains.  At $K=7$, conventional
pairwise antithetic coupling uses three antithetic pairs plus one unmatched
iid chain, all equally weighted; at $K=5$ it uses two pairs plus one unmatched
chain.  The $n_{\rm ens}$ column is the number of independent ensemble
posterior-mean estimates per method used to estimate trace variance.  The
$C_\star$/odd-pair+1 columns report median matched replicate ratios with
min--max ranges.  The $K=7$ main run and repeat use six replicates each;
the $K=5$ block is a three-replicate edge check.  Odd-pair+1/iid ratios sit
near $1/K$, consistent with
the residual unmatched-chain innovation mode, while $C_\star$ remains
zero-sum; the worst cell-median $C_\star$/pair+1 ratio is $0.0559$ at $K=7$
and $0.0437$ in the $K=5$ edge check.  This is a scoped comparison to the
conventional equal-weight odd-$K$ pairwise construction.}
\label{tab:real-odd-k}
\resizebox{\linewidth}{!}{\begin{tabular}{llclllll}
\toprule
Dataset & prior & $n_{\rm ens}$ & $K=7$ $C_\star$/pair+1 primary & $K=7$ $C_\star$/pair+1 stress & pair+1/iid primary;stress & $K=5$ $C_\star$/pair+1 edge & pair+1/iid edge \\
\midrule
WDBC & 0.1 & 256 & 0.0361 [0.0344, 0.0378] & 0.0349 [0.0314, 0.0366] & 0.1528 / 0.1528 & 0.0254 [0.0245, 0.0261] & 0.2048 \\
WDBC & 0.3 & 256 & 0.0557 [0.0511, 0.0619] & 0.0559 [0.0510, 0.0677] & 0.1497 / 0.1517 & 0.0437 [0.0365, 0.0498] & 0.2104 \\
Spambase & 0.1 & 192 & 0.0110 [0.0101, 0.0124] & 0.0105 [0.0101, 0.0112] & 0.1488 / 0.1453 & 0.0076 [0.0075, 0.0091] & 0.2031 \\
Spambase & 0.3 & 192 & 0.0163 [0.0151, 0.0183] & 0.0185 [0.0150, 0.0268] & 0.1435 / 0.1448 & 0.0155 [0.0115, 0.0161] & 0.1952 \\
Adult & 0.1 & 64 & 0.0037 [0.0034, 0.0040] & 0.0038 [0.0034, 0.0043] & 0.1396 / 0.1404 & 0.0025 [0.0024, 0.0027] & 0.1957 \\
Adult & 0.3 & 64 & 0.0039 [0.0033, 0.0057] & 0.0040 [0.0035, 0.0051] & 0.1430 / 0.1450 & 0.0030 [0.0027, 0.0035] & 0.2034 \\
\bottomrule
\end{tabular}
}
\end{table}

\begin{table}[htbp]
\centering
\scriptsize
\caption{Odd-$K$ zero-sum-class check on the real posteriors.  Matched
trace-variance ratios against iid (median over three replicates).  Any
zero-sum design cancels the linear ensemble-mean noise: a block baseline of
sign pairs plus one three-chain simplex block matches $C_\star$ up to a
$\le 1.5\times$ factor (replicate ranges disjoint on three of the four
WDBC cells, direction favoring $C_\star$), while the conventional
equal-weight pair+1 construction
sits at its predicted $\approx 1/K$ floor.  $C_\star$ remains the
exchangeable member of the class with the smallest jointly Gaussian
quadratic penalty.}
\label{tab:oddk-zerosum-class}
\resizebox{.8\linewidth}{!}{\begin{tabular}{clllll}
\toprule
$K$ & Dataset & prior & $C_\star$/iid & pairs+triplet/iid & pair+1/iid \\
\midrule
7 & WDBC & 0.1 & 0.00546 [0.00502, 0.00587] & 0.00778 [0.00731, 0.00784] & 0.144 [0.144, 0.147] \\
7 & WDBC & 0.3 & 0.00863 [0.00820, 0.00966] & 0.0107 [0.0106, 0.0111] & 0.141 [0.139, 0.153] \\
7 & Spambase & 0.1 & 0.00152 [0.00151, 0.00169] & 0.00176 [0.00169, 0.00179] & 0.146 [0.137, 0.149] \\
7 & Spambase & 0.3 & 0.00240 [0.00222, 0.00252] & 0.00242 [0.00226, 0.00243] & 0.151 [0.150, 0.154] \\
7 & Adult & 0.1 & $5.46\times 10^{-4}$ [$4.69\times 10^{-4}$, $6.80\times 10^{-4}$] & $5.57\times 10^{-4}$ [$5.56\times 10^{-4}$, $5.70\times 10^{-4}$] & 0.144 [0.139, 0.148] \\
7 & Adult & 0.3 & $4.87\times 10^{-4}$ [$4.72\times 10^{-4}$, $5.42\times 10^{-4}$] & $6.70\times 10^{-4}$ [$5.84\times 10^{-4}$, $7.51\times 10^{-4}$] & 0.143 [0.128, 0.146] \\
\midrule
5 & WDBC & 0.1 & 0.00622 [0.00588, 0.00631] & 0.00715 [0.00644, 0.00817] & 0.207 [0.207, 0.208] \\
5 & WDBC & 0.3 & 0.0102 [0.00827, 0.0115] & 0.0101 [0.00981, 0.0115] & 0.223 [0.204, 0.237] \\
5 & Spambase & 0.1 & 0.00156 [0.00142, 0.00180] & 0.00167 [0.00158, 0.00175] & 0.206 [0.197, 0.206] \\
5 & Spambase & 0.3 & 0.00252 [0.00204, 0.00362] & 0.00239 [0.00223, 0.00302] & 0.204 [0.184, 0.207] \\
5 & Adult & 0.1 & $5.26\times 10^{-4}$ [$5.03\times 10^{-4}$, $5.55\times 10^{-4}$] & $4.95\times 10^{-4}$ [$4.91\times 10^{-4}$, $5.81\times 10^{-4}$] & 0.198 [0.195, 0.202] \\
5 & Adult & 0.3 & $5.09\times 10^{-4}$ [$4.72\times 10^{-4}$, $6.21\times 10^{-4}$] & $6.76\times 10^{-4}$ [$6.75\times 10^{-4}$, $6.87\times 10^{-4}$] & 0.198 [0.193, 0.211] \\
\bottomrule
\end{tabular}
}
\end{table}

\FloatBarrier
\subsection{Changing the transition with minibatches and MALA}\label{app:exp-transition}
Table~\ref{tab:minibatch-sgld-real} reports trace-variance robustness checks
for fixed-step nonquadratic minibatched SGLD on the UCI logistic posteriors.
At prior precision $\lambda$ we used a base step size of
$0.25\min\{0.03/(\lambda+0.25),0.08\}$. We ran WDBC and Spambase for
$3600$ steps and Adult for $2800$, discarding the first half in each
case. The main run used $256$ independent ensembles on WDBC, $192$ on
Spambase and $64$ on Adult, with respective batch sizes $8$, $64$ and
$512$. In the small-batch check we halved the batch size on each dataset.

\begin{table}[htbp]
\centering
\scriptsize
\caption{Which features of a minibatch run matter for the variance reduction?
All entries compare $K=8$ centered and iid Gaussian innovations.
We divide posterior-mean trace variances within a replicate, matching
the dataset, prior, horizon, step and batch sizes, and ensemble count,
then take the median of the ratios. The primary and stress columns
repeat the deterministic-start design with independent batches per
chain. Small-batch halves the indicated batch size. Random-start/WR
and common-stream both draw independent initial states at scale $0.25$
from a Gaussian and sample minibatches with replacement. Common-stream
reuses the minibatch random stream between methods in the matched design.}
\label{tab:minibatch-sgld-real}
\resizebox{\linewidth}{!}{\begin{tabular}{llrrrrrr}
\toprule
Dataset & prior & batch & primary & stress & small-batch & random start/WR & common stream \\
\midrule
WDBC & 0.1 & 8 & 0.0055 & 0.0052 & 0.0056 & 0.0059 & 0.0058 \\
WDBC & 0.3 & 8 & 0.0092 & 0.0088 & 0.0081 & 0.0076 & 0.0091 \\
\addlinespace
Spambase & 0.1 & 64 & 0.0016 & 0.0016 & 0.0015 & 0.0015 & 0.0016 \\
Spambase & 0.3 & 64 & 0.0024 & 0.0027 & 0.0026 & 0.0028 & 0.0025 \\
\addlinespace
Adult & 0.1 & 512 & $5.35{\times}10^{-4}$ & $5.23{\times}10^{-4}$ & $5.49{\times}10^{-4}$ & $5.81{\times}10^{-4}$ & $6.00{\times}10^{-4}$ \\
Adult & 0.3 & 512 & $5.31{\times}10^{-4}$ & $5.66{\times}10^{-4}$ & $6.35{\times}10^{-4}$ & $6.23{\times}10^{-4}$ & $8.17{\times}10^{-4}$ \\
\bottomrule
\end{tabular}
}
\end{table}

\paragraph{Metropolis adjustment.} Each chain makes a ULA proposal
and applies its own Metropolis--Hastings test. To choose the step size
for a dataset and prior, we ran a 100-step iid pilot on a geometric grid
around the ULA default. We took the largest step with acceptance in
$[0.4,0.8]$. If the entire grid exceeded the band, we took the step
whose acceptance was closest to $0.6$. Both kernels and all couplings
then used that same step size. Realized acceptance across the cells
ranged from $0.67$ to $0.89$.

\FloatBarrier
\begin{table}[htbp]
\centering
\small
\caption{How much of the ULA gain survives MALA at the same step size?
Each entry compares posterior-mean trace variances at $K=8$ and gives
the median and range over three matched replicates. Within a cell
every kernel and coupling uses the same step size. Each coupled MALA
chain still has its ordinary MALA law. If only some proposals are accepted, the Gaussian draws attached to
those proposals can have a nonzero sum. Centering
then gives a three- to eightfold reduction relative to iid MALA,
smaller than the reduction for ULA in the same cells.}
\label{tab:mala-real}
\resizebox{.9\linewidth}{!}{\begin{tabular}{llrrrr}
\toprule
Dataset & prior & ULA $C_\star$/iid & MALA $C_\star$/iid & MALA pairwise/iid & MALA accept \\
\midrule
WDBC & 0.1 & 0.0443 [0.0376, 0.0454] & 0.168 [0.140, 0.178] & 0.147 [0.139, 0.161] & 0.89 \\
WDBC & 0.3 & 0.0544 [0.0527, 0.0653] & 0.267 [0.256, 0.328] & 0.247 [0.246, 0.297] & 0.74 \\
Spambase & 0.1 & 0.00816 [0.00716, 0.00953] & 0.130 [0.129, 0.141] & 0.113 [0.109, 0.127] & 0.87 \\
Spambase & 0.3 & 0.0205 [0.0203, 0.0212] & 0.337 [0.333, 0.342] & 0.245 [0.221, 0.268] & 0.67 \\
Adult & 0.1 & 0.0170 [0.0136, 0.0225] & 0.155 [0.154, 0.236] & 0.155 [0.130, 0.204] & 0.81 \\
Adult & 0.3 & 0.0378 [0.0265, 0.0467] & 0.163 [0.155, 0.189] & 0.146 [0.137, 0.169] & 0.84 \\
\bottomrule
\end{tabular}
}
\end{table}

\begin{table}[htbp]
\centering
\small
\caption{How sensitive is the MALA comparison to the pilot step size?
Across two replicates the variance ratio for $C_\star$ reached $0.86$
at an acceptance rate of $0.24$. At four times the pilot step most
cells rejected nearly every proposal. We excluded rates below $0.15$
from the table. In one such Adult cell, with acceptance near $0.11$,
the replicate ratios spanned $0.77$--$1.06$. The pairwise ratios were
lower than those for $C_\star$ throughout the tabulated comparison.}
\label{tab:mala-sensitivity}
\resizebox{.85\linewidth}{!}{\begin{tabular}{llrrrr}
\toprule
Dataset & prior & $\eta/\eta_{\mathrm{cal}}$ & MALA accept & MALA $C_\star$/iid & MALA pairwise/iid \\
\midrule
WDBC & 0.1 & 2$\times$ & 0.70 & 0.362 [0.310, 0.413] & 0.273 [0.262, 0.285] \\
WDBC & 0.1 & 4$\times$ & 0.24 & 0.855 [0.806, 0.904] & 0.628 [0.565, 0.692] \\
WDBC & 0.3 & 2$\times$ & 0.32 & 0.700 [0.624, 0.776] & 0.494 [0.449, 0.540] \\
Spambase & 0.1 & 2$\times$ & 0.61 & 0.362 [0.353, 0.372] & 0.270 [0.257, 0.284] \\
Adult & 0.3 & 2$\times$ & 0.38 & 0.762 [0.650, 0.873] & 0.677 [0.609, 0.746] \\
\bottomrule
\end{tabular}
}
\end{table}

\FloatBarrier
\subsection{Accuracy against a long MALA reference}\label{app:exp-reference}

For the archived reference we ran $K=64$ MALA chains per cell from a
common zero start. A pilot selected the step size (long-run acceptance
$0.56$--$0.79$). We discarded the first $5{,}000$ steps, kept the next
$30{,}000$ and pooled three fixed seeds. Their spread gives a check on Monte Carlo variability, but all the
chains could still carry the same bias from starting at zero. For the
archived reference we checked agreement with independent VI and EP
fits (Table~\ref{tab:gold-endtask-archived}). For the main reference
we could check the start directly. We compared $K=64$ chains from a
Cholesky-overdispersed start with the same design from zero, using three
seeds. Both arms had to pass the acceptance and split-half equivalence
checks as well as split-$\widehat R\le1.01$ and bulk ESS $\ge400$.
A third dispersed arm was descriptive. Adult/$0.3$ passed on the one
doubled-length repeat allowed by the protocol. We report the Bonferroni
batch-means simultaneous $95\%$ radius $r_{95}$ with $q_{P4}=3.829$.

For the end-task comparison we used fresh prespecified seeds and scored
each retained estimate against both references. This pairing holds the
estimates fixed when measuring the effect of changing the reference.
We matched the total gradient budgets ($2{,}400$ steps at $K=8$, or
$1{,}800$ on Adult) and discarded the first half.

We found the MAP with full-Hessian Newton steps in float64 on the
same potential. The gradient norm reached $10^{-10}$, and we checked
the gradient against an independent implementation.

For VI we minimized the reverse KL over full-rank Gaussians
$q=N(\mu,LL^\top)$. We evaluated $\mathbb E_q[\mathrm{softplus}]$
by integrating a hinge term analytically and the bounded remainder on
fixed Gauss--Legendre panels. This gives machine precision even for the
largest marginal scale here ($s_i\approx460$). We checked the kernels
against adaptive quadrature to $10^{-10}$. For log-concave targets the
negative ELBO is jointly convex in $(\mu,L)$~\cite{challis2013gaussian}.
The unpenalized intercept preserves convexity, and with both labels
present the likelihood makes $U$ strictly convex, giving a unique
optimum. That argument uses the entries of $L$ directly. We optimize
its log-diagonal parametrization, so we also checked convergence
numerically. Float64 L-BFGS fits from a Laplace start and a cold start
agreed to $10^{-8}$, the gradient norm was below $10^{-6}$, and
doubling the panels changed the result by less than $10^{-10}$.

For EP the site factor at $z_i=x_i^\top\theta$ was
$\exp(-(\mathrm{softplus}(z_i)-y_iz_i)/n)$. We computed tilted moments
with the same panels and integrated the truncated-Gaussian tails in
closed form. We initialized damped parallel updates at the Laplace site
decomposition. At convergence we required a relative site residual
below $10^{-8}$, zero skipped cavities, nonnegative site precisions
and a stable mean on the same sweep.

\begin{table}[htbp]
\centering
\small
\caption{Squared error against the long MALA reference from dispersed starts.  Entries are median [min--max] posterior-mean
squared error $\|\widehat\theta-\mu_{\mathrm{gold}}\|^2$ over three
fresh-seed replicates at gradient-evaluation budgets matched across
the sampler arms; VI, EP, and MAP are deterministic fits rescored
against the same reference.  The last column is the per-cell
simultaneous $95\%$ gold-error radius $r_{95}$ (Bonferroni batch-means
construction).  MAP is the deterministic mode/Laplace-mean baseline.
It misses the posterior mean
badly because these tempered-likelihood posteriors are wide (Laplace
trace $97$--$552$) and skewed, and the gap does not hinge on the
unpenalized intercept.  VI and EP are deterministic converged fits,
not budget-matched estimators: more budget improves the sampler columns
(ULA toward its fixed-step bias floor) but cannot improve a fixed fit,
and a fixed fit carries no internal estimate of its posterior-mean
error. The two fits solve different fixed-point problems; their reference squared
errors differ by less than $2\%$ in every cell, an implementation check. To order two entries we required their squared errors to differ by a
factor of at least $1.5$ and each to exceed $25r_{95}^2$. The latter
condition puts both distances beyond five reference-error radii. The $95\%$ level describes the simultaneous
reference radii.  By that rule every
sampler-versus-VI/EP ordering is floor-unresolved in every cell (all
$48$ pairs; Table~\ref{tab:gold-ordering-grid}); unresolved is not
equivalent to matched---no equivalence test was passed.  VI/EP raw
squared errors are numerically lower than $C_\star$ ULA on all six
cells ($1.9$--$6.5\times$), descriptively.  Sampler-versus-sampler
differences are likewise descriptive, at reference precision: every
$C_\star$ ULA cell median sits below its cell's $25r_{95}^2$ floor,
while every iid ULA cell median sits above it.  $C_\star$ ULA uses
step $4\eta_c$ and the fine column $2\eta_c$; at the matched budget
the finer step covers half the time horizon, so its larger Adult error
is unconverged-transient error, not a step-size pathology.  The MALA
column is MALA with centered proposal noise at the pilot-calibrated step and equal
step count: with the current-point gradient cached, both kernels spend
one fresh gradient evaluation per chain per step, and MALA's fused
acceptance test makes its measured per-step wall-clock
$1.8$--$3.1\times$ ULA's, uncharged here.  ULA's measured stationary-bias estimate at
$4\eta_c$ is within $r_{95}$ (squared bias $<r_{95}^2$) in every fresh
replicate.  The archived common-start reference and the original
sampler seeds are reported in Table~\ref{tab:gold-endtask-archived}. The Spambase/$0.3$ ordering did not replicate with fresh sampler seeds ($1.22$ against the $1.5$ threshold).}
\label{tab:gold-endtask}
\resizebox{\linewidth}{!}{\begin{tabular}{llrrrrrrrr}
\toprule
Dataset & prior & MAP & VI & EP & iid ULA & $C_\star$ ULA & $C_\star$ ULA (fine) & MALA & $r_{95}$ \\
\midrule
WDBC & 0.1 & 26.3 & 0.015 & 0.0148 & 2.79 [2.75, 2.85] & 0.0654 [0.0458, 0.0863] & 0.108 [0.099, 0.112] & 0.19 [0.184, 0.224] & 0.125 \\
WDBC & 0.3 & 8.59 & 0.0169 & 0.0167 & 0.845 [0.751, 0.861] & 0.0352 [0.0327, 0.0356] & 0.045 [0.0376, 0.0466] & 0.103 [0.101, 0.133] & 0.0948 \\
Spambase & 0.1 & 25.1 & 0.00821 & 0.00815 & 3.73 [3.61, 3.81] & 0.0165 [0.0148, 0.018] & 0.0275 [0.0252, 0.0423] & 0.226 [0.214, 0.238] & 0.128 \\
Spambase & 0.3 & 6.67 & 0.00307 & 0.00304 & 0.913 [0.846, 0.944] & 0.0135 [0.0105, 0.0144] & 0.0155 [0.0143, 0.0199] & 0.0933 [0.0928, 0.0961] & 0.102 \\
Adult & 0.1 & 245 & 0.0165 & 0.0165 & 10.2 [9.71, 10.9] & 0.107 [0.0773, 0.135] & 1.63 [1.54, 1.64] & 0.531 [0.501, 0.532] & 0.247 \\
Adult & 0.3 & 85.9 & 0.0418 & 0.0418 & 2.52 [2.38, 3.05] & 0.0783 [0.054, 0.0979] & 0.576 [0.495, 0.634] & 0.335 [0.314, 0.436] & 0.152 \\
\bottomrule
\end{tabular}
}
\end{table}

\begin{table}[htbp]
\centering
\small
\caption{Archived common-start reference (historical; superseded as
the main comparison by Table~\ref{tab:gold-endtask}).  Original-seed entries are
the previously reported table; fresh-seed entries score the
rescore-run estimates against the same archived reference.  Under the
original seeds the comparison rule called $C_\star$ ULA
$1.7\times$ closer than VI/EP on Spambase/$0.3$; this is a historical,
non-replicating result: with fresh sampler seeds against the same
archived reference the $C_\star$ error moves from $0.00989$ to
$0.0139$ and the ratios drop to $1.22\times$ (VI) and $1.22\times$
(EP), below the $1.5\times$ rule---sampler seed noise, not the
reference change (the paired dispersed-start/archived error ratio on that
cell is $0.975$ (from unrounded errors; the displayed values give
$0.971$)).  Rescoring the fits against the dispersed-start reference
lowers VI/EP on five of six cells (Spambase/$0.3$:
$0.0170\to0.0031$) and raises them on Adult/$0.3$
($0.0240\to0.0418$): the archived common-start transient inflated the
Gaussian fits' apparent error.  Under the original seeds ULA's
measured stationary-bias estimate at $4\eta_c$ is below the archived
across-seed spread (last column) in every replicate of every cell.
The across-seed spread bounds Monte Carlo variability but not
transient bias shared through the common start, for which the
independent VI/EP agreement was the corroborating check---the
dispersed-start reference of Table~\ref{tab:gold-endtask} now
supersedes that role.}
\label{tab:gold-endtask-archived}
\resizebox{\linewidth}{!}{\begin{tabular}{llrrrrrrrr}
\toprule
Dataset & prior & MAP & VI & EP & iid ULA & $C_\star$ ULA & $C_\star$ ULA (fine) & MALA & gold spread \\
\midrule
\multicolumn{10}{l}{\emph{Original sampler seeds (as previously reported)}} \\
WDBC & 0.1 & 26.4 & 0.0181 & 0.0178 & 2.75 [2.57, 2.94] & 0.0608 [0.047, 0.076] & 0.105 [0.0963, 0.118] & 0.189 [0.166, 0.211] & 4.8e-03 \\
WDBC & 0.3 & 8.68 & 0.0211 & 0.0209 & 0.874 [0.747, 0.931] & 0.0327 [0.0299, 0.0412] & 0.0638 [0.0527, 0.0658] & 0.109 [0.102, 0.128] & 1.0e-02 \\
Spambase & 0.1 & 25.2 & 0.0102 & 0.0101 & 3.54 [3.53, 3.74] & 0.0185 [0.0129, 0.0229] & 0.0287 [0.0271, 0.0347] & 0.223 [0.204, 0.227] & 7.2e-03 \\
Spambase & 0.3 & 7.04 & 0.017 & 0.0169 & 1.01 [0.827, 1.05] & 0.00989 [0.0067, 0.0102] & 0.0209 [0.0185, 0.0243] & 0.123 [0.107, 0.125] & 1.8e-03 \\
Adult & 0.1 & 247 & 0.0308 & 0.0307 & 10.2 [9.49, 10.4] & 0.121 [0.0624, 0.132] & 1.67 [1.63, 1.76] & 0.568 [0.543, 0.653] & 4.4e-02 \\
Adult & 0.3 & 85 & 0.024 & 0.0239 & 2.71 [2.51, 2.97] & 0.0702 [0.0483, 0.0783] & 0.464 [0.418, 0.556] & 0.387 [0.361, 0.391] & 2.3e-02 \\
\midrule
\multicolumn{10}{l}{\emph{Fresh rescore seeds, scored against the same archived reference}} \\
WDBC & 0.1 & 26.4 & 0.0181 & 0.0178 & 2.79 [2.78, 2.85] & 0.0673 [0.0448, 0.0856] & 0.113 [0.102, 0.116] & 0.188 [0.186, 0.223] & 4.8e-03 \\
WDBC & 0.3 & 8.68 & 0.0211 & 0.0209 & 0.842 [0.771, 0.846] & 0.0358 [0.0344, 0.0389] & 0.045 [0.0396, 0.0459] & 0.108 [0.0963, 0.134] & 1.0e-02 \\
Spambase & 0.1 & 25.2 & 0.0102 & 0.0101 & 3.75 [3.61, 3.83] & 0.0157 [0.0147, 0.0185] & 0.0264 [0.024, 0.044] & 0.226 [0.217, 0.239] & 7.2e-03 \\
Spambase & 0.3 & 7.04 & 0.017 & 0.0169 & 0.936 [0.85, 0.978] & 0.0139 [0.0133, 0.0163] & 0.0165 [0.0164, 0.0215] & 0.093 [0.0903, 0.0999] & 1.8e-03 \\
Adult & 0.1 & 247 & 0.0308 & 0.0307 & 10.2 [9.56, 11] & 0.119 [0.104, 0.128] & 1.76 [1.67, 1.77] & 0.526 [0.51, 0.53] & 4.4e-02 \\
Adult & 0.3 & 85 & 0.024 & 0.0239 & 2.52 [2.37, 2.95] & 0.0734 [0.0522, 0.0745] & 0.501 [0.426, 0.577] & 0.328 [0.321, 0.433] & 2.3e-02 \\
\bottomrule
\end{tabular}
}
\end{table}

\begin{table}[htbp]
\centering
\small
\caption{Ordering labels for all $48$ sampler-versus-fit pairs under
both references (entry: common start $\to$ dispersed starts).  Labels: U =
floor-unresolved (at least one error below the floor); N = neither
ordered at $1.5\times$ (floors passed, ratio $<1.5$); VI/EP = that
baseline ordered ahead under the comparison rule.  Archived floor:
$5s_c$ on squared errors ($s_c$ the archived across-seed spread);
dispersed-start threshold: $25r_{95}^2$ (the five-$r_{95}$ distance rule).  All
$48$ pairs are floor-unresolved under the dispersed-start reference; the
eight paired label changes are confined to Spambase/$0.3$, of which
four (iid and MALA versus VI and EP) are directional and fire the
pre-specified disclosure rule; the two original$\to$fresh $C_\star$
changes on the same cell are a separate sampler-noise category (see
Table~\ref{tab:gold-endtask-archived}).  Paired dispersed-start/archived
sampler-error ratios are $0.899$--$1.149$ across arms and cells: the
reference change is benign for sampler-error magnitudes---a statement
about magnitudes, never ordering labels.  The rule could also identify a sampler as more accurate than a fit.
That outcome did not occur under either reference.}
\label{tab:gold-ordering-grid}
\resizebox{\linewidth}{!}{\begin{tabular}{llcccccccc}
\toprule
Dataset & prior & iid--VI & iid--EP & $C_\star\eta_1$--VI & $C_\star\eta_1$--EP & $C_\star\eta_2$--VI & $C_\star\eta_2$--EP & MALA--VI & MALA--EP \\
\midrule
WDBC & 0.1 & U$\to$U & U$\to$U & U$\to$U & U$\to$U & U$\to$U & U$\to$U & U$\to$U & U$\to$U \\
WDBC & 0.3 & U$\to$U & U$\to$U & U$\to$U & U$\to$U & U$\to$U & U$\to$U & U$\to$U & U$\to$U \\
Spambase & 0.1 & U$\to$U & U$\to$U & U$\to$U & U$\to$U & U$\to$U & U$\to$U & U$\to$U & U$\to$U \\
Spambase & 0.3 & VI$\to$U & EP$\to$U & N$\to$U & N$\to$U & N$\to$U & N$\to$U & VI$\to$U & EP$\to$U \\
Adult & 0.1 & U$\to$U & U$\to$U & U$\to$U & U$\to$U & U$\to$U & U$\to$U & U$\to$U & U$\to$U \\
Adult & 0.3 & U$\to$U & U$\to$U & U$\to$U & U$\to$U & U$\to$U & U$\to$U & U$\to$U & U$\to$U \\
\bottomrule
\end{tabular}
}
\end{table}

\FloatBarrier
\subsection{Preconditioning and Hamiltonian transitions}\label{app:exp-hmc}

\paragraph{Fixed preconditioner.} The fixed diagonal preconditioner is the
Jacobi/Gauss--Newton curvature at zero,
$P=\operatorname{diag}(1/(0.25\,\overline{x_j^2}+\lambda))$, applied as
$x\leftarrow x-\eta P\nabla U+\sqrt{2\eta}\,P^{1/2}\xi$ with the coupling
on $\xi$; a per-cell pilot picks the largest stable step.

\paragraph{Momentum refresh.} We ran HMC for $300$ outer steps and
discarded the first $150$. Each trajectory took $L=10$ leapfrog steps.
All couplings used the same pilot-calibrated step size within a cell and
shared noise only when refreshing the momenta.

\paragraph{Path lengths.} In jittered HMC we drew
$L\sim\mathrm{Uniform}\{1,\dots,20\}$, giving an average of $10.5$
leapfrog steps against ten in the fixed-length run. One version draws
$L$ once per ensemble at each outer step, so all its chains have the
same path length. The other draws $L$ separately for each chain. Both
mix exact HMC kernels and both average over schedules across ensembles.
We seeded the schedules by design to match them across couplings while
keeping replicates independent.

For multinomial NUTS each chain builds its own tree, with maximum
depth $8$. The chains share noise only in the momentum refresh.
We accumulate energies in float64, use biased progressive sampling
across doublings and check the original endpoint U-turn criterion on
every aligned power-of-two block by iterative checkpointing.
We checked the sampler on a Gaussian with known moments and found
errors within $0.04\sigma$ against a tolerance of $4\sigma$. For the
selection distribution the largest quantile discrepancy was
$|z|=1.39$. With constant or mixed $L$, the per-chain jittered
integrator agreed numerically with plain leapfrog at every check.
The code repository contains the self-test log. Each of the three kernels
used the pilot step size for its cell and its own long iid run for
the reference mean. At the tabulated step sizes we observed neither a
divergence nor a tree reaching the maximum depth.

\begin{table}[htbp]
\centering
\small
\caption{Fixed diagonal (Jacobi/Gauss--Newton) preconditioned ULA at
$K=8$.  Preconditioning by a fixed matrix is a coordinate change, so the
zero-sum identity transfers verbatim; the measured matched trace-variance
ratios stay in the unpreconditioned band.  On these standardized datasets
the Jacobi preconditioner is nearly isotropic; strongly anisotropic and
adaptive preconditioning remain open.}
\label{tab:preconditioned-real}
\resizebox{.65\linewidth}{!}{\begin{tabular}{llrr}
\toprule
Dataset & prior & $C_\star$/iid & pairwise/iid \\
\midrule
WDBC & 0.1 & 0.0472 [0.0462, 0.0599] & 0.0538 [0.0513, 0.0656] \\
WDBC & 0.3 & 0.0374 [0.0290, 0.0423] & 0.0471 [0.0452, 0.0484] \\
Spambase & 0.1 & 0.00806 [0.00761, 0.00837] & 0.00976 [0.00911, 0.0105] \\
Spambase & 0.3 & 0.0150 [0.0134, 0.0170] & 0.0164 [0.0148, 0.0254] \\
Adult & 0.1 & 0.0222 [0.0187, 0.0310] & 0.0314 [0.0304, 0.0382] \\
Adult & 0.3 & 0.0391 [0.0387, 0.0396] & 0.0681 [0.0490, 0.0703] \\
\bottomrule
\end{tabular}
}
\end{table}

\begin{table}[htbp]
\centering
\small
\caption{Coupled HMC at $K=8$ ($L=10$ leapfrog steps, calibrated step
sizes; only the momentum refresh is coupled).  Matched trace-variance
ratios (median [min--max] over three replicates).  The reduction transfers
in five of six cells; the rejection-leak prediction $p\,s$ becomes a floor
rather than a match because trajectory endpoints are nonlinear in the
refreshed momenta, and the Spambase/$0.3$ cell lands on a leapfrog
resonance where iid is anomalously efficient and the ratio reaches parity
(one step-size notch down restores $0.17$).}
\label{tab:hmc-real}
\resizebox{.85\linewidth}{!}{\begin{tabular}{llrrrr}
\toprule
Dataset & prior & HMC accept & $C_\star$/iid & pairwise/iid & pred.\ $p\,s$ \\
\midrule
WDBC & 0.1 & 0.82 & 0.449 [0.424, 0.462] & 0.487 [0.428, 0.541] & 0.200 \\
WDBC & 0.3 & 0.88 & 0.214 [0.205, 0.229] & 0.214 [0.214, 0.232] & 0.131 \\
Spambase & 0.1 & 0.78 & 0.316 [0.313, 0.345] & 0.281 [0.276, 0.298] & 0.244 \\
Spambase & 0.3 & 0.78 & 0.993 [0.875, 1.03] & 0.967 [0.768, 1.04] & 0.230 \\
Adult & 0.1 & 0.90 & 0.323 [0.260, 0.353] & 0.358 [0.262, 0.392] & 0.111 \\
Adult & 0.3 & 0.89 & 0.172 [0.167, 0.176] & 0.140 [0.138, 0.145] & 0.114 \\
\bottomrule
\end{tabular}
}
\end{table}

\begin{table}[htbp]
\centering
\small
\caption{Adaptive path lengths at $K=8$ (matched trace-variance ratios,
median [min--max] over three replicates; per-cell step sizes shared with
Table~\ref{tab:hmc-real}).  With the jittered trajectory length shared by
an ensemble's chains---what parallel-chain tuners such as
ChEES~\cite{hoffman2021chees} produce---the reduction survives in five of
six cells (two cells transfer at $\le 0.5$, three are partial at
$0.51$--$0.67$; the leapfrog-resonance cell Spambase/$0.3$ sits at
parity, $0.807$).  Per-chain lengths sit at parity in all six cells
($0.91$--$1.01$), and per-chain multinomial NUTS (endpoint U-turn
criterion, mean tree depth $2.96$--$3.98$, accept-statistic
$0.61$--$0.90$, zero divergences, zero max-depth saturation) in five of
six with Adult/$0.3$ partial at $0.80$: the cancellation requires the
same integrator map across chains, not merely coupled refreshes.  The
pre-specified paired shared-vs-per-chain contrast is positive in all
six cells (median $0.38$).  Replicate ranges straddle a band boundary in
six tabulated entries; in particular one Adult/$0.1$ NUTS replicate
reaches $1.25$, the pre-specified ``hurts'' threshold, so NUTS can
mildly amplify variance there.  Pairwise antithetic tracks at or below
$C_\star$ in most shared-$L$ and NUTS cells, consistent with the paper's
even-$K$ disclosures.  Closed-loop ChEES adaptation on the coupled
chains is tested separately (Table~\ref{tab:chees-closedloop}); all
schedules here are exogenous and design-seeded.}
\label{tab:nuts-real}
\resizebox{.8\linewidth}{!}{\begin{tabular}{lllrr}
\toprule
Dataset & prior & kernel & $C_\star$/iid & pairwise/iid \\
\midrule
WDBC & 0.1 & jitter (shared $L$) & 0.474 [0.468, 0.518] & 0.481 [0.456, 0.496] \\
WDBC & 0.1 & jitter (per-chain $L$) & 0.924 [0.910, 0.986] & 1.00 [0.932, 1.01] \\
WDBC & 0.1 & NUTS & 1.12 [0.859, 1.14] & 0.919 [0.803, 1.07] \\
\addlinespace
WDBC & 0.3 & jitter (shared $L$) & 0.599 [0.554, 0.604] & 0.520 [0.503, 0.531] \\
WDBC & 0.3 & jitter (per-chain $L$) & 0.912 [0.876, 0.989] & 0.990 [0.973, 1.03] \\
WDBC & 0.3 & NUTS & 0.957 [0.884, 0.974] & 0.898 [0.895, 0.919] \\
\addlinespace
Spambase & 0.1 & jitter (shared $L$) & 0.514 [0.507, 0.523] & 0.435 [0.433, 0.443] \\
Spambase & 0.1 & jitter (per-chain $L$) & 1.01 [0.991, 1.03] & 0.969 [0.957, 1.02] \\
Spambase & 0.1 & NUTS & 1.04 [0.875, 1.04] & 0.973 [0.903, 1.02] \\
\addlinespace
Spambase & 0.3 & jitter (shared $L$) & 0.807 [0.745, 0.854] & 0.655 [0.636, 0.682] \\
Spambase & 0.3 & jitter (per-chain $L$) & 0.988 [0.920, 1.01] & 1.01 [0.990, 1.03] \\
Spambase & 0.3 & NUTS & 0.959 [0.927, 0.985] & 0.975 [0.928, 1.04] \\
\addlinespace
Adult & 0.1 & jitter (shared $L$) & 0.295 [0.290, 0.321] & 0.273 [0.261, 0.284] \\
Adult & 0.1 & jitter (per-chain $L$) & 0.955 [0.886, 0.986] & 0.960 [0.911, 0.985] \\
Adult & 0.1 & NUTS & 1.04 [0.901, 1.25] & 1.13 [0.878, 1.21] \\
\addlinespace
Adult & 0.3 & jitter (shared $L$) & 0.673 [0.574, 0.689] & 0.536 [0.489, 0.550] \\
Adult & 0.3 & jitter (per-chain $L$) & 0.955 [0.954, 0.958] & 0.980 [0.940, 1.04] \\
Adult & 0.3 & NUTS & 0.796 [0.733, 0.806] & 0.697 [0.604, 0.925] \\
\bottomrule
\end{tabular}
}
\end{table}

\begin{table}[htbp]
\centering
\scriptsize
\caption{Tuning ChEES on the coupled chains failed the stability
criterion in two of six logistic cells at $K=8$. The $C_\star$ arm's
median IQR$_e(\log_2 T^\star)$ exceeded $1.0$ on Spambase/$0.3$
($1.086$) and Adult/$0.3$ ($1.494$). The pairwise arm also gave a large
IQR on Adult/$0.3$ ($1.377$), compared with $0.168$--$0.921$ across iid
cells. These differences concern the dispersion of the adapted schedules.
Their median locations met the practical-equivalence criteria
($A_T=0.879$--$1.007$, $A_\epsilon=0.989$--$1.002$).
The first two ratios compare the closed-loop trace variances with iid
(median [range] over three replicates). All six $C_\star$ cell medians
met the specified upper limit of $0.81$ ($0.412$--$0.781$), although one
Adult/$0.3$ replicate gave $0.870$. RS(A) and RS(B) use fresh runs to
replay frozen schedules learned with iid and coupled chains, respectively.
Their ratios were $0.386$--$0.539$ and $0.352$--$0.538$.
The replay-source differences $M_{\mathrm{src}}=0.009$--$0.273$ and
$\mathrm{INT}=-0.260$--$0.114$ do not establish equivalence.
We adapted $\log T$ for $450$ steps with Adam (learning rate $0.025$,
$\beta_2=0.95$), dual averaging to harmonic-mean acceptance $0.651$ and
rotated-Halton jitter. We froze the EMA/dual-averaged schedule, allowed
$50$ more steps for equilibration and scored only the subsequent steps.
At the recorded dynamics, cross-fitted centering shifted all $54$ adapted
lengths upward ($+0.100$ to $+0.512$, median $+0.194$). This calculation
does not identify the effect of changing the closed loop. All $144$
scored rows passed the prescribed checks, including acceptance
$0.758$--$0.793$, zero divergences and zero clamps. The comparison with
the reference and probe met its criterion and warmup events stayed below
their specified limits. With three replicates, medians near a cutoff
remain sensitive to individual runs. These $650$-step runs use a
different schedule from the $300$-step runs in Table~\ref{tab:nuts-real}.}

\label{tab:chees-closedloop}
\resizebox{\linewidth}{!}{\begin{tabular}{llrrrrr}
\toprule
Dataset & prior & closed $C_\star$/iid & closed pairwise/iid & RS(A) & RS(B) & $A_T$ \\
\midrule
WDBC & 0.1 & 0.537 [0.420, 0.573] & 0.521 [0.464, 0.611] & 0.524 [0.485, 0.545] & 0.494 [0.432, 0.515] & 0.975 [0.968, 0.999] \\
WDBC & 0.3 & 0.601 [0.498, 0.622] & 0.496 [0.455, 0.543] & 0.539 [0.517, 0.562] & 0.538 [0.498, 0.552] & 1.01 [0.974, 1.01] \\
Spambase & 0.1 & 0.412 [0.384, 0.453] & 0.328 [0.296, 0.352] & 0.470 [0.403, 0.498] & 0.426 [0.421, 0.456] & 0.972 [0.972, 0.983] \\
Spambase & 0.3 & 0.522 [0.509, 0.545] & 0.421 [0.409, 0.588] & 0.401 [0.356, 0.450] & 0.377 [0.327, 0.399] & 0.904 [0.867, 0.918] \\
Adult & 0.1 & 0.518 [0.408, 0.588] & 0.393 [0.347, 0.448] & 0.428 [0.349, 0.492] & 0.471 [0.398, 0.480] & 0.916 [0.889, 0.945] \\
Adult & 0.3 & 0.781 [0.499, 0.870] & 0.558 [0.487, 0.624] & 0.386 [0.381, 0.548] & 0.352 [0.294, 0.411] & 0.879 [0.684, 0.966] \\
\bottomrule
\end{tabular}
}
\end{table}

\FloatBarrier
\subsection{Changing the posterior and measuring
uncertainty}\label{app:exp-posterior}

For five checks we retain the main campaign's estimator of trace variance over
independent ensembles, with $K=8$ chains started at zero and three replicates.
These are the Student-$t(3)$ and hierarchical random-intercept models, MNIST
(7 versus 9), Gisette, and the uncertainty/prediction observables. Run lengths
remain 900 steps (450 burn-in) for WDBC and Spambase, and 700 (350 burn-in)
for Adult, the high-dimensional runs and the hierarchical model. Independent
ensemble counts also stay fixed: 256 for WDBC, 192 for Spambase, 64 for Adult
and MNIST, 32 for Gisette. MALA, exact-regime, HMC and GP checks use the
separate protocols given in this appendix.

\paragraph{Regression priors.} We take $s^2=(3\lambda)^{-1}$ for the
Student-$t$ prior with 3 degrees of freedom. Its variance is thus
$\lambda^{-1}$, as for the Gaussian prior, and the intercept is again
unpenalized. The hierarchical variant is an Adult logistic regression with an
unpenalized global intercept and $G=16$ group intercepts, one for each
education group. In addition to the Gaussian coefficient prior
($\lambda=0.1$), its specification is

\[ u_g\sim N(0,\tau^2), \qquad \tau=e^\rho, \qquad \rho\sim N(0,1). \]

We evaluate gradients analytically and validate them by finite differences.

\paragraph{Gaussian process hyperparameters.} The RBF-kernel GP provides a
computational stress test because gradient evaluation involves a dense solve.
A fixed random permutation selects $n=320$ Superconductivity rows. We sample
$\theta=(\log\ell,\log\sigma_f,\log\sigma_n)$, using standard normal priors, a
fixed nugget $10^{-6}$ and the tempered potential

\[ U(\theta)=-\frac{\log p(y\mid X,\theta)}{n}-\log p(\theta). \]

We check the potential against an independent implementation and the $n=1$
closed form. The analytic trace gradient is checked by finite differences
about zero and the MAP. In the float64 GP potential evaluator, a Cholesky
factorization checks the kernel matrix and gives its log determinant; we
obtain the inverse by a separate solve and require its entrywise residual to
be below $10^{-6}$.

For ULA, we use 24 independent ensembles of $K=8$ chains per replicate, with
2,400 steps and 1,200 burn-in. The gradient is clipped at norm 10 before
adding noise, since $\log\sigma_n$ can be a stiff direction. The clipped
marginal transition is identical for all couplings. We chose the largest grid
step with pilot MALA acceptance at least 0.75, no singular events, and finite
ULA-probe second moments below 100 in every coordinate: $\eta=0.4$. Applying
the unclipped reference MALA acceptance band $[0.55,0.85]$ to the same pilot
also selects this step.

No singular events occurred in the scored ULA runs. Clipping rates were
0.73--0.77\%, and the maximum solve residual was $1.3\times10^{-8}$. The
repository does not retain clip counts for the step-size and invariance probes,
although they were computed. The maximum absolute invariance $z$-statistic
was checked against 4, but its value was not saved.

The reference calculation used $K=64$ with 30,000 kept steps. Each of three
seeds gave acceptance 0.798. No coordinate had skewness outside
$[-0.11,0.11]$; the largest squared Euclidean distance between seed means was
$1.5\times10^{-5}$, and the squared distance between pooled first-half and
second-half means was $4.1\times10^{-6}$.

\paragraph{Posterior uncertainty.} The uncertainty observables are
coordinate-wise variance and equal-tail 90\% interval width. Both are computed
from all post-burn-in draws in an independent ensemble, pooled over chains and
time. Using these same draws, we compute a sigmoid average at each of 128
fixed test points (114 for WDBC). We estimate the variance of each estimate
across ensembles, then sum over coordinates or test points.

\begin{table}[htbp]
\centering
\small
\caption{Non-GLM cell: GP-regression hyperparameter posterior on
Superconductivity ($d=3$, tempered likelihood, dense-solve gradients;
clipped ULA at $K=8$, matched budgets, three replicates).  The
coupling's variance reduction transfers---trace-variance ratios
$0.036$--$0.070$ for $C_\star$ and $0.041$--$0.065$ for pairwise, the
pre-specified ``transfers'' band.  The end-task mean errors are bias
dominated at the calibrated $\eta=0.4$ ($86$--$100\%$ of MSE; clipping
and horizon not separately ablated), and no coupling is ordered by
end-task MSE under the pre-specified $1.5\times$ rule; on this mildly
skewed posterior (skewness $\le 0.11$) the MAP point---deterministic,
not budget matched, and supplying no posterior samples---sits closer to
the gold mean ($0.0058$) than the sampler arms ($0.010$--$0.011$).
The cell demonstrates that the coupling's \emph{variance} mechanism is
not GLM-specific; it is not an end-task accuracy win.}
\label{tab:gp-hyper}
\resizebox{.75\linewidth}{!}{\begin{tabular}{lrrr}
\toprule
 & iid & $C_\star$ & pairwise \\
\midrule
trace variance / iid & 1 & 0.0596 [0.0359, 0.0701] & 0.0570 [0.0410, 0.0650] \\
median $\|\widehat\theta-\theta_{\mathrm{gold}}\|^2$ & 0.0107 [0.00961, 0.0112] & 0.0107 [0.0100, 0.0107] & 0.0103 [0.0101, 0.0105] \\
\midrule
MAP $\|\theta-\theta_{\mathrm{gold}}\|^2$: 0.00579 & \multicolumn{3}{r}{gold spread 1.5e-05; gold skewness -0.02, -0.11, 0.02} \\
\bottomrule
\end{tabular}
}
\end{table}

\begin{table}[htbp]
\centering
\small
\caption{What if we use the same posterior draws to estimate uncertainty?
For each observable we divide its variance across ensembles by that
under iid sampling and report the median [min--max]. Estimating a
posterior variance or interval width costs precision with either
coupling, more with pairs than with $C_\star$. This agrees in ordering
with the Gaussian quadratic factors $2/K$ and $1/(K-1)$ in
Proposition~\ref{prop:quad-observable}. For mean predictive
probabilities both couplings still give substantial variance reductions.}
\label{tab:uncertainty-observables}
\resizebox{\linewidth}{!}{\begin{tabular}{llrrrrrr}
\toprule
Dataset & prior & $\hat{v}$ $C_\star$/iid & $\hat{v}$ pairwise/iid & CIW $C_\star$/iid & CIW pairwise/iid & pred.\ $C_\star$/iid & pred.\ pairwise/iid \\
\midrule
WDBC & 0.1 & 1.22 [0.935, 1.48] & 2.04 [1.82, 2.18] & 1.12 [1.07, 1.27] & 1.85 [1.82, 1.90] & 0.182 [0.150, 0.191] & 0.242 [0.203, 0.242] \\
WDBC & 0.3 & 1.22 [0.941, 1.39] & 1.93 [1.58, 2.38] & 1.06 [0.984, 1.28] & 1.72 [1.64, 2.18] & 0.119 [0.109, 0.129] & 0.156 [0.136, 0.159] \\
Spambase & 0.1 & 1.22 [1.04, 1.28] & 2.16 [2.01, 2.47] & 1.14 [1.13, 1.20] & 1.97 [1.91, 1.97] & 0.125 [0.123, 0.126] & 0.129 [0.125, 0.134] \\
Spambase & 0.3 & 1.45 [1.28, 1.52] & 2.41 [2.12, 2.85] & 1.16 [1.12, 1.31] & 2.04 [1.72, 2.07] & 0.0946 [0.0926, 0.0962] & 0.0745 [0.0743, 0.0781] \\
Adult & 0.1 & 1.29 [1.22, 1.38] & 2.11 [2.09, 2.12] & 1.20 [1.13, 1.22] & 1.97 [1.97, 2.02] & 0.138 [0.120, 0.140] & 0.126 [0.118, 0.132] \\
Adult & 0.3 & 1.39 [1.37, 2.36] & 2.70 [2.62, 2.98] & 1.22 [1.21, 1.40] & 2.18 [1.97, 2.25] & 0.111 [0.105, 0.117] & 0.109 [0.0995, 0.111] \\
\bottomrule
\end{tabular}
}
\end{table}

\begin{table}[htbp]
\centering
\small
\caption{Does the even-$K$ comparison change with the prior? For each fitted
posterior the table gives median [min--max] matched trace-variance
ratios against iid. Replacing the Gaussian prior with a Student-$t(3)$
prior of the same variance favors pairs over $C_\star$ on the tested
cells. Both designs still reduce variance relative to iid, as they do
in the hierarchical random-intercept fit to Adult.}
\label{tab:nonlogconcave-real}
\resizebox{.85\linewidth}{!}{\begin{tabular}{llrrrr}
\toprule
\multicolumn{6}{l}{\emph{Student-$t$ prior ($\nu=3$)}} \\
Dataset & prior & $C_\star$/iid & pairwise/iid & & \\
\midrule
WDBC & 0.1 & 0.215 [0.210, 0.216] & 0.141 [0.140, 0.144] & & \\
WDBC & 0.3 & 0.245 [0.236, 0.268] & 0.161 [0.159, 0.178] & & \\
Spambase & 0.1 & 0.231 [0.229, 0.255] & 0.0864 [0.0758, 0.0955] & & \\
Spambase & 0.3 & 0.274 [0.262, 0.285] & 0.0797 [0.0773, 0.0817] & & \\
Adult & 0.1 & 0.194 [0.194, 0.201] & 0.0182 [0.0171, 0.0202] & & \\
Adult & 0.3 & 0.272 [0.257, 0.276] & 0.0191 [0.0185, 0.0211] & & \\
\midrule
\multicolumn{6}{l}{\emph{Hierarchical logistic (Adult)}} \\
Dataset & prior & full $C_\star$/iid & $\beta$-block $C_\star$/iid & full pairwise/iid & $\beta$-block pairwise/iid \\
\midrule
Adult & 0.1 & 0.0270 [0.0231, 0.0304] & $7.83\times 10^{-4}$ [$7.81\times 10^{-4}$, $8.76\times 10^{-4}$] & 0.0147 [0.00623, 0.0190] & $7.95\times 10^{-4}$ [$7.77\times 10^{-4}$, $9.66\times 10^{-4}$] \\
\bottomrule
\end{tabular}
}
\end{table}
\FloatBarrier

\subsection{Composition with zero-variance control variates}\label{app:exp-zv}
Starting at zero with $K=8$, we ran three replicates on WDBC, Spambase,
Adult, WDBC with a $t(3)$ prior, and hierarchical Adult, respectively
using $G=256,192,192,256,192$ independent ensembles per arm per replicate.
The first, second and fourth targets used 900 steps, the others 700. We
discarded the first half. At step size $0.08$ we accumulated the
state and score before the update, so the correction used the gradient
at the recorded point. Each fit used a separate iid sample from 256
ensembles at four times the campaign length. We froze the resulting map
for all three couplings. The alternative fits one map to all retained
samples within each evaluation ensemble. This measures the usual same-run
procedure, but the chain contributions then share fitted coefficients.
Noise seeds were $961{,}000{,}000+100{,}000r+1{,}000c+j$, with replicate
$r=0,1,2$, posterior index $c=0,\ldots,4$ in the order above and method
$j=0,1,2$ for iid, pairwise and $C_\star$. Fitting seeds were
$960{,}000{,}000+1{,}000c$. This campaign used three NVIDIA
A100-PCIE-40GB GPUs.

\begin{table}[htbp]
\centering\footnotesize\setlength{\tabcolsep}{3pt}
\caption{Variance after ZV correction, relative to iid with the same
fitting scheme and feature degree. Mean is posterior-mean trace variance.
Pred.\ is variance of the mean predictive probability on test data, or on
training data for hierarchical Adult. Each entry gives the pooled ratio
and its two-standard-error interval. The bands are T (transfers, $R\le0.5$),
P (partial, $0.5<R\le0.8$), E (parity, $0.8<R<1.25$) and H (hurts,
$R\ge1.25$). U means the interval crossed a cutoff or replicate bands
disagreed. $^p$The same-run pairwise contrast is descriptive and post hoc,
computed from the archived per-ensemble estimates. $\dagger$The frozen
Adult mean result at degree one is fit-noise-limited for both couplings.
Refitting on the two halves of the training run changed the band. We show
the numbers for completeness and withhold their interpretation.}
\label{tab:zv-grid}
\begin{tabular}{llccccc}
\toprule
 & & & \multicolumn{2}{c}{Frozen fit} & \multicolumn{2}{c}{Same-run fit} \\
\cmidrule(lr){4-5}\cmidrule(lr){6-7}
Posterior & Obs. & Degree & $C_\star$ & Pairwise & $C_\star$ & Pairwise$^p$ \\
\midrule
WDBC & Mean & 1 & \shortstack{1.04 E \\{} [0.893, 1.22]} & \shortstack{1.48 U \\{} [1.27, 1.72]} & \shortstack{0.351 T \\{} [0.299, 0.413]} & \shortstack{0.29 T \\{} [0.252, 0.334]} \\[2pt]
 &  & 2 & \shortstack{0.758 U \\{} [0.645, 0.89]} & \shortstack{1.06 E \\{} [0.909, 1.24]} & \shortstack{0.265 T \\{} [0.233, 0.302]} & \shortstack{0.374 T \\{} [0.327, 0.429]} \\[2pt]
 & Pred. & 1 & \shortstack{1.23 U \\{} [1.06, 1.43]} & \shortstack{1.63 H \\{} [1.41, 1.88]} & \shortstack{0.459 U \\{} [0.396, 0.533]} & \shortstack{0.497 U \\{} [0.43, 0.574]} \\[2pt]
 &  & 2 & \shortstack{0.968 E \\{} [0.842, 1.11]} & \shortstack{1.18 U \\{} [1.02, 1.37]} & \shortstack{0.495 U \\{} [0.425, 0.577]} & \shortstack{0.666 P \\{} [0.572, 0.777]} \\[2pt]
\addlinespace
Spambase & Mean & 1 & \shortstack{1.25 U \\{} [1.07, 1.47]} & \shortstack{1.38 U \\{} [1.21, 1.58]} & \shortstack{0.508 U \\{} [0.425, 0.606]} & \shortstack{0.37 T \\{} [0.322, 0.424]} \\[2pt]
 &  & 2 & \shortstack{1.08 U \\{} [0.907, 1.28]} & \shortstack{0.887 U \\{} [0.766, 1.03]} & \shortstack{0.316 T \\{} [0.278, 0.359]} & \shortstack{0.344 T \\{} [0.301, 0.395]} \\[2pt]
 & Pred. & 1 & \shortstack{1.29 U \\{} [1.1, 1.52]} & \shortstack{1.11 U \\{} [0.95, 1.29]} & \shortstack{0.382 T \\{} [0.323, 0.452]} & \shortstack{0.339 T \\{} [0.289, 0.399]} \\[2pt]
 &  & 2 & \shortstack{1.24 U \\{} [1.06, 1.45]} & \shortstack{0.996 E \\{} [0.853, 1.16]} & \shortstack{0.536 U \\{} [0.455, 0.631]} & \shortstack{0.657 P \\{} [0.551, 0.783]} \\[2pt]
\addlinespace
Adult & Mean & 1 & \shortstack{0.526 $\dagger$ \\{} [0.45, 0.616]} & \shortstack{0.649 $\dagger$ \\{} [0.553, 0.762]} & \shortstack{0.622 P \\{} [0.548, 0.707]} & \shortstack{0.658 U \\{} [0.579, 0.747]} \\[2pt]
 &  & 2 & \shortstack{0.788 U \\{} [0.646, 0.962]} & \shortstack{0.81 U \\{} [0.673, 0.975]} & \shortstack{0.343 T \\{} [0.29, 0.406]} & \shortstack{0.471 U \\{} [0.393, 0.566]} \\[2pt]
 & Pred. & 1 & \shortstack{0.856 U \\{} [0.729, 1]} & \shortstack{0.92 U \\{} [0.782, 1.08]} & \shortstack{0.704 U \\{} [0.594, 0.835]} & \shortstack{0.982 E \\{} [0.826, 1.17]} \\[2pt]
 &  & 2 & \shortstack{0.87 U \\{} [0.741, 1.02]} & \shortstack{0.948 E \\{} [0.807, 1.11]} & \shortstack{0.543 U \\{} [0.449, 0.657]} & \shortstack{0.677 U \\{} [0.558, 0.821]} \\[2pt]
\addlinespace
$t(3)$ & Mean & 1 & \shortstack{0.824 U \\{} [0.771, 0.881]} & \shortstack{0.616 P \\{} [0.571, 0.664]} & \shortstack{0.663 P \\{} [0.617, 0.713]} & \shortstack{0.469 U \\{} [0.433, 0.508]} \\[2pt]
 &  & 2 & \shortstack{0.823 U \\{} [0.767, 0.882]} & \shortstack{0.366 T \\{} [0.335, 0.4]} & \shortstack{0.608 P \\{} [0.568, 0.652]} & \shortstack{0.402 T \\{} [0.371, 0.436]} \\[2pt]
 & Pred. & 1 & \shortstack{0.9 U \\{} [0.774, 1.05]} & \shortstack{0.944 E \\{} [0.8, 1.11]} & \shortstack{0.6 P \\{} [0.514, 0.7]} & \shortstack{0.578 U \\{} [0.488, 0.685]} \\[2pt]
 &  & 2 & \shortstack{0.86 U \\{} [0.737, 1]} & \shortstack{0.644 P \\{} [0.553, 0.748]} & \shortstack{0.672 P \\{} [0.58, 0.779]} & \shortstack{0.819 U \\{} [0.699, 0.96]} \\[2pt]
\addlinespace
Hierarchical & Mean & 1 & \shortstack{0.631 U \\{} [0.55, 0.723]} & \shortstack{0.344 T \\{} [0.286, 0.413]} & \shortstack{0.504 U \\{} [0.458, 0.553]} & \shortstack{0.38 T \\{} [0.305, 0.473]} \\[2pt]
 &  & 2 & \shortstack{0.646 U \\{} [0.56, 0.745]} & \shortstack{0.313 T \\{} [0.254, 0.384]} & \shortstack{0.458 U \\{} [0.422, 0.497]} & \shortstack{0.397 T \\{} [0.363, 0.434]} \\[2pt]
 & Pred. & 1 & \shortstack{0.0175 T \\{} [0.0147, 0.0209]} & \shortstack{0.0209 T \\{} [0.0176, 0.0249]} & \shortstack{0.236 T \\{} [0.195, 0.286]} & \shortstack{0.323 T \\{} [0.26, 0.4]} \\[2pt]
 &  & 2 & \shortstack{0.165 T \\{} [0.134, 0.205]} & \shortstack{0.26 T \\{} [0.213, 0.318]} & \shortstack{0.782 U \\{} [0.208, 2.93]} & \shortstack{0.335 U \\{} [0.132, 0.849]} \\[2pt]
\addlinespace
\bottomrule
\end{tabular}

\end{table}

\noindent Before ZV the $C_\star$/iid posterior-mean ratios, in table order,
were $0.0054,0.0016,0.00052,0.22,0.025$. Table~\ref{tab:zv-diagnostics}
gives the absolute iid variances on a log scale.
Frozen fits retain coefficient error and can inflate the iid denominator.
Same-run fitting often gave smaller coupled/iid ratios but the ordering
depended on the observable. These are variance comparisons. Two predictions
missed their marks. The degree-two Adult posterior-mean pairwise ratio was
$0.81$ [$0.67,0.97$] despite a prediction above one. On $t(3)$ the mean ratios were $0.824$ and $0.823$ at degrees one and two. The predictive ratios were $0.900$ and $0.860$. All four exceeded the predicted $0.8$, which lay inside each interval.

\begin{table}[htbp]
\centering\small
\caption{Absolute variances and the gain from fitting a control variate.
The first three numeric columns are base-ten log iid variances.
Multiplying the corresponding variance by a ratio from
Table~\ref{tab:zv-grid} recovers that coupled variance up to rounding.
The next three are base-ten log ratios of frozen ZV variance to the plain
variance within each arm. The last column is the natural log of the
$C_\star$/iid ratio after frozen ZV minus its value on the log scale before
ZV, with an approximate two-standard-error interval that omits the covariance from shared runs. The degree-one Adult
mean entries share the fitting caveat in Table~\ref{tab:zv-grid}.}
\label{tab:zv-diagnostics}
\resizebox{\linewidth}{!}{\begin{tabular}{llcrrrrrrc}
\toprule
 & & & \multicolumn{3}{c}{iid log variance} & \multicolumn{3}{c}{Log variance change} & \\
\cmidrule(lr){4-6}\cmidrule(lr){7-9}
Posterior & Obs. & Deg. & Plain & Frozen & Same-run & iid & $C_\star$ & Pairwise & Interaction \\
\midrule
WDBC & Mean & 1 & 1.242 & -0.478 & -0.034 & -1.720 & 0.566 & 0.532 & 5.26 [5.08, 5.44] \\
 &  & 2 & 1.242 & -0.494 & 0.147 & -1.736 & 0.412 & 0.373 & 4.94 [4.76, 5.13] \\
 & Pred. & 1 & -2.769 & -4.958 & -4.525 & -2.189 & -0.633 & -0.571 & 3.58 [3.37, 3.79] \\
 &  & 2 & -2.769 & -5.048 & -4.446 & -2.279 & -0.828 & -0.800 & 3.34 [3.14, 3.54] \\
\addlinespace
Spambase & Mean & 1 & 1.450 & -0.990 & -0.546 & -2.440 & 0.455 & 0.472 & 6.66 [6.48, 6.85] \\
 &  & 2 & 1.450 & -1.090 & -0.397 & -2.540 & 0.290 & 0.180 & 6.52 [6.32, 6.72] \\
 & Pred. & 1 & -2.720 & -5.910 & -5.349 & -3.190 & -1.204 & -1.313 & 4.57 [4.32, 4.82] \\
 &  & 2 & -2.720 & -5.937 & -5.228 & -3.217 & -1.250 & -1.387 & 4.53 [4.28, 4.78] \\
\addlinespace
Adult & Mean & 1 & 1.774 & -0.532 & -0.036 & -2.306 & 0.700 & 0.818 & 6.92 [6.76, 7.08] \\
 &  & 2 & 1.774 & -0.704 & 0.442 & -2.478 & 0.704 & 0.743 & 7.32 [7.12, 7.53] \\
 & Pred. & 1 & -2.852 & -6.939 & -6.001 & -4.087 & -2.230 & -2.306 & 4.28 [4.04, 4.51] \\
 &  & 2 & -2.852 & -6.960 & -5.777 & -4.108 & -2.244 & -2.314 & 4.29 [4.06, 4.53] \\
\addlinespace
$t(3)$ & Mean & 1 & 1.132 & 0.631 & 0.746 & -0.501 & 0.076 & 0.114 & 1.33 [1.24, 1.42] \\
 &  & 2 & 1.132 & 0.539 & 0.675 & -0.593 & -0.016 & -0.204 & 1.33 [1.24, 1.42] \\
 & Pred. & 1 & -2.691 & -4.613 & -4.436 & -1.922 & -0.596 & -0.616 & 3.05 [2.83, 3.27] \\
 &  & 2 & -2.691 & -4.759 & -4.562 & -2.068 & -0.761 & -0.928 & 3.01 [2.79, 3.23] \\
\addlinespace
Hierarchical & Mean & 1 & 1.725 & 0.376 & 0.915 & -1.349 & 0.050 & 0.120 & 3.22 [3.04, 3.40] \\
 &  & 2 & 1.725 & 0.350 & 1.016 & -1.375 & 0.034 & 0.052 & 3.25 [3.06, 3.43] \\
 & Pred. & 1 & -2.791 & -13.131 & -10.656 & -10.340 & -10.321 & -10.328 & 0.05 [-0.20, 0.30] \\
 &  & 2 & -2.791 & -10.939 & -7.311 & -8.148 & -7.153 & -7.041 & 2.29 [2.01, 2.57] \\
\addlinespace
\bottomrule
\end{tabular}
}
\end{table}
\FloatBarrier

\subsection{Mode occupancy and regression benchmarks}\label{app:exp-occupancy}
In a mixture of two unit-covariance Gaussians in $d=10$, the component
means lie at $\pm a$ on the first axis. We varied the separation, weights,
initialization, chain count and step size as specified in
Table~\ref{tab:bimodal}. Across three replicates we used $G=256$ independent
ensembles per arm and replicate, with starts shared across methods and
independent innovation streams. Runs from the mixture law discarded the
first half of 200,000 steps. When all chains started in one basin we kept
the whole trajectory. For cell $c$ in table order and replicate $r$
(starting at zero), the start seed was $951{,}000{,}000+100{,}000r+1{,}000c$.
Noise seeds used base $952{,}000{,}000$ plus $100{,}000r+1{,}000c$ and
method index $0,1,2,3$ for iid, $C_\star$, pairwise and common.

The original first-passage check used the median within each ensemble.
That median can depend on the coupling even when per-chain hitting-time laws
are identical; common noise failed the check in all six mixture-start cells.
Before examining the scored variance ratios, we replaced it with a comparison
of ensemble means of uncensored per-chain first-passage times. We retained
the four-standard-error threshold, reported pooled quartiles separately and
reused the original trajectories. The dated amendment is included in the
code repository.

\begin{table}[htbp]
\centering\footnotesize\setlength{\tabcolsep}{3pt}
\caption{Variance ratio of the estimated occupancy fraction relative to iid,
with two-standard-error intervals. The last column gives the centered
replicate median [minimum, maximum]. Band letters denote transfers
(T, $R\le0.5$), partial (P, $0.5<R\le0.8$), parity (E, $0.8<R<1.25$)
and hurts (H, $R\ge1.25$). U means an interval crosses a band boundary or
replicates disagree on the band. The default settings were $K=8$,
$\eta=0.25$ and equal mixture weights. The one-basin start and $K=16$
used $a=4$, while the asymmetric and $\eta=0.1$ comparisons used $a=2$.
$^*$At equal weights the drift is odd. Once a pair mirrors, opposite innovations keep it mirrored and its occupancy is $1/2$. Starts were iid across chains. The reported zeros are at displayed precision. Three of the five ratios are positive below $2\times10^{-9}$, and two are exactly zero. The predicted near-equality of pairwise and centered occupancy variance failed here. In the asymmetric run the intervals for the two
antithetic constructions overlap.}
\label{tab:bimodal}
\begin{tabular}{lcccc}
\toprule
Setting & $C_\star$ & Pairwise & Common & $C_\star$ replicates \\
\midrule
Mild ($a=2$) & \shortstack{0.440 U \\{} [0.380, 0.510]} & $0^{*}$ T & \shortstack{7.845 H \\{} [6.780, 9.077]} & 0.454 [0.367, 0.496] \\[2pt]
Intermediate ($a=3.5$) & \shortstack{1.030 E \\{} [0.897, 1.182]} & $0^{*}$ T & \shortstack{7.633 H \\{} [6.661, 8.746]} & 1.084 [0.865, 1.171] \\[2pt]
High barrier ($a=4$) & \shortstack{1.005 E \\{} [0.874, 1.155]} & $0^{*}$ T & \shortstack{8.281 H \\{} [7.225, 9.490]} & 0.991 [0.939, 1.086] \\[2pt]
Asymmetric ($w=0.7$) & \shortstack{0.546 U \\{} [0.471, 0.631]} & \shortstack{0.489 U \\{} [0.423, 0.565]} & \shortstack{7.743 H \\{} [6.717, 8.926]} & 0.533 [0.532, 0.570] \\[2pt]
One-basin start & \shortstack{0.996 E \\{} [0.866, 1.146]} & \shortstack{0.031 T \\{} [0.026, 0.037]} & \shortstack{7.730 H \\{} [6.709, 8.907]} & 0.943 [0.925, 1.132] \\[2pt]
$K=16$ & \shortstack{1.053 E \\{} [0.915, 1.211]} & $0^{*}$ T & \shortstack{15.836 H \\{} [13.805, 18.166]} & 0.982 [0.943, 1.223] \\[2pt]
$\eta=0.1$ & \shortstack{0.500 U \\{} [0.436, 0.573]} & $0^{*}$ T & \shortstack{9.320 H \\{} [8.096, 10.729]} & 0.456 [0.440, 0.625] \\[2pt]
\bottomrule
\end{tabular}

\end{table}

\noindent From a one-basin start the $C_\star$/iid occupancy ratios at
$2{,}000,\allowbreak 5{,}000,\allowbreak 10{,}000,\allowbreak 20{,}000,\allowbreak
50{,}000,\allowbreak 100{,}000,\allowbreak 200{,}000$
steps were $1.182,\allowbreak 1.008,\allowbreak 0.951,\allowbreak 1.067,\allowbreak
1.211,\allowbreak 1.066,\allowbreak 0.996$. The first
interval $[0.969,1.441]$ and the fifth $[1.056,1.389]$ crossed the upper
parity cutoff. Both were unresolved, and at the fifth checkpoint the
entire interval was above one. The other five checkpoints classified as
parity. No main centered comparison classified as hurts.

\paragraph{Regression benchmarks.} We used the Adult horizon for MNIST 7-vs-9~\cite{lecun1998mnist}
($12214/2037$ train/test, $d=784$ standardized pixels plus intercept) and
Gisette~\cite{uci_gisette} ($6000/1000$, $d=5000$ plus intercept). For the Gaussian check
we fitted Bayesian linear regression to the standardized UCI
Superconductivity data~\cite{uci_superconduct} ($17010/4253$, $d=81$).
We estimated the noise variance by ridge regression and used a step
size of $0.5/\lambda_{\max}$, where $\lambda_{\max}$ is the largest
eigenvalue of the posterior precision. The closed-form posterior mean
gives an exact reference.

\begin{table}[htbp]
\centering
\small
\caption{High-dimensional real logistic posteriors.  Matched trace-variance
ratios against iid (median [min--max]) at $K=8$.  The real-data reduction
is not a small-dimension artifact.}
\label{tab:highdim-real}
\resizebox{\linewidth}{!}{\begin{tabular}{llrrrr}
\toprule
Dataset & prior & $d$ & $C_\star$/iid & pairwise/iid & common/iid \\
\midrule
MNIST 7v9 & 0.1 & 785 & $7.93\times 10^{-5}$ [$7.59\times 10^{-5}$, $8.46\times 10^{-5}$] & $1.14\times 10^{-4}$ [$1.13\times 10^{-4}$, $1.16\times 10^{-4}$] & 8.01 [7.98, 8.06] \\
MNIST 7v9 & 0.3 & 785 & $4.77\times 10^{-5}$ [$4.74\times 10^{-5}$, $5.19\times 10^{-5}$] & $7.04\times 10^{-5}$ [$6.93\times 10^{-5}$, $7.57\times 10^{-5}$] & 7.97 [7.75, 8.12] \\
Gisette & 0.1 & 5001 & $2.54\times 10^{-5}$ [$2.50\times 10^{-5}$, $2.59\times 10^{-5}$] & $3.28\times 10^{-5}$ [$3.26\times 10^{-5}$, $3.31\times 10^{-5}$] & 8.02 [7.98, 8.03] \\
Gisette & 0.3 & 5001 & $9.23\times 10^{-6}$ [$9.03\times 10^{-6}$, $9.26\times 10^{-6}$] & $1.04\times 10^{-5}$ [$1.01\times 10^{-5}$, $1.05\times 10^{-5}$] & 7.97 [7.93, 7.98] \\
\bottomrule
\end{tabular}
}
\end{table}

\begin{table}[htbp]
\centering
\small
\caption{Exact-identity showcase: Bayesian linear regression on the UCI
Superconductivity data.  The posterior is exactly Gaussian, so
Theorem~\ref{thm:panel} applies verbatim on real data: measured centered
trace variance sits at the floating-point floor and the error against the
closed-form posterior mean is purely bias.  This row is an exactness
measurement, not a proposed workflow; the closed form exists precisely
because the target is Gaussian.}
\label{tab:bayes-linreg-exact}
\resizebox{\linewidth}{!}{\begin{tabular}{llrrrrr}
\toprule
$\lambda$ & dtype & iid trace var & $C_\star$ trace var & $C_\star$/iid & $C_\star$ MSE & $C_\star$ bias$^2$ \\
\midrule
0.1 & float32 & 18.5 [18.4, 18.6] & $4.03\times 10^{-12}$ [$4.02\times 10^{-12}$, $4.21\times 10^{-12}$] & $2.17\times 10^{-13}$ [$2.17\times 10^{-13}$, $2.29\times 10^{-13}$] & 0.153 [0.153, 0.153] & 0.153 [0.153, 0.153] \\
0.1 & float64 & 18.7 [18.4, 18.9] & $1.40\times 10^{-29}$ [$1.38\times 10^{-29}$, $1.40\times 10^{-29}$] & $7.50\times 10^{-31}$ [$7.31\times 10^{-31}$, $7.59\times 10^{-31}$] & 0.153 [0.153, 0.153] & 0.153 [0.153, 0.153] \\
0.3 & float32 & 12.3 [12.3, 12.6] & $2.34\times 10^{-12}$ [$2.33\times 10^{-12}$, $2.34\times 10^{-12}$] & $1.90\times 10^{-13}$ [$1.85\times 10^{-13}$, $1.90\times 10^{-13}$] & 0.0150 [0.0150, 0.0150] & 0.0150 [0.0150, 0.0150] \\
0.3 & float64 & 12.3 [12.3, 12.4] & $8.15\times 10^{-30}$ [$7.97\times 10^{-30}$, $8.22\times 10^{-30}$] & $6.59\times 10^{-31}$ [$6.45\times 10^{-31}$, $6.70\times 10^{-31}$] & 0.0150 [0.0150, 0.0150] & 0.0150 [0.0150, 0.0150] \\
\bottomrule
\end{tabular}
}
\end{table}
\FloatBarrier

\subsection{Finite-sum gradients and step timings}\label{app:exp-additional}
We used the quadratic model to separate the two noises in an SGLD
step. There were $64$ components in eight dimensions, so eight chains
with batches of eight could use each component exactly once between
them. With identical component Hessians and balanced batches, the
only remaining noise should come from the Gaussian draws
(Theorem~\ref{thm:finite-sum}). For iid draws we found $0.1258$ of
the single-chain terminal variance, close to the predicted $1/8$.
Returning to iid batches and centering only the Gaussian draws gave
$1.44\times10^{-5}$. When we used both constructions together the
result depended on the Hessians. The ratio was $1.22\times10^{-32}$
when they agreed and $2.02\times10^{-5}$ when they differed, in which
case $R_t$ need not vanish. The full diagnostic table is in the
code repository.

\paragraph{Distributed step times.} Our timings on three GPUs answer
a narrower question than the single-device cost comparison in
Table~\ref{tab:campaign-wallclock-frontier}. If the chains are already
split across devices, how much longer does one step take? We sharded
$8$ or $32$ chains across three PCIe A100s on one node, using NCCL to
sum the innovations. In five repetitions of $200$ timed steps the
coupled-to-iid step-time ratio was $1.34$--$1.36$. On one GPU the
largest real-data ratio was $1.14$. Only a synthetic step taking
microseconds reached $1.36$ (Table~\ref{tab:campaign-wallclock}).

\begin{table}[htbp]
\centering
\small
\caption{Single-device timing at $K=8$. Overhead gives the median $C_\star$
sampling seconds per ensemble divided by the iid seconds on the same
target at the same chain count. For comparison we extrapolated the
iid chain count that would match the synthetic terminal projection
variance $\operatorname{Var}(a^\top\bar X_t)$ or the real-data MSE
against the finite reference if the metric fell as $1/K$. The
synthetic entries at the numerical floor are capped. Table~\ref{tab:campaign-wallclock-frontier} reports the measured
variance-time comparison.}
\label{tab:campaign-wallclock}
\resizebox{.9\linewidth}{!}{\begin{tabular}{lrrrr}
\toprule
Target & $C_\star$ K=8 metric & iid $1/K$ extrap. K & $C_\star$ sec/ens & overhead \\
\midrule
gaussian & 1.12e-15 & $>10^6$ & 1.33e-04 & 1.361 \\
quartic & 1.01e-12 & $>10^6$ & 2.86e-04 & 1.239 \\
synthetic logistic & 3.20e-10 & $>10^6$ & 1.71e-04 & 1.288 \\
wdbc & 0.864 & 122 & 0.003 & 1.142 \\
spambase & 1.245 & 119 & 0.004 & 1.128 \\
adult & 9.045 & 45.3 & 0.013 & 0.998 \\
\bottomrule
\end{tabular}
}
\end{table}

\FloatBarrier
\clearpage

\section*{NeurIPS Paper Checklist}

\begin{enumerate}

\item {\bf Claims}
    \item[] Question: Do the main claims made in the abstract and introduction accurately reflect the paper's contributions and scope?
    \item[] Answer: \answerYes{}.
    \item[] Justification: The abstract and introduction state the exact marginal-preservation, quadratic cancellation, Hermite-chaos, minibatch, and nonlinear-remainder claims, and the Limitations section states their scope.

\item {\bf Limitations}
    \item[] Question: Does the paper discuss the limitations of the work performed by the authors?
    \item[] Answer: \answerYes{}.
    \item[] Justification: Section~\ref{sec:limitations} distinguishes the proved range from measured behavior, including synchronization cost, quadratic penalties, the failed tuning and reference checks, and the scope of the bimodal result.

\item {\bf Theory assumptions and proofs}
    \item[] Question: For each theoretical result, does the paper provide the full set of assumptions and a complete (and correct) proof?
    \item[] Answer: \answerYes{}.
    \item[] Justification: Theorems~\ref{thm:panel}--\ref{thm:variance-ratio} state their hypotheses in the main text. Appendix~\ref{app:marginal} proves the conditional path law and consistency, and Appendices~\ref{app:identities}--\ref{app:nonquadratic} supply the covariance and curvature arguments.

    \item {\bf Experimental result reproducibility}
    \item[] Question: Does the paper fully disclose all the information needed to reproduce the main experimental results of the paper to the extent that it affects the main claims and/or conclusions of the paper (regardless of whether the code and data are provided or not)?
    \item[] Answer: \answerYes{}.
    \item[] Justification: The repository scripts generate the synthetic targets and download and preprocess the cited UCI datasets for logistic regression. The README gives the environment, commands and steps we used to generate the paper artifacts.

\item {\bf Open access to data and code}
    \item[] Question: Does the paper provide open access to the data and code, with sufficient instructions to faithfully reproduce the main experimental results, as described in supplemental material?
    \item[] Answer: \answerYes{}.
\item[] Justification: The code repository linked in Appendix~\ref{app:experiments} includes the experiment scripts, dependency file, and README instructions needed to reproduce the paper artifacts from fresh outputs.  The real-data campaigns use UCI Machine Learning Repository datasets (Spambase~\cite{uci_spambase}, Wisconsin Diagnostic Breast Cancer~\cite{uci_wdbc}, Adult~\cite{uci_adult}, Gisette~\cite{uci_gisette}, and Superconductivity~\cite{uci_superconduct}) and MNIST digits 7/9~\cite{lecun1998mnist}; the included download scripts (\texttt{scripts/download\_real\_data.py}, \texttt{scripts/download\_highdim\_data.py}, \texttt{scripts/download\_regression\_data.py}) fetch and preprocess them from the original sources and record DOI and license metadata.

\item {\bf Experimental setting/details}
    \item[] Question: Does the paper specify all the training and test details (e.g., data splits, hyperparameters, how they were chosen, type of optimizer) necessary to understand the results?
    \item[] Answer: \answerYes{}.
    \item[] Justification: The experiment section specifies the swept dimensions, chain counts, target families, horizons, and aggregation protocol; Appendix~\ref{app:experiments} gives ensemble counts and seed rules for the ZV and bimodal studies. The dated amendment to the bimodal protocol is included in the code repository.

\item {\bf Experiment statistical significance}
    \item[] Question: Does the paper report error bars suitably and correctly defined or other appropriate information about the statistical significance of the experiments?
    \item[] Answer: \answerYes{}.
    \item[] Justification: Tables report medians and quantiles or maxima over independent problem/seed cells, and figures aggregate many independent coupled ensembles rather than a single run.

\item {\bf Experiments compute resources}
    \item[] Question: For each experiment, does the paper provide sufficient information on the computer resources (type of compute workers, memory, time of execution) needed to reproduce the experiments?
    \item[] Answer: \answerYes{}.
    \item[] Justification: The README states that the sweeps ran on NVIDIA A100-PCIE-40GB GPUs with CUDA 12 libraries, and the JSONL outputs in the code repository record elapsed times. The control-variate campaign used three A100-PCIE-40GB GPUs on TACC Lonestar6.

\item {\bf Code of ethics}
    \item[] Question: Does the research conducted in the paper conform, in every respect, with the NeurIPS Code of Ethics \url{https://neurips.cc/public/EthicsGuidelines}?
    \item[] Answer: \answerYes{}.
    \item[] Justification: The work is a theoretical and simulation study using standard public benchmark datasets (UCI, MNIST); it involves no human-subjects research, private or sensitive data collection, scraping, or deployment claims.

\item {\bf Broader impacts}
    \item[] Question: Does the paper discuss both potential positive societal impacts and negative societal impacts of the work performed?
    \item[] Answer: \answerNA{}.
    \item[] Justification: The contribution is a foundational sampling-analysis result with no direct deployment pathway, dataset release, or model release studied in the paper.

\item {\bf Safeguards}
    \item[] Question: Does the paper describe safeguards that have been put in place for responsible release of data or models that have a high risk for misuse (e.g., pre-trained language models, image generators, or scraped datasets)?
    \item[] Answer: \answerNA{}.
    \item[] Justification: No pretrained model, scraped dataset, or high-risk released asset is introduced.

\item {\bf Licenses for existing assets}
    \item[] Question: Are the creators or original owners of assets (e.g., code, data, models), used in the paper, properly credited and are the license and terms of use explicitly mentioned and properly respected?
    \item[] Answer: \answerYes{}.
    \item[] Justification: The paper cites prior methods.  The real-data campaigns use UCI datasets and MNIST, each cited in the bibliography (Spambase~\cite{uci_spambase}, Wisconsin Diagnostic Breast Cancer~\cite{uci_wdbc}, Adult~\cite{uci_adult}, Gisette~\cite{uci_gisette}, Superconductivity~\cite{uci_superconduct}, MNIST~\cite{lecun1998mnist}); license metadata for each dataset is recorded by the download scripts.

\item {\bf New assets}
    \item[] Question: Are new assets introduced in the paper well documented and is the documentation provided alongside the assets?
    \item[] Answer: \answerYes{}.
    \item[] Justification: The README explains the released scripts and results. We introduce no new dataset, benchmark or model checkpoint.

\item {\bf Crowdsourcing and research with human subjects}
    \item[] Question: For crowdsourcing experiments and research with human subjects, does the paper include the full text of instructions given to participants and screenshots, if applicable, as well as details about compensation (if any)?
    \item[] Answer: \answerNA{}.
    \item[] Justification: The work does not involve crowdsourcing or human-subject experiments.

\item {\bf Institutional review board (IRB) approvals or equivalent for research with human subjects}
    \item[] Question: Does the paper describe potential risks incurred by study participants, whether such risks were disclosed to the subjects, and whether Institutional Review Board (IRB) approvals (or an equivalent approval/review based on the requirements of your country or institution) were obtained?
    \item[] Answer: \answerNA{}.
    \item[] Justification: No human-subjects research is conducted.

\item {\bf Declaration of LLM usage}
    \item[] Question: Does the paper describe the usage of LLMs if it is an important, original, or non-standard component of the core methods in this research? Note that if the LLM is used only for writing, editing, or formatting purposes and does \emph{not} impact the core methodology, scientific rigor, or originality of the research, declaration is not required.
    \item[] Answer: \answerYes{}.
    \item[] Justification: We used LLM tools in brainstorming, editing prose, revising and debugging code, and checking the presentation. The author made the research decisions and checked the theory, code, experiments and reported results.

\end{enumerate}

\end{document}